\documentclass[letterpaper]{article} 
\usepackage{aaai25}  
\usepackage{times}  
\usepackage{helvet}  
\usepackage{courier}  
\usepackage[hyphens]{url}  
\usepackage{graphicx} 
\usepackage{natbib}  
\usepackage{caption} 
\usepackage{algorithm}
\usepackage{algorithmic}
\usepackage{adjustbox}

\usepackage{newfloat}
\usepackage{listings}
\DeclareCaptionStyle{ruled}{labelfont=normalfont,labelsep=colon,strut=off} 
\floatstyle{ruled}
\newfloat{listing}{tb}{lst}{}
\floatname{listing}{Listing}
\title{A Practical Approach to Causal Inference over Time}
\author{
    Martina Cinquini\equalcontrib\textsuperscript{\rm 1}, Isacco Beretta\equalcontrib\textsuperscript{\rm 1}, Salvatore Ruggieri\textsuperscript{\rm 1}, Isabel Valera\textsuperscript{\rm 2}
}
\affiliations{
    \textsuperscript{\rm 1}Department of Computer Science, University of Pisa, Pisa, Italy\\
    \textsuperscript{\rm 2}Department of Computer Science, Saarland University, Saarbr\"ucken, Germany\\

    \{martina.cinquini, isacco.beretta\}@phd.unipi.it, salvatore.ruggieri@unipi.it, ivalera@cs.uni-saarland.de\\
}

\usepackage[]{our_preamble}
\usetikzlibrary{positioning, fit, arrows.meta, cd, backgrounds, patterns, calc,shapes}

\tikzstyle{hidden vertex} = [circle, draw=black,style=dashed]
\tikzstyle{vertex} = [circle, fill=black!10]
\tikzstyle{box} = [fill=black!10,draw = black,minimum width=0.7cm, minimum height=0.7cm]
\tikzstyle{fixed vertex} = [circle, draw=red,fill=red!10]
\tikzstyle{edge} = [-{Latex[scale=0.7]},thick]
\tikzstyle{o->} = [{Circle[open]}->, thick]
\tikzstyle{o-o} = [{Circle[open]}-{Circle[open]}, thick]
\tikzstyle{<-->} = [<->, thick]
\tikzstyle{redge} = [->,thick, draw=red]
\tikzstyle{uedge} = [thick, draw=black!20]
\tikzstyle{bluedge} = [->,thick, draw=blue]
\tikzstyle{snakedge} = [-stealth,thick, shorten <=1pt,decorate, decoration={snake,amplitude=.3mm}]
\tikzstyle{array}=[draw, fill=green!40, minimum width = 7mm, minimum height = 7mm] 

\newcommand{\orightarrow}{\hbox{$\circ$}\kern-1.5pt\hbox{$\rightarrow$}}
\newcommand{\leftarrowo}{\hbox{$\leftarrow$}\kern-1.5pt\hbox{$\circ$}}
\newcommand{\oarrowo}{\hbox{$\circ$}\kern-1.5pt\hbox{$-$}\kern-1.5pt\hbox{$\circ$}}

\begin{document}

\maketitle

\begin{abstract}
In this paper, we focus on estimating the causal effect of an intervention over time on a dynamical system. 
To that end, we formally define causal interventions and their effects over time on discrete-time stochastic processes (\SP{}s). 
Then, we show under which conditions the equilibrium states of a \SP{}, both before and after a causal intervention, can be captured by a structural causal model (\SCM{}). 
With such an equivalence at hand, we provide an explicit mapping from vector autoregressive models (\VAR{}s), broadly applied in econometrics, to linear, but potentially cyclic and/or affected by unmeasured confounders, \SCM{}s. The resulting \emph{causal \VAR{} framework} allows us to perform \emph{causal inference over time} from observational time series data.  
Our experiments on synthetic and real-world datasets show that the proposed framework achieves strong performance in terms of observational forecasting while enabling accurate estimation of the causal effect of interventions on dynamical systems. 
We demonstrate, through a case study, 
the potential practical questions that can be addressed using the proposed causal \VAR{} framework. 
%
\end{abstract}

\section{Introduction} \label{sec:intro}
Dynamical systems often exhibit complex behaviors that unfold over time, leading to delayed responses and feedback loops. 
Importantly, understanding the causal effect of interventions within such systems is crucial across disciplines such as climate~\citep{runge2019inferring} and social sciences~\citep{wunsch2022time}, where different time scales play a central role. For instance, monetary policy adjustments may have immediate effects on consumer spending, but their impact on inflation, employment, and economic growth only becomes evident in the medium-/long-term. 
Similarly, the consequences of human actions on climate change may take decades to manifest, with the risk of endorsing public policies that underestimate their relevance. To address these issues, it is essential to estimate the causal effect of interventions, or generally, to perform \emph{causal inference over time}. 

From the perspective of causality, Structural Causal Models (SCMs) provide a formal framework to perform causal inference from cross-sectional data. However, adapting existing methods to capture temporal dynamics remains a challenge~\citep{bongers2021foundations}. Alternatively, temporal models, such as autoregressive models, offer practical methods for time-series analysis and forecasting~\citep{lutkepohl2005new}, but their formalization of causal effects is limited. 
First, they model interventions as \emph{shocks} applied at a specific point in time, with effects that fade away after a certain period~\citep{hyvarinen2010estimation, moneta2011causal}. Second, 
they rely on Granger causality~\citep{granger1969investigating} which 
is concerned with how well one variable can predict another rather than identifying causal relationships between them.

Our work combines the strengths of both frameworks, i.e., \SCM{}s and  
autoregressive models, to enable robust reasoning about the causal effect of interventions on dynamical systems over time.
To that end, we first introduce a formal definition of causal interventions on discrete-time stochastic processes (\SP{}s), proposing two alternatives, additive and forcing interventions. Second, we establish conditions under which the equilibrium state of a \SP{} can be represented by an \SCM{}. Third,  we develop a framework that maps \VAR{}s to linear \SCM{}s, handling potentially cyclic structures and unmeasured confounders.   
Finally, our practical framework for causal inference over time from observational time-series data is empirically validated on synthetic and real-world datasets.

\paragraph{Related work} 
 The works most closely related to ours are these from~\citet{Mooij2013deterministic_case} and \citet{bongers2018causal}, as they theoretically connect dynamical systems to the causal semantics of SCMs via the equilibration of deterministic and random differential equations, and thus are capable of  
modeling cyclic causal mechanisms~\citep{bongers2021foundations}.
Our approach differs from this line of work in two key aspects: i) we focus on \emph{discrete-time} dynamical systems parameterized using \emph{stochastic equations} which, as stated by~\citet{bongers2018causal}, become particularly challenging for continuous-time processes; and ii) our mapping from autoregressive \SP{}s to \SCM{}s provides not only a theoretical but also, to the best of our knowledge, \emph{the first data-driven framework for performing causal inference over time in dynamical systems.}

\section{Preliminaries and background} \label{sec:background}

\subsection{Structural Causal Models}
\paragraph{Equation} 
A \SCM{} $\scm[{\rmF,\rmE}]$ determines how a set of \sizex endogenous (observed) random variables $\rmX := \{\Xcomp[1], \dots, \Xcomp[\sizex]\}$ are obtained from a set of 
exogenous variables $ \rmE := \{E_1,\ldots, E_d\}$, with prior distribution $\distribution(\rmE)$, via a set of structural equations $\rmF := \{\Xcomp[i] := f_i(\parents[i], \rmE^{(i)})\}_{i=1}^{\sizex}$. 
Each $f_i$ computes $\Xcomp[i]$ from its causal parents\footnote{Unlike in acyclic \SCM{}s, $\parents[i]$ loses its hierarchical interpretation since two nodes can be mutually parents.}  $\parents[i]\subseteq\X$ and a set $\rmE^{(i)}\subseteq \rmE$. 
We refer to $\X$ as a solution of $\scm$.
We assume $\parents[i]$ to be minimal, i.e.,~it only contains variables $\Xcomp[j]$ such that$ \quad \partial_{\Xcomp[j]}{f_i}\neq~0$. 
This formulation extends the  definition in~\citep{pearl2009causality} 
to include cycles as in~\citep{bongers2018causal}. 

\paragraph{Graph}
A \SCM{} $\scm$ induces a directed graph $ \graph_{\scm} = (\gV, \gE)$ that describes the functional dependencies in $\rmF$: $\gV$ is the set of nodes for which $V_i$ represents $\Xcomp[i]$ and $\gE$ is the set of the edges $(V_i,V_j) \in \gE \iff \Xcomp[i] \in \parents[j]$. 

%
\paragraph{Intervention}
Besides describing the observational distribution $p(\X)$, \SCM{}s allow answering \emph{interventional queries} about the effect of external manipulations, and enable \emph{counterfactual queries} assessing what would have happened to a particular observation if one observed variable $\Xcomp[i]$ had taken a different value. 
%
An intervention $\intervention$ on a \SCM{} $\scm$ yields a new \SCM{} $\scm^\intervention$ for which one or more mechanisms $f_i(\parents[i], \rmE^{(i)})$ change to
$\tilde{f}_i(\tilde{\parentsym}^{(i)}, \tilde{\rmE}^{(i)}),$
where $ \tilde{\parentsym}^{(i)}\subseteq \parents[i] \: \text{ and } \:\tilde{\rmE}^{(i)} \subseteq \rmE^{(i)}$.
We refer to a \textit{hard intervention} when $f_i$ is replaced by a constant value $\alpha^{(i)}$, and $\tilde{\parentsym}^{(i)}=\tilde{\rmE}^{(i)}=\varnothing$. This type of intervention is denoted by the do-operator $do(\Xcomp[i] = \alpha^{(i)})$. 
On the other hand, we refer to a \textit{soft intervention} when at least one argument of $f_i$ is
retained.
The causal effect $\ce$ of an intervention is evaluated in terms of differences between the values of the observable variables before and  after the intervention $\intervention$, i.e.,  
\begin{equation}\label{eq:causal_effect}
\ce^\intervention =\expected{\X^\intervention -\X}.
\end{equation}

%

\subsection{Discrete-time Stochastic Processes }

A discrete-time (vector) stochastic process (\SP{}) is a function
$ \X : \;T \times \Omega \to \sR^\sizex$
%
where $t\in{T}$ is a time index in $\sZ$, such that $\X_t$ (which denotes $\X(t,\cdot)$) is a random variable on a probability space $(\Omega,\gF,\Prb)$. 
We refer to $\X(\omega)$ (which denotes $\X(\cdot,\omega)$)  
as a \textit{realization} or \textit{trajectory} of $\X$ and denote the i-th component of $\X$ with $\Xcomp[i]$. 
%
%
%
Every \SP{} can be described through a difference equation (\DE{}), i.e.,~a recurrence relation that allows computing $\X_t$ based on its past values. 
\DE{}s can be categorized into three types~\citep{bongers2018causal}: 
ordinary difference equations (\ODE{}) describing deterministic processes; random difference equations (\RDE{}), which involve randomness in the initial state $\X_0$ and in the evolution parameters (see \cref{app:difference_equations}); and stochastic difference equations, which describe inherently stochastic trajectories.
%

\paragraph{Equation}
A stochastic difference equation (\SDE{}) describes a \SP{} via a functional relationship of the form
%
\begin{equation}\label{eq:sde}
\X_t = \bm{f}(\X_{<t}) + \bm{g}(\X_{<t}) \odot \vectornoise_t, 
\end{equation}
where $\X_{<t}:=\{\X_{t-1},\X_{t-2},\dots\}$ represents the trajectory up to time $t$, 
$\bm{f}$ represents the system's deterministic mechanism, and the Hadamard product $\bm{g} \odot \vectornoise_t$ is the inhomogeneous stochastic part, where $\vectornoise_t$ denotes \textit{white noise}, i.e.,~$\forall t,t^\prime \in T \quad \expected{\vectornoise_t}=\bm{0}, \expected{\vectornoise_t^2}=\Sigma^2_{\vectornoise}, \expected{\vectornoise_t\vectornoise_{t^\prime}}=\bm{0}$.  
While for \ODE{}s and \RDE{}s trajectories $\X(\omega)$ may asymptotically converge to an equilibrium, \SDE{}s cannot exhibit such convergence due to the ongoing influence of $\bm{g} \odot \vectornoise_t$. 


\paragraph{Graph} 
Analogously to \SCM{}s, we can associate a directed graph \(\graph_{\dequation}\) to a \DE{} \(\dequation\), consisting of nodes \(V_i\) representing individual components \(\Xcomp[i]\), while an edge \((V_i,V_j)\) is present if $\exists k>0$ such that \(\partial_{\Xcomp[i]_t} \Xcomp[j]_{t+k}\neq 0\) in \(\dequation\).\footnote{Depending on the type of \DE{}, the derivative must be evaluated with respect to \(f_i\) or with respect to both \(f_i\) and \(g_i\) (see \Eqref{eq:sde}).}

\subsection{Vector Autoregressive models}
In this paper, we focus on a specific type of \SDE{}, the \VAR{} model~\citep{kilian2017structural}.
%
\paragraph{Equation}
Consider a \sizex-dimensional vector-valued stationary time series $\{\X_0, \dots, \X_T\}$ generated by a \VAR{} model with \emph{lag} $\order$, where a lag represents 
the number of previous time steps used to predict the current value of each variable.
Specifically, the \VAR{}($\order$) model is defined by 
\begin{equation}\label{eq:VAR}
    \X_t = \bm{\nu} + \A_1 \X_{t-1} + \dots + \A_p \X_{t-p} + \vectornoisevar_t, 
\end{equation}
where $\bm{\nu}$ is a \sizex-dimensional vector of intercept terms, 
$\{\A_i\}_{i=1}^p$ are $ (\sizex \times \sizex) $ matrices 
and 
$\vectornoisevar_t$ is a \sizex-dimensional white noise term.
%
If the process $\X_t$ is stable and stationary~\citep{hamilton1994time}, 
%
%
\Eqref{eq:VAR} can also be written as
    \begin{equation}\label{eq:var_lagoperator}
    \begin{split}
        &\A(L)\X_t = \bm{\nu} + \vectornoisevar_t, \\
        \text{with} \quad \:  &\A(L) := \bm{I}_\sizex - \A_1 L - \dots - \A_\order L^\order, 
    \end{split}
    \end{equation}
where $L$ is the \emph{lag operator} such that $L \X_t \equiv \X_{t-1}$ and $\bm{I}_\sizex$ is a \sizex-dimensional identity matrix.  

A key limitation of \VAR{}s is the inability to interpret the system in causal terms since the components of $\vectornoisevar_t$ are cross-correlated and act as hidden confounders.
A common approach to overcome this problem is to orthogonalize the noise terms. 
In this context, the process of \textit{causal discovery}, i.e., inferring the causal structure of the data, is analogous to the one of \SCM{s}~\citep{hyvarinen2010estimation, moneta2013causal, geiger15, malinskyS18}, and involves identifying a triangular matrix $\hat{\A}_0$ such that $\vectornoisesvar_t = \hat{\A}_0 \vectornoisevar_t$ consists of mutually uncorrelated elements. The transformed \VAR{}, commonly known as the Structural VAR (\SVAR{}) model in the literature~\citep{kilian2017structural}, is defined by
$\hat{\A}_0 \X_t = \hat{\A}_0 \bm{\nu} + \hat{\A}_1 \X_{t-1} + \dots + \hat{\A}_\order \X_{t-\order} + \vectornoisesvar_t,
$ where $\hat{\A}_i = \hat{\A}_0 \A_i$. 
%
%
From a modeling perspective, \VAR{} and \SVAR{} are equivalent, 
as any \SVAR{} can be expressed in its reduced-form \VAR{} by computing $\A_i = \hat{\A}_0^{-1}\hat{\A}_i$ for $i=0, \ldots, p$ in \cref{eq:VAR}. Notably, choosing one over the other does not affect its causal interpretation, provided that $\hat{\A}_0$ is known. For simplicity, in this work, we adopt the \VAR{} notation, 
to introduce a \emph{novel framework for causal inference over time, which complements the \SVAR{}'s causal discovery approach}. 

\paragraph{Graph}
An edge $(V_i, V_j)$ is present iff $\exists k.\,\A_k[i, j] \neq 0$.  


%
%
%
%
%


%

\section{Causal perspective on Discrete-time Stochastic Processes} \label{sec:equivalence}
This section provides the theoretical basis for causal inference over time. First, we formally define causal interventions on \SDE{}s (\cref{subsec:causal_int_sde}). 
Then, we show how a \SCM{} can be considered a compressed description of the asymptotic behavior of an underlying dynamical system 
(\cref{subsec:mapping_sde_scm}). 
%

\subsection{Causal Interventions on \SDE{}s}\label{subsec:causal_int_sde}

We define an intervention $\intervention$ on a \SDE{} $\dequation$ as a modification of one or more component equations denoted by the mapping:
\begin{equation}\label{eq:sde_intervention}
\begin{split}
\intervention:  \, &f_i(\parents[i]_{<t}) + g_i(\parents[i]_{<t})\odot\vectornoise_t \quad \longmapsto\\
&\tilde{f_i}(\tilde{\parentsym}_{<t}^{(i)}) + \tilde{g_i}(\tilde{\parentsym}_{<t}^{(i)})\odot \vectornoise_t, \; \forall t\geq t_\intervention
\end{split}
\end{equation}
where \(\tilde{\parentsym}_{<t}^{(i)}\subseteq\parents[i]_{<t}\). 
Unlike \SCM{}s, the intervention applies \textit{starting from a specific time} $t_\intervention$. 
In other words, the process follows the original equations for $t<t_\intervention$ and the modified ones for \(t\geq t_\intervention\). We denote the modified \SDE{} as $\dequation^\intervention$ to 
generalize \cref{eq:causal_effect} to account for time. To differentiate between interventions on a \SCM{} $\scm$ and on a \SDE{} $\dequation$, $\intervention_{\scm}$ and $\intervention_{\dequation}$ will be  respectively adopted when necessary.

\begin{definition}[Causal Effect over time ($\ce{}_t$)]
    Let \(\X\) be a solution of a specific \SDE{} $\dequation$. We define the causal effect at time $t$ of an intervention $\intervention$ as
    \begin{equation}\label{eq:causal_effet_time}
        \ce{}^\intervention_t := \expected{\X^\intervention_t-\X_t|\X_{<t_\intervention}}, 
    \end{equation}
where \(\X^\intervention\) is the solution of the modified \SDE{} $\dequation^\intervention$.
\end{definition}
The interpretation of  $\ce{}^\intervention_t$ is closely related to the causal effect of an intervention on a \SCM{} (\cref{eq:causal_effect}), $\ce{}^\intervention$: while the latter measures the causal effect of an exogenous intervention, \(\ce_t\) does so for any time step $t$ of the \SP, i.e.,~\textit{it measures the causal effect of an intervention over time}. 
%
%
Importantly, as we will show in the next section, \(\ce{}^\intervention_t \to \ce^\intervention{}\) as \(t\to\infty\), i.e.,
there is an asymptotic correspondence between the two quantities.

\subsection{Mapping \SDE{}s to \SCM{}s}
\label{subsec:mapping_sde_scm}
Given an \SDE{} $\dequation$ and its solution $\X$, we study the conditions on $\dequation$ such that:
i) $\X_t$ converges in distribution to $\X_\infty$ as $t \to \infty$; and
ii) there exists an \SCM{} $\scm$ such that $\X_\infty$ is a solution of $\scm$ and, for every intervention $\intervention$, it holds that $(\X_\infty)^{\intervention_{\scm}} =(\X^{\intervention_{\dequation}})_\infty$. While i) is automatically satisfied by any finite memory stationary process, ii) requires more careful analysis, as discussed below. 


%
\paragraph{A negative result from \citep{janzing2018structural}} 
Consider the stable bivariate system defined by the equations $X_t = \varepsilon^{x}_t, Y_t= 0.5 \cdot X_{t-1} + \varepsilon^{y}_t$. For every \( t \), \( X_t \) and \( Y_t \) are independent of each other. Consequently, the joint distribution $p(X_t,Y_t)$ cannot capture the causal dependencies of the system (\textit{$X$ causes $Y$}). The lack of causal information in the cross-sectional dimension arises because the variables are \textit{localized in time}; their values change rapidly, leading to minimal or no correlation with their past values. On this specific point \citet{janzing2018structural} provide an explicit negative result: without first making the variables de-localized in time, there is no \SCM{} that can capture the \SDE{}. In fact, our definition of intervention (\cref{eq:sde_intervention}) \textit{acts on a variable of the system for a prolonged and indefinite period}.

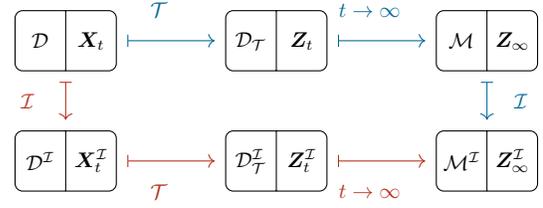
\begin{figure}[t!]
    \centering
\begin{tikzpicture}
    [scale=0.8, every node/.style={rectangle split, rectangle split parts=2,rectangle split horizontal=true, rounded corners=0.1cm, minimum height=1cm, align=center,scale=0.8}]

    \node[draw] (1) at (0,2) {\nodepart[text width=0.6cm]{one}$\dequation$\nodepart[text width=0.6cm]{two}$\X_t$};
    \node[draw] (2) at (3.5,2) {\nodepart[text width=0.6cm]{one}$\dequation_\gT$\nodepart[text width=0.6cm]{two}$\Z_t$};
    \node[draw] (3) at (7,2) {\nodepart[text width=0.6cm]{one}$\scm$\nodepart[text width=0.6cm]{two}$ \Z_\infty$};
    \node[draw] (4) at (0,0) {\nodepart[text width=0.6cm]{one}$\dequation^\intervention$\nodepart[text width=0.6cm]{two}$\X_t^\intervention$};
    \node[draw] (5) at (3.5,0) {\nodepart[text width=0.6cm]{one}$\dequation^\intervention_\gT$\nodepart[text width=0.6cm]{two}$ \Z_t^\intervention$};
    \node[draw] (6) at (7,0) {\nodepart[text width=0.6cm]{one}$\scm^\intervention$\nodepart[text width=0.6cm]{two}$ \Z_\infty^\intervention$};

    \draw[|->,shorten >=4pt,shorten <=4pt,color=MidnightBlue] (1) -- (2) node[midway, above] {$\gT$};
    \draw[|->,shorten >=4pt,shorten <=4pt,color=MidnightBlue] (2) -- (3) node[midway, above] {$t \to \infty$};
    \draw[|->,shorten >=4pt,shorten <=4pt,color=BrickRed] (4) -- (5) node[midway, below] {$\gT$};
    \draw[|->,shorten >=4pt,shorten <=4pt,color=BrickRed] (5) -- (6) node[midway, below] {$t \to \infty$};
    \draw[|->,shorten >=4pt,shorten <=4pt,color=BrickRed] (1) -- (4) node[midway, left] {$\intervention$};
    \draw[|->,shorten >=4pt,shorten <=4pt,color=MidnightBlue] (3) -- (6) node[midway, right, xshift=0.33cm] {$\intervention$};

\end{tikzpicture}
\caption{$\gT$-transformation transfers causal information from the temporal to the cross-sectional dimension, and thus to the joint distribution $P(\Z_t)$. The diagram commutes, i.e., red and blue paths produce the same result.}
\label{fig:diagram_z_commutes}
\end{figure}

\paragraph{$\gT$-transformation} To overcome this limitation,  \citep{janzing2018structural} propose a transformation of $\X_t$ based on a frequency analysis of the time series.\footnote{Our $\Z_t$ (\cref{eq:long_run_normalized_mean}) can also be interpreted as a form of discrete Fourier transform of the time-series $\X_{1:t}$ with frequency zero.} Instead, our choice is inspired by the long-run normalized mean via the transformation $\gT: \SP{} \mapsto \SP{}$ defined by
\begin{equation}\label{eq:long_run_normalized_mean}
\gT(\X)_t := \Z_t = \bm{\mu} + \frac{1}{\sqrt{t}}\sum_{i=1}^t (\X_i-\bm{\mu}),
\end{equation}
%
%
where $\bm{\mu} := \expected{\X}$.\footnote{The expectation here is taken over time as well. Nonetheless, for stationary processes, this simplifies to \(\expected{\X} = \expected{\X_t}\) for all \(t\).} 
Moreover, $\expected{\Z_\infty} = \expected{\X_\infty}= \bm{\mu}$ so that for every intervention $\intervention$, $\ce{}^\intervention_\infty$ (\cref{eq:causal_effet_time}) \textit{yields the same values}. However, unlike $\X$, $\Z$ can be mapped into an \SCM{} that precisely models \textit{its distribution shift over any intervention}, thereby satisfying property ii), represented as the commutativity of the diagram in \cref{fig:diagram_z_commutes}. 

It is important to clarify that $\Z_t$ is not the process of interest, and the focus of the causal analysis remains on $\X_t$. However, due to the equivalence of long-run causal effects calculated in both processes, and the ability to associate $\Z_t$ with the \SCM{} that models these effects, $\Z_t$ serves as a convenient intermediate mathematical tool. To demonstrate how this transformation ensures these desirable properties, we will focus on the subclass of linear systems, particularly on \VAR{} models. The reason for this choice is twofold. First, linear models, despite their simplicity, are still on par performance-wise with state-of-the-art Machine-Learning based forecast techniques ~\citep{linear_analysis_toner_2024}, in particular when dealing with stochastic time series ~\citep{parmezan2019evaluation}. Second, the mathematical treatment of interventions and the estimation of causal effects is particularly straightforward to implement and interpret, making this a useful first step for a possible extension to the nonlinear case.


\section{From Vector Autoregressive models to Structural Causal Models} \label{sec:interventions}
In this section, we show that linear \SCM{}s can model the long-term effects of stable \VAR{}s, explaining the properties of its \SP{} equilibrium (\cref{subsec:mapping_var_scm}). 
Then, we provide implementations of two types of causal interventions, leveraging the strengths of the \VAR{} framework (\cref{subsec:causal_interventions}). 
Finally, we discuss the practical implications of our theoretical results 
(\cref{subsec:practical_implications}).

\subsection{Mapping from \VAR{}s to \SCM{}s}\label{subsec:mapping_var_scm}
We provide the explicit mapping from \VAR{}s to linear \SCM{}s in the following theorem (proved in \cref{app:explicit_solution_var_svar}).
\begin{theorem}\label{thm:var_to_scm}
    Given a stable $\VAR{}(p)$ $\dequation$ defined by \cref{eq:VAR}, 
    there exists a linear \SCM{} $\scm$ with structural equations 
    \footnote{For simplicity we set $\bm{\nu}=0$, i.e.,~we assume $\expected{\X_t}=0$. The theorem applies in the general case up to a translation of both the \VAR{} and the associated \SCM{}.} 
    \begin{equation}\label{eq:var_to_scm}
    \begin{split}
        &\tilde{\bm{X}} = \tilde{\A} \tilde{\bm{X}} + \tilde{\vectornoisevar}, 
        \\\text{where }\quad &\tilde{\A} := \left[ \A_1 + \dots + \A_p \right] \text{ and } \tilde{\vectornoisevar} \sim \normal(\bm{0},\boldsymbol{\Sigma}_{\tilde{\vectornoisevar}}),  
    \end{split}
    \end{equation}
    such that, given the transformation $\bm{Z}_t = \frac{1}{\sqrt{t}}\sum_{i=1}^t\X_i$, the following properties hold: 
    \begin{enumerate}
    \item  $\dequation$ and  $\scm$ share the same \emph{causal graph}, i.e., $\graph_{\scm}=\graph_{\dequation}$;\label{samegraph}
    \item The \emph{observational distribution} induced by $\dequation$ at equilibrium $p (\bm{Z}_\infty)$ is equal to the one induced by $\scm$, $p( \tilde{\bm{X}})$;\label{scmissolution}
     \item The \emph{interventional distribution}
     $p (\bm{Z}_\infty^{\intervention})$
     is equal to the one induced by the same intervention on $\scm$,
         $p( \tilde{\bm{X}}^{\intervention_{\scm}})$. \label{scminterventioncomutes}
    %
    \end{enumerate}
\end{theorem}
\begin{remark*}
    Note that due to the influence of time in \VAR{}s, the equivalent  \SCM{}s at equilibrium, while linear, may lead to cycles in the causal graph (see, e.g., \cref{fig:census_graph}) and correlations between the exogenous variables, captured by the full covariance matrix $\boldsymbol{\Sigma}_{\tilde{\vectornoisevar}}$ in \cref{eq:var_to_scm}. Note also that the above Theorem implies that there is a direct relationship between interventions on \SP{}s, $\intervention$, and interventions on \SCM{}s, here denoted by $\intervention_{\scm}$. Refer to \cref{app:explicit_solution_var_svar} for further details. 
\end{remark*}


\subsection{Implementation of Causal Interventions}\label{subsec:causal_interventions}

Different application scenarios may need different types of interventions. Consider a government's fiscal policy. 
In such a setting, a feasible approach would be to implement an \emph{additive intervention} in the form of an annual tax increase of, e.g., $300$ euros per household on top of existing taxes. 
Alternatively, in other scenarios, e.g., when studying the effect of the \textit{key European Central Bank's interest rate}~\citep{belke2007ecb}, a more natural choice is to implement a \emph{forcing intervention} that enforces the convergence of an observed variable (e.g., interest rate) to a target value. 
In the following, we propose an implementation for \VAR{}s of these two forms of interventions, 
showing their effects on the system and discussing their stability conditions.

\subsubsection{Additive Interventions}\label{subsec:forcing}

Given a stable \VAR{}(\order) as in \cref{eq:var_lagoperator}, we define an additive intervention $\intervention_{a}$ at time $t_\intervention$ with force $\bm{F}$ as the mapping:
\begin{equation}\label{eq:forcing_intervention}
\begin{split}
   \intervention_{a}: \,  &\A(L)\X_t = \bm{\nu} + \vectornoisevar_t \quad \longmapsto \\
    &\A(L)\X_t = \bm{\nu} + \vectornoisevar_t + \mathds{I}(t\geq t_\intervention)\bm{F},
    \end{split}
\end{equation}
where $\mathds{I}(t\geq t_\intervention)$ is the indicator function, which equals $1$ if $t\geq t_\intervention$, otherwise $0$. 
In other words, we perform a translation while keeping the process dynamics unchanged.
%
In such case, the temporal causal effect $\ce{}_t$ is deterministic and takes values $\ce_{t}=0$  for $t <t_\intervention$ while, for 
$k\geq 0$:
\begin{equation*}
    \ce_{t_\intervention+k} 
    = \sum_{l=0}^k \response_l \bm{F}, 
\end{equation*}
where $\Phi$ is the \emph{impulse response function} of the \VAR{} model. Refer to \cref{app:svarproperties} for further details.
%
%
\begin{remark*}
    For this type of intervention, $\ce_t$ is deterministic and does not depend on the specific trajectory. The same property can be observed on the linear $\SCM{}$ associated with the process, defining the intervention in a similar way: $\tilde{\X}=\tilde{\A} \tilde{\X}+\tilde{\vectornoisevar}$ changes into $\tilde{\X}=\tilde{\A} \tilde{\X}+\bm{F}+\tilde{\vectornoisevar}$.
\end{remark*}

\paragraph{Stability}
Additive intervention preserves the stability regardless of the value of $\bm{F}$, since $\A(L)$ does not change. See \cref{app:svarproperties} for stability conditions of \VAR{}s.

\subsubsection{Forcing Interventions}\label{subsec:do-like}

We define a forcing intervention $\intervention_f$ at time $t_\intervention$ with force $\bm{F}$ and target value $\hat{\X}$ as:
\begin{equation}\label{eq:do-like}
\begin{split}
    \intervention_{f}&: \, \A(L)\X_t = \bm{\nu} + \vectornoisevar_t \quad \longmapsto \\
    &\A(L)\X_t = \bm{\nu} + \vectornoisevar_t + \mathds{I}(t\geq t_\intervention)\bm{F}\odot(\hat{\X}-\X_t).
\end{split}
\end{equation}
We assume $\bm{F}$ to be positive in each component. 
This intervention acts as an attraction towards $\hat{\X}$, and $\bm{F}$ modulates the intensity of the attraction force. Applying an intervention on a single component $\Xcomp[i]$ toward the fixed value $\hat{X}$ and letting $F^{(i)}\to +\infty$ yields the do operator $do(\Xcomp[i] = \hat{X})$. We refer to~\citep{Mooij2013deterministic_case} for a detailed discussion of this point.

\paragraph{Stability}
Forcing interventions $\intervention_{f}$ perturb the system dynamics by modifying the operator $\A(L)$. 
Specifically, by shifting the term $\bm{F} \odot \X_t$ to the left of the equation and rewriting it in matrix form as $\bm{F}_{diag}\X_t$, we obtain $\tilde{\A}(L):=\A(L)+\bm{F}_{diag}$. 
Hence, the stability of the intervened system is not guaranteed (we provide an example in \cref{subsec:stability_not_granted}), and it is necessary to verify \textit{that all the eigenvalues of} $\tilde{\A}(L)$ \textit{are still inside the unit circle.} Intuitively, the stability of an observational system often relies on negative feedback loops. Fixing one variable can disrupt this balance, leading to runaway behavior. For example, turning off a pressure release valve in a pressurized tank can cause the pressure to build up uncontrollably, eventually leading to an explosion.


\subsection{Practical implications}\label{subsec:practical_implications}

\paragraph{Causal queries}
Our formulation of causal interventions on \VAR{}s differs from the standard approach based on Granger causality by being closer 
to that of \SCM{}s. 
Consequently, it enables the generalization of interventional and counterfactual queries 
to account for time (see~\cref{app:queries_practice}).
That is, it allows for answering the following causal questions: 

\begin{itemize}
    \item \textbf{Forecasted Interventions} What are the expected effects on an individual trajectory (or a population) when intervening in the present, and how do they vary over time?
    
    \item \textbf{Retrospective Counterfactuals} What would have happened to an individual trajectory if an intervention had been applied at a specific point in the past? What state would it be in now?
\end{itemize}
Both causal queries acquire a meaning embedded in the temporal dimension in terms of \textit{forecasting for the future} (\cref{subsec:interventional_forecasting}) and \textit{retrospection for the past} 
 (\cref{app:counterfactuals}), respectively. 

\paragraph{Expressiveness and universality}
\VAR{}s, despite their linearity, possess a high level of expressiveness~\citep{kilian2017structural}. 
In fact, the Wold decomposition Theorem \citep{wold1938study} implies that the dynamics of \textit{any purely nondeterministic covariance-stationary process can be approximated arbitrarily well by an autoregressive model, making them universal approximators}. In practice, linear autoregressive models are broadly used in time-series analysis. Yet, we intend to explore non-linear \SP{}s in future work, as they may lead to better convergence rates and allow for 
causal interpretation of a broader family of dynamical models.

\paragraph{Feedback loops}
To properly understand complex systems, it is often useful to model feedback loops between their variables. 
Time-series models naturally capture this property, while \SCM{}s require significant reformulation. 
The theory of cyclic \SCM{}s has seen a significant advancement in recent years \citep{bongers2021foundations}, but
practical approaches, both for causal discovery and causal inference, are still underdeveloped~\citep{bongers2016cyclic, Lorbeer2023cycliccausaldiscovery}. 
%
%
Our formalization of causal inference on \VAR{}s is a step forward in this direction.

\paragraph{Fitting}
\VAR{}s estimation is typically performed using ordinary least squares. Various alternative methods are available, both in terms of constrained optimization (e.g., to use prior knowledge about some coefficients of the \VAR{} matrices ~\citep{sims1980macroeconomics}) and within a Bayesian framework ~\citep{koop2010bayesian}. Refer to ~\citep[chapters~3,4,5]{lutkepohl2005new} for a comprehensive discussion. 
%
Importantly, although 
\VAR{}s are most commonly used on
time-series data (i.e., data from one single unit across a period of time), there are approaches tailored to the analysis of \emph{panel data} (i.e., the evolution of many units over time) ~\citep{sigmund2021panel}; and \emph{cross-sectional data} (i.e.,~many individuals at a single point of time), provided that they have at least some proxy variables of time ~\citep{deaton1985panel}. 
%
Such approaches open up a promising line of future work that can further generalize \VAR{}s applicability for causal reasoning over time. 

\section{Empirical evaluation} \label{sec:experiments}
In this section, we evaluate \VAR{} models' accuracy and expressiveness in multivariate time series, focusing on two forecasting dimensions: observational (\cref{subsec:observational_forecasting}) and interventional (\cref{subsec:interventional_forecasting}).
Additional results and in-depth descriptions can be found in~\cref{app:extra_results}. 

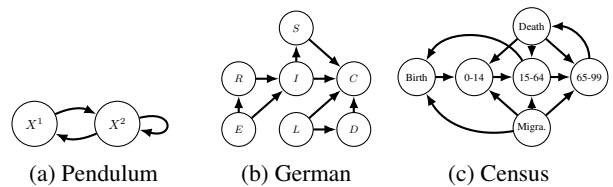
\begin{figure}[t]
    \centering
    \begin{subfigure}[b]{0.32\columnwidth}
        \centering
        \begin{tikzpicture}[
            scale=0.5,
            transform shape,
            vertex/.style={circle, draw, minimum size=1.2cm, inner sep=0pt, font=\normalsize}
        ]
        \node[vertex] (x1) at (0, 1.5) {$X^{1}$};
        \node[vertex] (x2) at (2.2, 1.5) {$X^{2}$};
        
        \draw[edge] (x1) to [bend left=25] (x2);
        \draw[edge] (x2) to [bend left=25] (x1);
        \draw[edge] (x2) to [loop right] (x2);
        \end{tikzpicture}
        \caption{\footnotesize Pendulum}
        \label{fig:inverted_pendulum_graph}
    \end{subfigure}%
    \begin{subfigure}[b]{0.32\columnwidth}
        \centering
        \begin{tikzpicture}[
            scale=0.45,
            transform shape,
            vertex/.style={circle, draw, minimum size=1cm, inner sep=0pt, font=\footnotesize}
        ]
        \node[vertex] (r) at (0,1.5) {$R$};
        \node[vertex] (e) at (0,0) {$E$};
        \node[vertex] (l) at (1.7,0) {$L$};
        \node[vertex] (d) at (3.4,0) {$D$};
        \node[vertex] (i) at (1.7,1.5) {$I$};
        \node[vertex] (s) at (1.7,3) {$S$};
        \node[vertex] (y) at (3.4,1.5) {$C$};
        
        \draw[edge] (e) -- (r);
        \draw[edge] (l) -- (d);
        \draw[edge] (r) -- (i);
        \draw[edge] (e) -- (i);
        \draw[edge] (i) -- (s);
        \draw[edge] (l) -- (y);
        \draw[edge] (d) -- (y);
        \draw[edge] (i) -- (y);
        \draw[edge] (s) -- (y);
        \end{tikzpicture}
        \caption{\footnotesize German}
        \label{fig:german_graph}
    \end{subfigure}%
    \begin{subfigure}[b]{0.32\columnwidth}
        \centering
        \begin{tikzpicture}[
            scale=0.45,
            transform shape,
            vertex/.style={circle, draw, minimum size=1.1cm, inner sep=0pt, font=\small}
        ]
        \node[vertex] (b) at (0, 1.5) {Birth};
        \node[vertex] (d) at (3.4, 3) {Death};
        \node[vertex] (m) at (3.4,0) {Migra.};
        \node[vertex] (a-0) at (1.7,1.5) {0-14};
        \node[vertex] (a-1) at (3.4,1.5) {15-64};
        \node[vertex] (a-2) at (5.1,1.5) {65-99};
        
        \draw[edge] (b) -- (a-0);
        \draw[edge] (a-1) to [out=120,in=60](b); 
        \draw[edge] (d) -- (a-2);
        \draw[edge] (d) -- (a-1);
        \draw[edge] (d) -- (a-0);
        \draw[edge] (m) -- (a-0);
        \draw[edge] (m) -- (a-1);
        \draw[edge] (m) -- (a-2);
        \draw[edge] (a-0) -- (a-1);
        \draw[edge] (a-1) -- (a-2);
        \draw[edge] (m) to [out=190,in=-60](b);
        \draw[edge] (a-2) to [out=90, in=0](d);
        \end{tikzpicture}
        \caption{\footnotesize Census}
        \label{fig:census_graph}
    \end{subfigure}
    \caption{\textbf{Causal graphs}. The causal graph for (a) and (b) is known, while for (c), it is assumed.
    In (b), nodes are labeled with the initials of each feature: Expertise, Responsibility, Loan Amount, Duration, Income, Savings, and Credit Score. In (c), $0-14$, $15-64$, and $65-99$ represent age groups.}
    \label{fig:datasets_graphs}
    \vspace{-0.2cm}
\end{figure}

\paragraph{Datasets} We rely on two synthetic datasets, German\footnote{This dataset is inspired on \url{https://archive.ics.uci.edu/dataset/144/statlog+german+credit+data}} and Pendulum,
and the real-world Census dataset\footnote{\url{https://data.census.gov/}}.
German simulates a loan approval scenario with seven variables.
%
Pendulum is a two-variable system where
$X^{(1)}$ operates as a stabilizer for $X^{(2)}$, which exhibits a divergent dynamic. 
%
Census includes demographic variables across three age groups, along with migration, birth, and death rates from $1992$ to $2023$ for $50$ countries. 
\cref{fig:datasets_graphs} illustrates the causal graphs for all datasets.
%
%
See~\cref{app:dataset} for further details.

\paragraph{Metrics}
We measure the discrepancy between the \textit{h}-steps forecast $\hat{\X}_{t+h}|\X_{<t}$ and the true value $\X_{t+h}$ on the test set $\X_{test}$. 
%
We report Mean Absolute Error (MAE) focusing on the target variables (i.e., \textit{Credit Score} for German, $X^{(1)}$ for Pendulum, and age groups for Census). See~\cref{app:obs_forecasting} for other metrics. All results are averaged over ten runs.

\begin{figure*}[t!]
  \centering
  \includegraphics[width=0.9\textwidth]{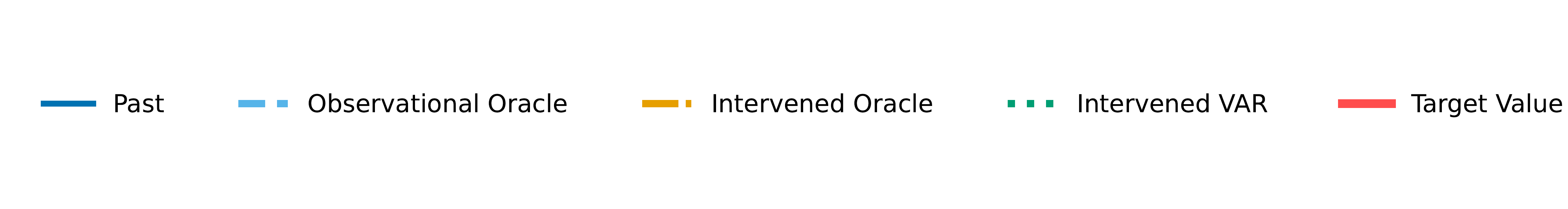}
  \begin{minipage}[t]{0.49\textwidth}
    \centering
    \begin{minipage}{\textwidth}
    \vspace{-0.7cm} 
    \begin{subfigure}[b]{0.49\textwidth}
      \centering
      \includegraphics[width=\textwidth]{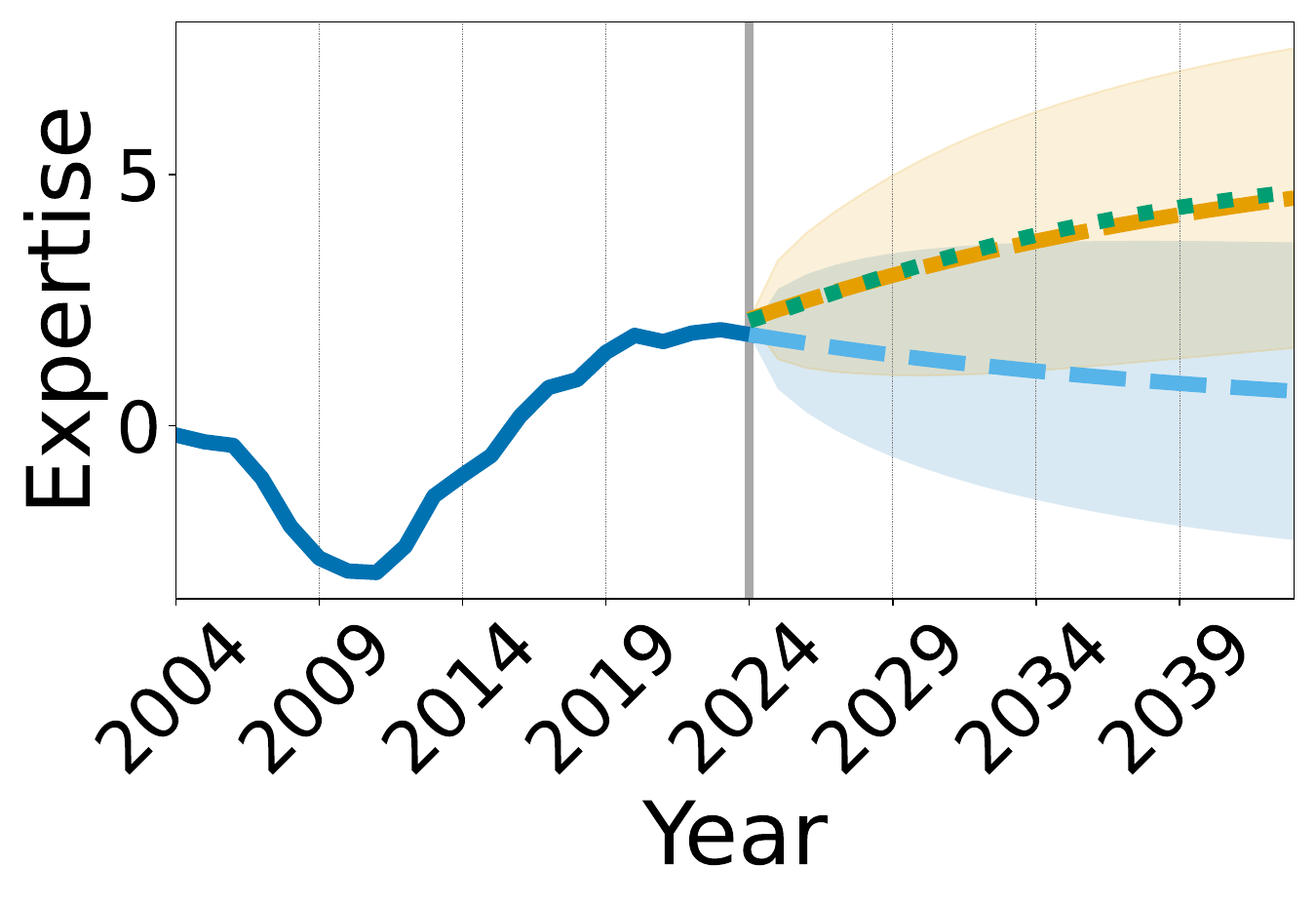}
    \end{subfigure}
    \begin{subfigure}[b]{0.49\textwidth}
      \centering
      \includegraphics[width=\textwidth]{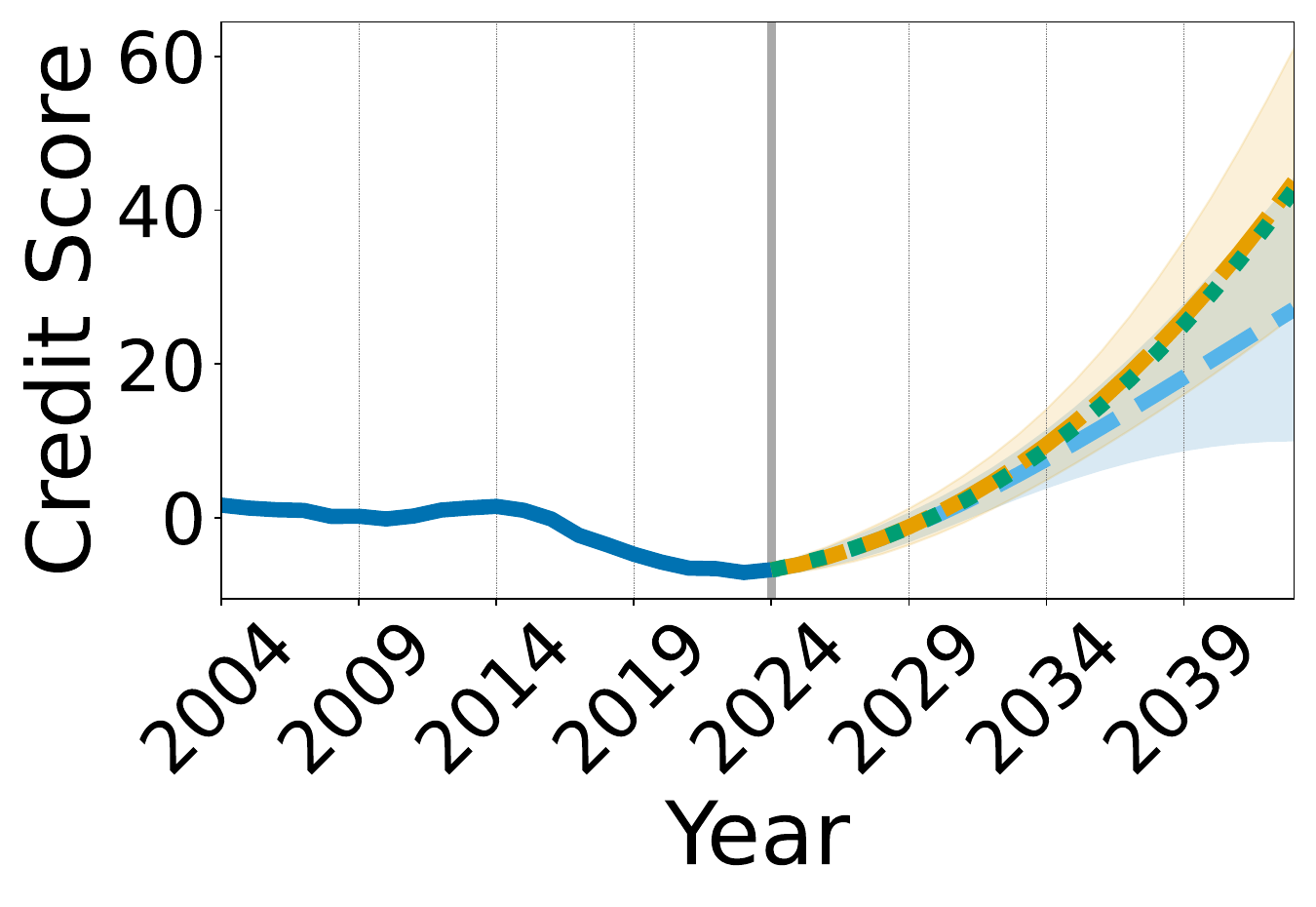}
    \end{subfigure}
    \end{minipage}
    \caption{\textbf{Additive Intervention.} (Left) Intevention on \textit{Expertise} with $\bm{F} = 0.2$. 
    (Right) Effect on \textit{Credit Score}. Shaded regions in both plots denote $95\%$ confidence bounds.}
    \label{fig:additive_int}
  \end{minipage}
  \hspace{0.15cm}
  \begin{minipage}[t]{0.49\textwidth}
    \centering
    \begin{minipage}{\textwidth}
    \vspace{-0.7cm} 
    \begin{subfigure}[b]{0.49\textwidth}
      \centering
      \includegraphics[width=\textwidth]{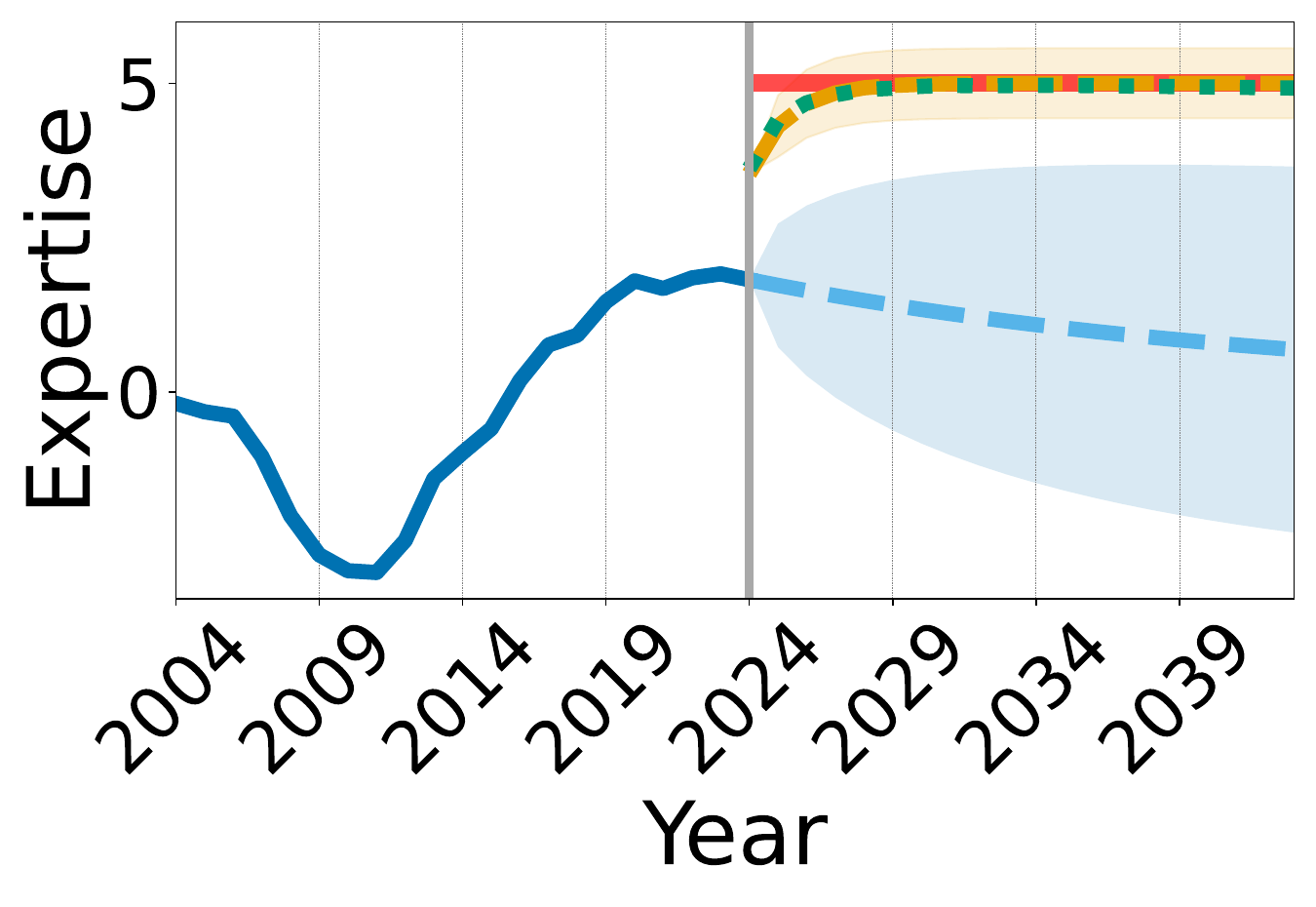}
    \end{subfigure}
    \hfill
    \begin{subfigure}[b]{0.49\textwidth}
      \centering
      \includegraphics[width=\textwidth]{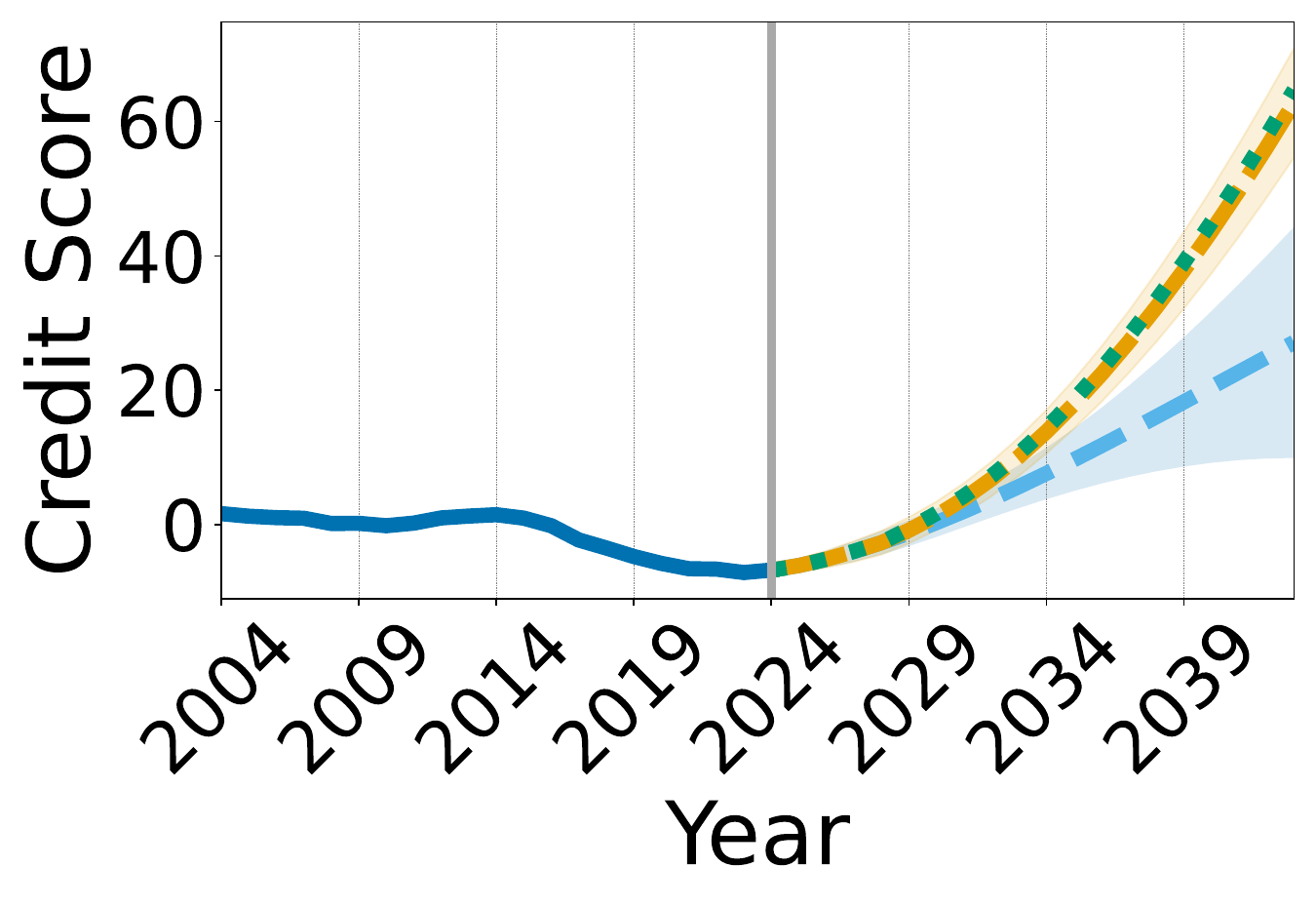}
    \end{subfigure}
     \end{minipage}
    \caption{\textbf{Forcing Intervention.} (Left) Intevention on \textit{Expertise} with $\bm{F} = 1$ and target $\hat{E}=5$. (Right) Effect on \textit{Credit Score}. Note that confidence bounds are narrower w.r.t.~\cref{fig:additive_int}.}
    \label{fig:forcing_int}
  \end{minipage} 
\end{figure*}

\subsection{Observational Forecasting}\label{subsec:observational_forecasting}
\paragraph{Baselines}
We compare \VAR{} with three relevant works: i) DLinear~\cite{zeng2023dlinear}, a decomposition-based linear model that separates trend and seasonal components;
ii) TSMixer~\cite{chen2023tsmixer}, a Multi-layer Perceptron (MLP) based model that focuses on mixing time and feature dimensions; iii) TiDE~\cite{das2023tide}, a MLP based encoder-decoder model. 
To assess the effectiveness of the forecasting methods, we introduce an observational oracle forecaster that has full knowledge about the true data generating process and produces the optimal predictor, i.e., $\hat{\X}_{t+h}|\X_{<t} = \expected{\X_{t+h}|\X_{<t}}$. 


\begin{table}[t]
    \centering
    \caption{\textbf{Observational Forecasting.}
    MAE scores (\emph{lower is better}) for 
    \VAR{}, DLinear~\citep{zeng2023dlinear},  TiDE~\cite{das2023tide} and TSMixer~\cite{chen2023tsmixer}.   
    Results averaged over $10$ runs. Due to space limitations, standard deviations are reported in~\cref{app:obs_forecasting}. 
    Best model in bold, Oracle in $\mathtt{typewriter}$.
    For Census, size equals the number of countries times the number of years.}
    \label{tab:comparative_forecasting}
    \setlength{\tabcolsep}{1.5pt}
    \footnotesize
    \renewcommand{\arraystretch}{1.18}
    \begin{tabular}{lcccccccc}
    \toprule
    & & & \multicolumn{5}{c}{Observational Forecasting} \\
    \cmidrule(lr){4-8}
    Dataset & Size & Horizon & Oracle & VAR & DLinear & TiDE & TSMixer \\
\midrule
    German & \multirow{4}{*}{\centering100} & \multirow{2}{*}{\centering1} & $\mathtt{.004}$ & $\mathbf{.008}$ & $.009$ & $.011$ & $.014$ \\
Pendulum & & & $\mathtt{.042}$ & $\mathbf{.043}$ & $\mathbf{.043}$ & $.218$ & $.217$ \\
\cmidrule(lr){3-8}
German & & \multirow{2}{*}{\centering10} & $\mathtt{.014}$ & $\mathbf{.055}$ & $\mathbf{.055}$ & $.094$ & $.139$ \\
Pendulum & & & $\mathtt{.399}$ & $\mathbf{.420}$ & $.440$ & $1.43$ & $1.43$ \\
\midrule
German & \multirow{4}{*}{\centering500} & \multirow{2}{*}{\centering1} & $\mathtt{.004}$ & $\mathbf{.004}$ & $\mathbf{.004}$ & $.011$ & $.014$ \\
Pendulum & & & $\mathtt{.042}$ & $\mathbf{.042}$ & $\mathbf{.042}$ & $.218$ & $.217$ \\
\cmidrule(lr){3-8}
German & & \multirow{2}{*}{\centering10} & $\mathtt{.014}$ & $\mathbf{.015}$ & $\mathbf{.015}$ & $.093$ & $.135$ \\
Pendulum & & & $\mathtt{.399}$ & $\mathbf{.401}$ & $.405$ & $1.43$ & $1.43$ \\
\hline \hline
\multirow{2}{*}{Census} &\multirow{2}{*} {\centering{$50\times32$}} & 1 & - & $\mathbf{.001}$ & $.006$ & $\mathbf{.001}$ & $.008$ \\
& & 5 & - & $.017$ & $.025$ & $\mathbf{.014}$ & $.024$\\
\bottomrule
    \end{tabular}
\end{table}

\begin{table}[ht]
    \centering
    \caption{\textbf{Interventional Forecasting.} 
    MAE scores (\emph{lower is better}) for the proposed causal \VAR{} framework on the German dataset. Results averaged over $10$ runs, with standard deviation in subscript. Scores are scaled by a factor of $10^2$ to ease readability.}
    \label{tab:interventional_forecasting_german}
    \setlength{\tabcolsep}{4pt}
    \footnotesize
    \renewcommand{\arraystretch}{1.05}
    \begin{tabular}{ccccc}
    \toprule
    & & & \multicolumn{2}{c}{Interventional Forecasting} \\
    \cmidrule(lr){4-5}
    Dataset & Size & Horizon & Additive & Forcing \\
    \midrule
    \multirow{4}{*}{German} & \multirow{2}{*}{100} & 1 & $.000_{.000}$ & $.000_{.000}$ \\
& & 10 & $.043_{.028}$ & $.364_{.297}$ \\
\cmidrule(lr){2-5}
 & \multirow{2}{*}{500} & 1 & $.000_{.000}$ & $.000_{.000}$ \\
& & 10 & $.018_{.014}$ & $.115_{.081}$ \\
\bottomrule
\end{tabular}
\end{table}

\paragraph{How does the \VAR{} performance compare with SOTA models for forecasting multivariate time series?}
The observational forecasting results in~\cref{tab:comparative_forecasting} show performance across varying data sizes (i.e., number of instances) and forecast horizons for all datasets. 
\VAR{} emerges as the top-performing model, consistently matching or closely approaching Oracle's scores for all datasets. 
DLinear usually achieves predictive accuracy close to \VAR{} for $1$-step forecasts, presumably due to the common linear nature of both models. 
TiDE and TSMixer consistently underperform compared to \VAR{} and DLinear for German and Pendulum. 
For all models (including Oracle), performance on the Pendulum dataset is uniformly worse than on the German, highlighting the greater challenge in forecasting given the system's stronger stochasticity and variables changing more rapidly over time.
On Census, \VAR {} and TiDE provide the best results,  TiDE slightly outperforming \VAR{} in $5$-step horizon.

\subsection{Interventional Forecasting}\label{subsec:interventional_forecasting}
We evaluate the causal \VAR{}'s forecast in estimating the causal effects on German. See~\cref{app:int_forecasting} for other datasets.
%

\paragraph{Baselines} 
Since state-of-the-art methods do not allow computing the causal effect of interventions on dynamical systems, we use an oracle forecaster 
as a benchmark for theoretically optimal performance.
Specifically, the ground truth values are estimated as $\ce_{t+h} = \expected{\X^\intervention_{t+h} - \X_{t+h}|\X_{<t}}$, while the predicted values from the proposed \VAR{} framework as $\hat{\ce}_{t+h} = (\hat{\X}^\intervention_t - \hat{\X}_{t+h})|\X_{<t}$.

\paragraph{Interventions} 
We perform causal interventions on the root node \textit{Expertise} and observe the effect on the target variable \textit{Credit Score}.
For the additive case, we apply $\bm{F}=0.2$, while for forcing, we use $\bm{F}=1$ with a target value of $\hat{E} = 5$. These values are selected for illustrative purposes such that the long-term expected value of \textit{Expertise} is the same for both interventions (i.e., $5$). See~\cref{app:int_forecasting} for other variants.

\begin{figure*}[t!]
  \centering
  \includegraphics[width=\textwidth]{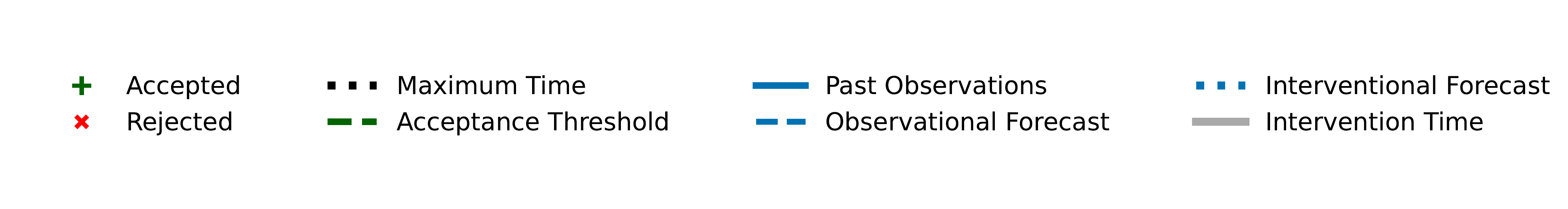}
  \begin{minipage}[t]{0.49\textwidth}
    \centering
    \begin{minipage}{\textwidth}
    \vspace{-0.6cm}  
      \begin{subfigure}[b]{0.49\textwidth}
        \centering
        \includegraphics[width=\textwidth]{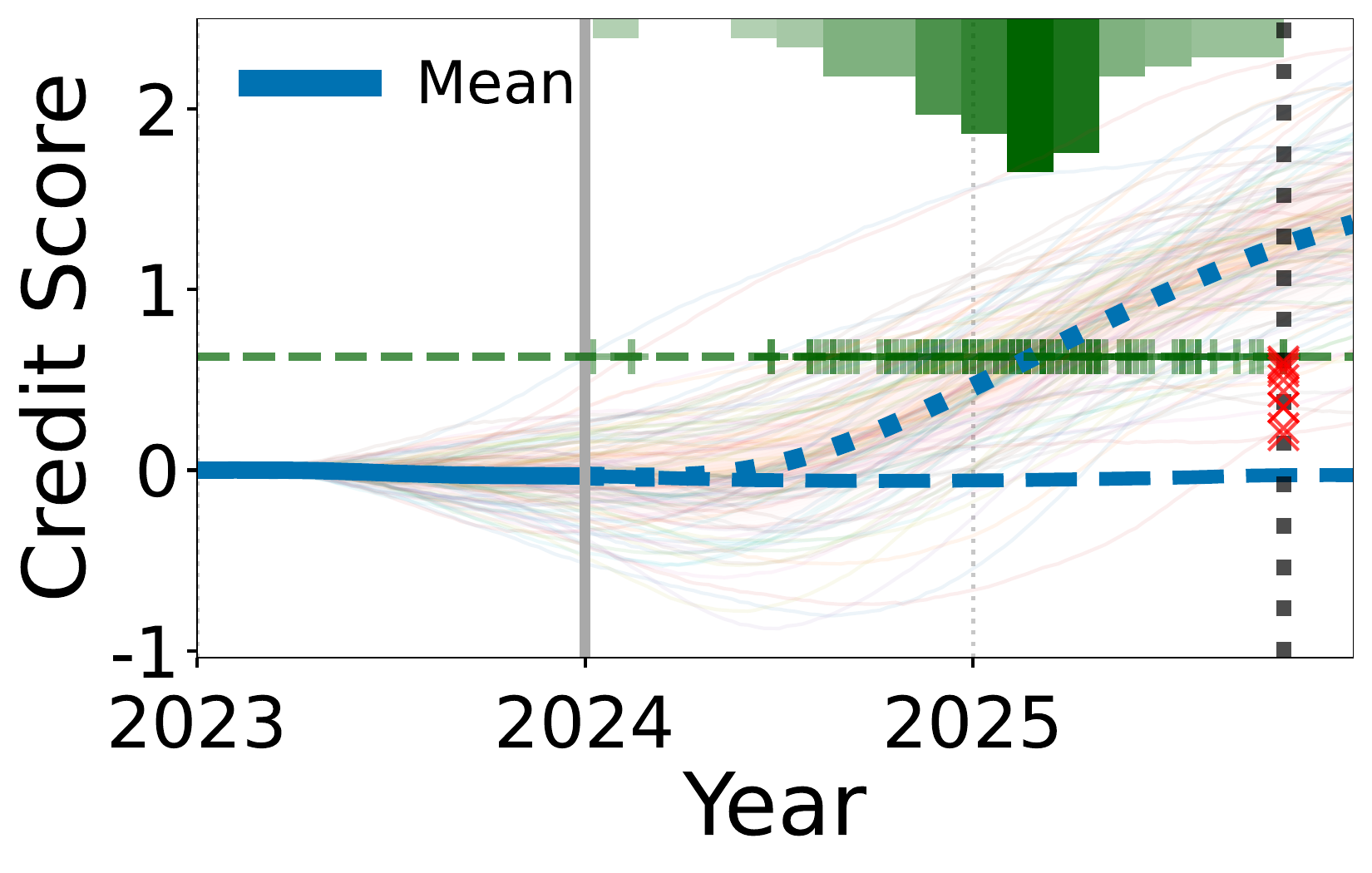}
      \end{subfigure}\hspace{-0.12cm}
      \begin{subfigure}[b]{0.49\textwidth}
        \centering
        \includegraphics[width=\textwidth]{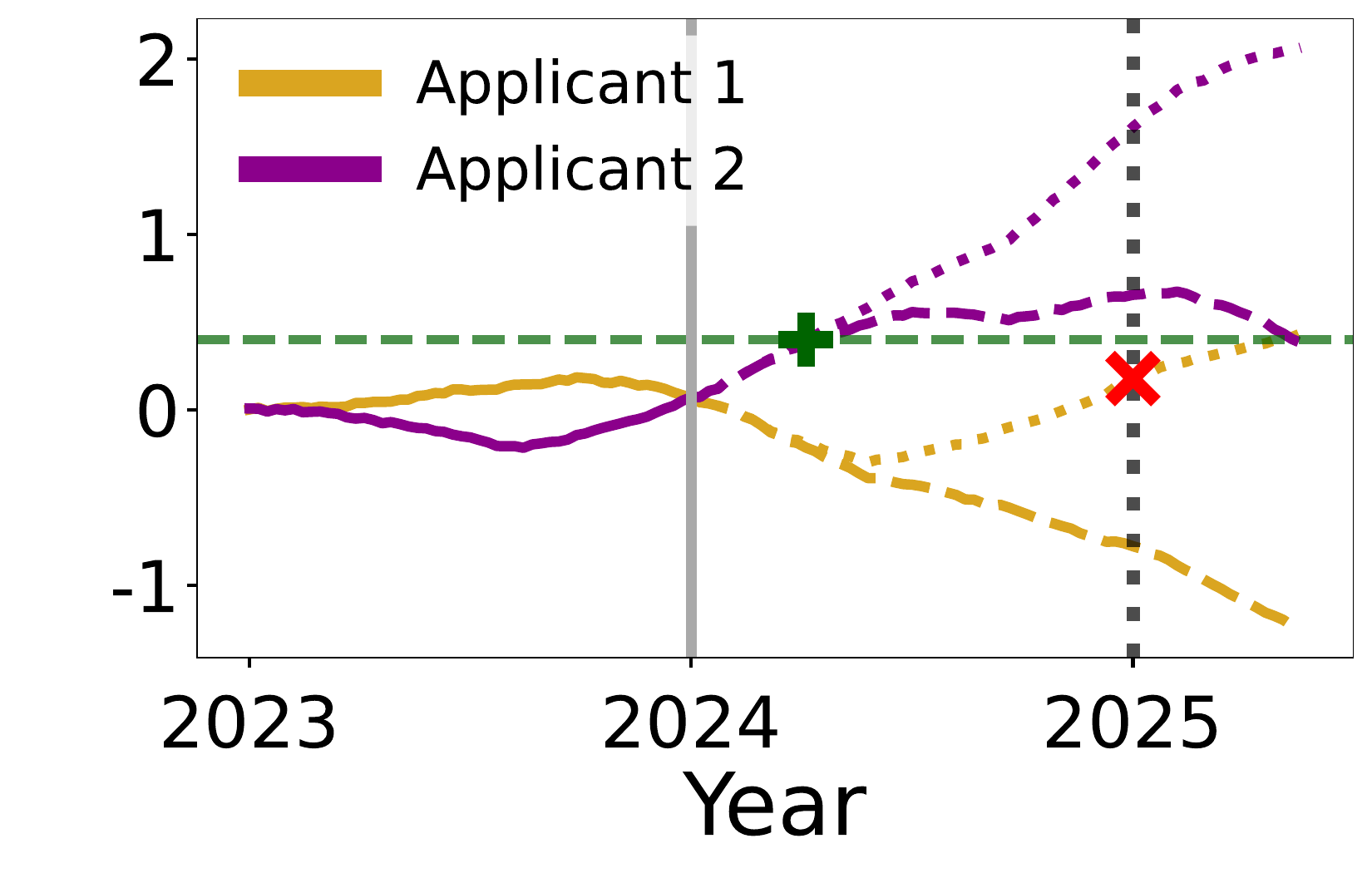}
      \end{subfigure}
    \end{minipage}
    \caption{\textbf{German}. Effect of increasing \textit{Expertise} on \textit{Credit Score}. ({Left}) The time each loan applicant takes to cross or not the acceptance threshold. The histogram shows the distribution of crossing times. 
    ({Right}) Comparison of two loan applicants, i.e., trajectories, with similar scores at intervention time. After the intervention, they diverge significantly, with only an applicant being accepted at the maximum time. Forecasts are dashed for observational and dotted for interventional.}
    \label{fig:use_case_german}
  \end{minipage}
  \hfill
  \begin{minipage}[t]{0.49\textwidth}
    \centering
    \begin{minipage}{\textwidth}
    \vspace{-0.6cm}  
      \begin{subfigure}[b]{0.49\textwidth}
        \centering
        \includegraphics[width=\textwidth]{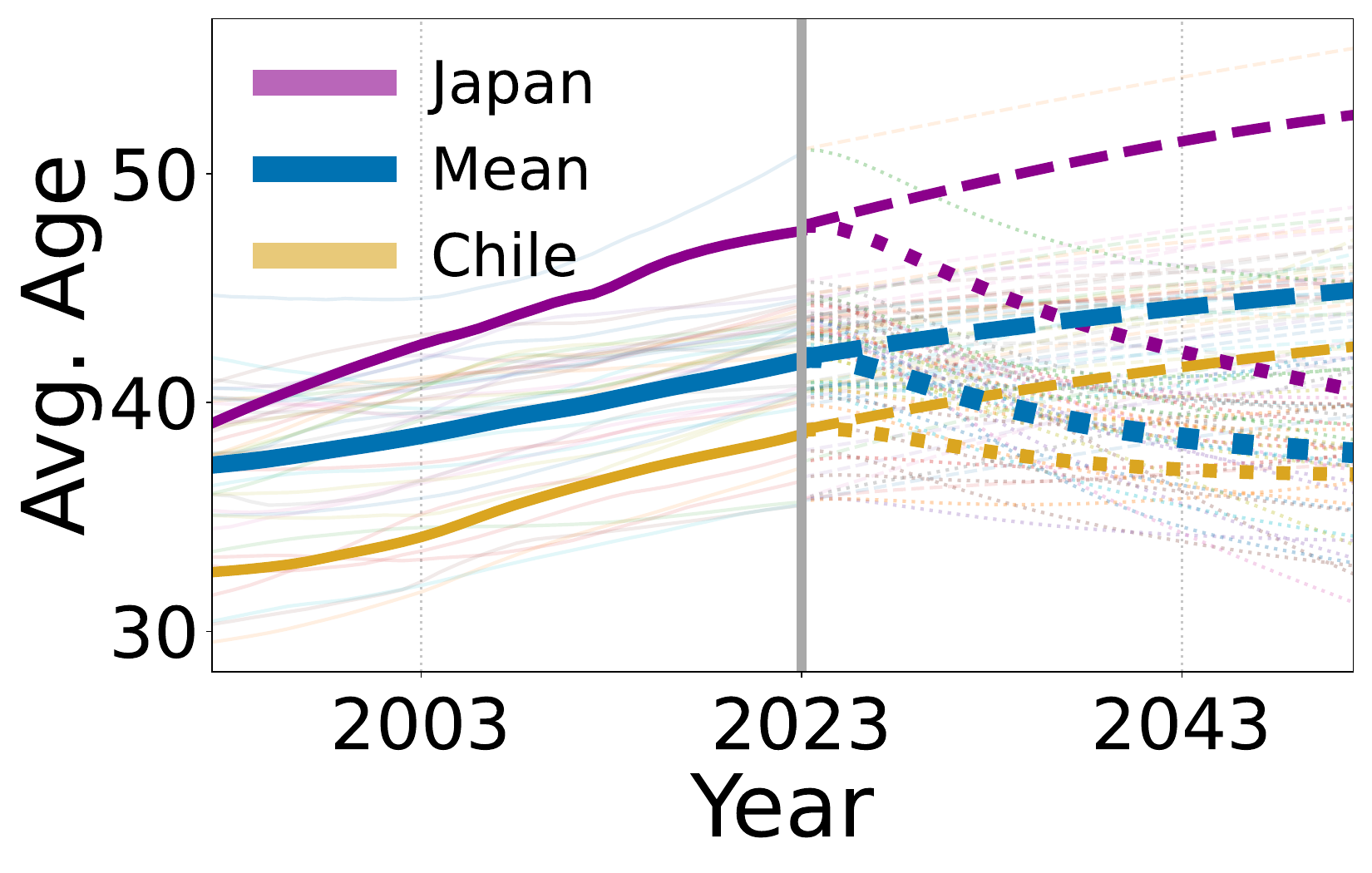}
      \end{subfigure}
      \begin{subfigure}[b]{0.49\textwidth}
        \centering
        \includegraphics[width=\textwidth]{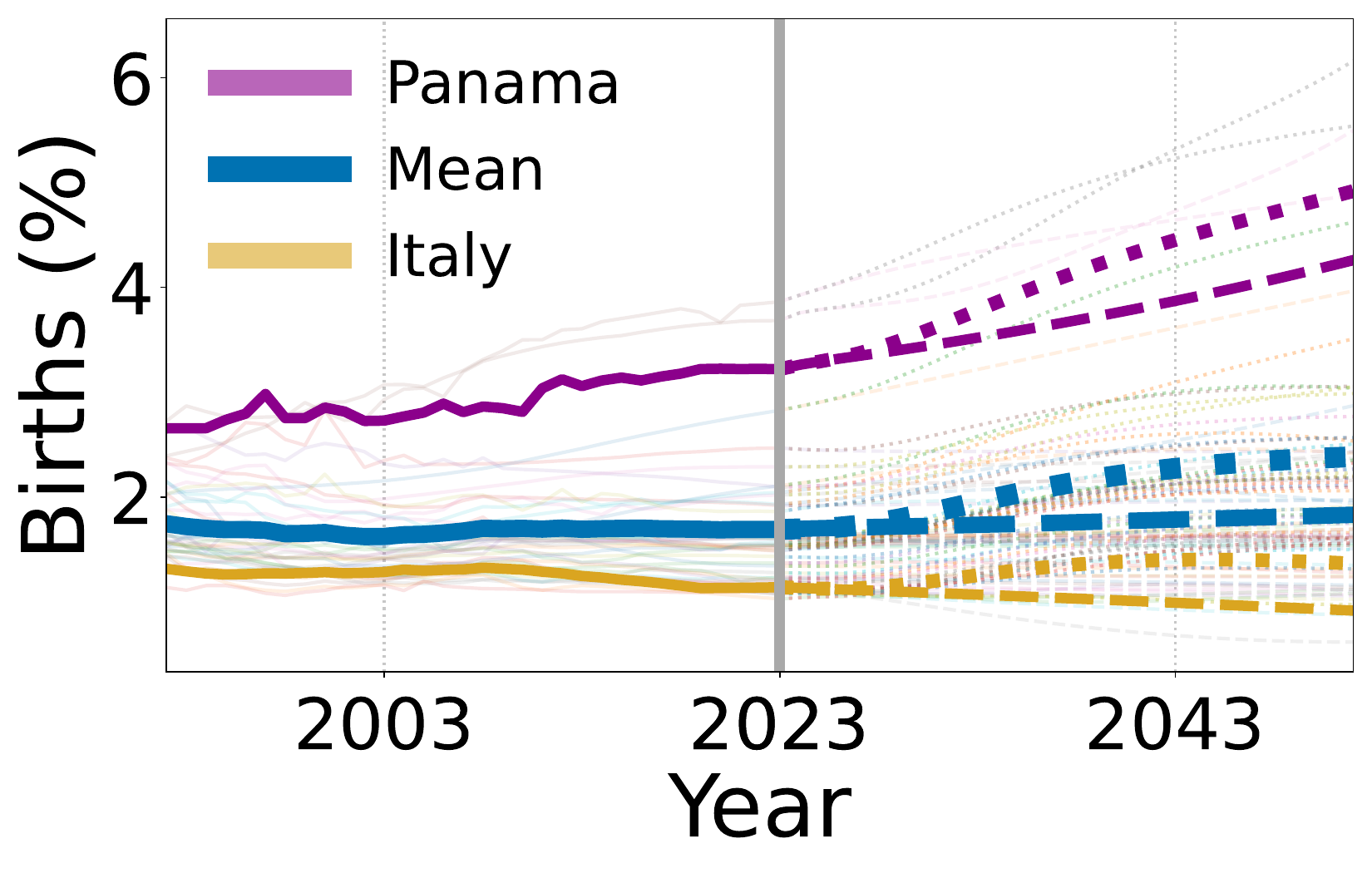}
      \end{subfigure}
    \end{minipage}
    \caption{\textbf{Census}. Additive intervention across all countries.
    (Left) Intervention on \textit{Births} with $\bm{F}=0.004$ and its effect on the population's average age.
    (Right) Intervention on \textit{Migration} with $\bm{F}=0.04$ and its effect on \textit{Births}. 
    In both plots, we highlight two countries (one above and one below the mean) to illustrate the differences after the intervention. Forecasts are dashed for observational and dotted for interventional.
    }
    \label{fig:use_case_census}
  \end{minipage}
\end{figure*}

\paragraph{How do additive and forcing interventions affect the system dynamics? }
\cref{fig:additive_int} illustrates the additive intervention.
\textit{Expertise} is a variable that typically necessitates several years for acquisition in practical scenarios. 
The causal \VAR{} accurately captures such delayed impact as its effect on \textit{Credit Score} appears after several years.
As the system maintains its dynamic characteristics unchanged, the forecasted covariance remains the same even after the intervention.
\cref{fig:forcing_int} shows the forcing intervention, 
 where the interventional forecasting exactly aligns with the target value. 
Moreover, we stress that even for a low value of $\bm{F}$, the forcing intervention resembles a do-intervention (shrinking the variances significantly) even though theoretically convergence is guaranteed only for $F \to \infty$.

\paragraph{How accurate is the causal VAR framework in estimating the causal effect of interventions over time?} 
\cref{tab:interventional_forecasting_german} summarizes results on interventional forecasting, showing  errors with varying data size 
and forecast horizons.  
At $1$-step, both interventions lead to perfect performance since \textit{Credit Score} is a slow-changing variable and requires at least $3$ time steps for an intervention to take effect.
At $10$-step, our causal effect estimates remain highly accurate.


\section{Use cases} \label{sec:use-cases}
In this section, we show real-world scenarios where estimating causal effects over time represents a useful step toward a 
realistic modeling of the phenomena.
We focus on the German and Census datasets, presenting two analyses for each.
For German, \cref{fig:use_case_german} illustrates the effect of increasing \textit{Expertise} by $\bm{F} = 0.38$ on \textit{Credit Score}. 

\textbf{German 1 -- Same intervention, different outcomes. }
The left panel of \cref{fig:use_case_german} shows the distribution of the time required for loan applicants to cross or not the acceptance threshold after the intervention. 
Access to this information allows quantification of intervention efficacy, identification of credit-building patterns, and infer the key factors influencing loan eligibility outcomes.
It can also inform the recommendation of actions (\eg in algorithmic recourse~\citep{karimi2022survey}) 
within a reasonable timeframe, fostering trust in the system and promoting user acceptance.

\textbf{German 2 -- Similar cross-sectional values, different causal effects over time. }
The right panel of \cref{fig:use_case_german} shows trajectories that, while seeming similar at a given time, may have significantly divergent historical and future behaviors. For instance, the purple trajectory may have autonomously crossed the threshold without intervention, whereas in the case of the yellow one, the applied intervention may be inadequate to ensure the desired outcome. 
Such divergence highlights the importance of moving beyond models that rely only on cross-sectional data, motivating the need for techniques, such as the proposed causal \VAR{} framework, that 
capture individual applicant behavior over time.

For Census,~\cref{fig:use_case_census} presents two additive interventions across all countries. 

\textbf{Census 1 -- Impact of Births on Avg. Age. } 
The left panel of ~\cref{fig:use_case_census} shows the increase on \textit{Births} with $\bm{F} = 0.004$ and its effect on
the population’s average age (computed as a weighted mean of age groups). The force value means that \textit{Births} increase by $0.4\%$ of each country's total population every year.
Examining how they affect population age over time could allow policymakers to identify which countries might benefit most from specific types of demographic interventions and evaluate the long-term viability of systems. 
We can also observe that the intervention in Japan causes a more evident decrease in the average age than in Chile.

\textbf{Census 2 -- Impact of Migration on Births. } 
The right panel of~\cref{fig:use_case_census} reports how a $4\%$ growth in \textit{Migration} w.r.t. the total population influences \textit{Birth} rates. 
We observe that increased migration leads to a rise in births. However, its impact is less evident (observational and interventional forecasting trajectories are closer) compared to the result on the average age shown in~\cref{fig:use_case_census} (left panel).

\section{Concluding remarks} \label{sec:conclusions}

In this work, we have established a link between discrete-time dynamical systems at equilibrium and \SCM{}s. 
Moreover, we have provided an explicit procedure for mapping \VAR{}s to linear \SCM{}s and demonstrated that, under specific model stability conditions, interventions on the dynamical system and the \SCM{} yield equivalent results.
To conduct causal inference over time, we have introduced two classes of interventions (additive and forcing) for  \VAR{}s.

\textbf{Limitations} 
When systems exhibit strongly nonlinear dynamics, linear \var{}s may prove less effective than alternative nonlinear approaches. 
Moreover, our framework requires prior knowledge of the causal graph.  In scenarios where this information is lacking, the process of causal discovery can present significant challenges.

\textbf{Future work}  We will investigate the use of non-linear \SP{}s, as they may lead to better convergence rates and allow for 
causal interpretation of a broader family of dynamical models.
Moreover, 
our work opens several interesting research directions (\cref{subsec:practical_implications} for concrete examples) and applications (\eg causal inference over time in high-dimensional contexts such as climate science).

\bibliography{references}

\begin{thebibliography}{42}
\providecommand{\natexlab}[1]{#1}

\bibitem[{Anderson(1994)}]{anderson1994statistical}
Anderson, T. 1994.
\newblock The Statistical Analysis of Time Series.
\newblock \emph{Wiley Series in Probability and Statistics}.

\bibitem[{Arnold et~al.(1995)Arnold, Jones, Mischaikow, Raugel, and Arnold}]{arnold1995random}
Arnold, L.; Jones, C.~K.; Mischaikow, K.; Raugel, G.; and Arnold, L. 1995.
\newblock \emph{Random dynamical systems}.
\newblock Springer.

\bibitem[{Bareinboim et~al.(2022)Bareinboim, Correa, Ibeling, and Icard}]{BareinboimCII22}
Bareinboim, E.; Correa, J.~D.; Ibeling, D.; and Icard, T. 2022.
\newblock On Pearl's Hierarchy and the Foundations of Causal Inference.
\newblock In \emph{Probabilistic and Causal Inference}, volume~36 of \emph{{ACM} Books}, 507--556. {ACM}.

\bibitem[{Belke and Polleit(2007)}]{belke2007ecb}
Belke, A.; and Polleit, T. 2007.
\newblock How the ECB and the US Fed set interest rates.
\newblock \emph{Applied Economics}, 39(17): 2197--2209.

\bibitem[{Bhattacharya and Majumdar(2003)}]{bhattacharya2003random}
Bhattacharya, R.; and Majumdar, M. 2003.
\newblock Random dynamical systems: a review.
\newblock \emph{Economic Theory}, 23: 13--38.

\bibitem[{Bongers, Blom, and Mooij(2018)}]{bongers2018causal}
Bongers, S.; Blom, T.; and Mooij, J.~M. 2018.
\newblock Causal modeling of dynamical systems.
\newblock \emph{arXiv preprint arXiv:1803.08784}.

\bibitem[{Bongers et~al.(2021)Bongers, Forr{\'e}, Peters, and Mooij}]{bongers2021foundations}
Bongers, S.; Forr{\'e}, P.; Peters, J.; and Mooij, J.~M. 2021.
\newblock Foundations of structural causal models with cycles and latent variables.
\newblock \emph{The Annals of Statistics}, 49(5): 2885--2915.

\bibitem[{Bongers et~al.(2016)Bongers, Peters, Sch{\"{o}}lkopf, and Mooij}]{bongers2016cyclic}
Bongers, S.; Peters, J.; Sch{\"{o}}lkopf, B.; and Mooij, J.~M. 2016.
\newblock Structural Causal Models: Cycles, Marginalizations, Exogenous Reparametrizations and Reductions.
\newblock \emph{CoRR}, abs/1611.06221.

\bibitem[{Chen et~al.(2023)Chen, Li, Arik, Yoder, and Pfister}]{chen2023tsmixer}
Chen, S.; Li, C.; Arik, S.~{\"{O}}.; Yoder, N.~C.; and Pfister, T. 2023.
\newblock TSMixer: An All-MLP Architecture for Time Series Forecast-ing.
\newblock \emph{Trans. Mach. Learn. Res.}, 2023.

\bibitem[{Das et~al.(2023)Das, Kong, Leach, Mathur, Sen, and Yu}]{das2023tide}
Das, A.; Kong, W.; Leach, A.; Mathur, S.; Sen, R.; and Yu, R. 2023.
\newblock Long-term Forecasting with TiDE: Time-series Dense Encoder.
\newblock \emph{Trans. Mach. Learn. Res.}, 2023.

\bibitem[{Deaton(1985)}]{deaton1985panel}
Deaton, A. 1985.
\newblock Panel data from time series of cross-sections.
\newblock \emph{Journal of econometrics}, 30(1-2): 109--126.

\bibitem[{Edwards, Lee, and Ray(1992)}]{edwards1992robust}
Edwards, R.~M.; Lee, K.~Y.; and Ray, A. 1992.
\newblock Robust optimal control of nuclear reactors and power plants.
\newblock \emph{Nuclear Technology}, 98(2): 137--148.

\bibitem[{Geiger et~al.(2015)Geiger, Zhang, Schoelkopf, Gong, and Janzing}]{geiger15}
Geiger, P.; Zhang, K.; Schoelkopf, B.; Gong, M.; and Janzing, D. 2015.
\newblock Causal Inference by Identification of Vector Autoregressive Processes with Hidden Components.
\newblock In Bach, F.; and Blei, D., eds., \emph{Proceedings of the 32nd International Conference on Machine Learning}, volume~37 of \emph{Proceedings of Machine Learning Research}, 1917--1925. Lille, France: PMLR.

\bibitem[{Granger(1969)}]{granger1969investigating}
Granger, C.~W. 1969.
\newblock Investigating causal relations by econometric models and cross-spectral methods.
\newblock \emph{Econometrica: journal of the Econometric Society}, 424--438.

\bibitem[{Hamilton(1994)}]{hamilton1994time}
Hamilton, J.~D. 1994.
\newblock Time Series Analysis.

\bibitem[{Herzen et~al.(2022)Herzen, L{\"a}ssig, Piazzetta, Neuer, Tafti, Raille, Van~Pottelbergh, Pasieka, Skrodzki, Huguenin et~al.}]{herzen2022darts}
Herzen, J.; L{\"a}ssig, F.; Piazzetta, S.~G.; Neuer, T.; Tafti, L.; Raille, G.; Van~Pottelbergh, T.; Pasieka, M.; Skrodzki, A.; Huguenin, N.; et~al. 2022.
\newblock Darts: User-friendly modern machine learning for time series.
\newblock \emph{Journal of Machine Learning Research}, 23(124): 1--6.

\bibitem[{Hildebrand(1987)}]{hildebrand1987introduction}
Hildebrand, F.~B. 1987.
\newblock \emph{Introduction to numerical analysis}.
\newblock Courier Corporation.

\bibitem[{Hyv{\"{a}}rinen et~al.(2010)Hyv{\"{a}}rinen, Zhang, Shimizu, and Hoyer}]{hyvarinen2010estimation}
Hyv{\"{a}}rinen, A.; Zhang, K.; Shimizu, S.; and Hoyer, P.~O. 2010.
\newblock Estimation of a Structural Vector Autoregression Model Using Non-Gaussianity.
\newblock \emph{J. Mach. Learn. Res.}, 11: 1709--1731.

\bibitem[{Janzing, Rubenstein, and Sch{\"o}lkopf(2018)}]{janzing2018structural}
Janzing, D.; Rubenstein, P.; and Sch{\"o}lkopf, B. 2018.
\newblock Structural causal models for macro-variables in time-series.
\newblock \emph{arXiv preprint arXiv:1804.03911}.

\bibitem[{Jornet(2023)}]{jornet2023theory}
Jornet, M. 2023.
\newblock Theory and methods for random differential equations: a survey.
\newblock \emph{SeMA Journal}, 80(4): 549--579.

\bibitem[{Karimi et~al.(2022)Karimi, Barthe, Sch{\"o}lkopf, and Valera}]{karimi2022survey}
Karimi, A.-H.; Barthe, G.; Sch{\"o}lkopf, B.; and Valera, I. 2022.
\newblock A survey of algorithmic recourse: contrastive explanations and consequential recommendations.
\newblock \emph{ACM Computing Surveys}, 55(5): 1--29.

\bibitem[{Kilian and L{\"u}tkepohl(2017)}]{kilian2017structural}
Kilian, L.; and L{\"u}tkepohl, H. 2017.
\newblock \emph{Structural vector autoregressive analysis}.
\newblock Cambridge University Press.

\bibitem[{Kingma and Ba(2015)}]{KingmaB14}
Kingma, D.~P.; and Ba, J. 2015.
\newblock Adam: {A} Method for Stochastic Optimization.
\newblock In \emph{{ICLR} (Poster)}.

\bibitem[{Koop, Korobilis et~al.(2010)}]{koop2010bayesian}
Koop, G.; Korobilis, D.; et~al. 2010.
\newblock Bayesian multivariate time series methods for empirical macroeconomics.
\newblock \emph{Foundations and Trends{\textregistered} in Econometrics}, 3(4): 267--358.

\bibitem[{Lasota and Mackey(1989)}]{lasota1989stochastic}
Lasota, A.; and Mackey, M.~C. 1989.
\newblock Stochastic perturbation of dynamical systems: The weak convergence of measures.
\newblock \emph{Journal of Mathematical Analysis and Applications}, 138(1): 232--248.

\bibitem[{Lorbeer and Mohsen(2023)}]{Lorbeer2023cycliccausaldiscovery}
Lorbeer, B.; and Mohsen, M. 2023.
\newblock Comparative Study of Causal Discovery Methods for Cyclic Models with Hidden Confounders.
\newblock In \emph{CogMI}, 103--111. {IEEE}.

\bibitem[{{\L}oskot and Rudnicki(1995)}]{loskot1995limit}
{\L}oskot, K.; and Rudnicki, R. 1995.
\newblock Limit theorems for stochastically perturbed dynamical systems.
\newblock \emph{Journal of applied probability}, 32(2): 459--469.

\bibitem[{L{\"u}tkepohl(2005)}]{lutkepohl2005new}
L{\"u}tkepohl, H. 2005.
\newblock \emph{New introduction to multiple time series analysis}.
\newblock Springer Science \& Business Media.

\bibitem[{Malinsky and Spirtes(2018)}]{malinskyS18}
Malinsky, D.; and Spirtes, P. 2018.
\newblock Causal Structure Learning from Multivariate Time Series in Settings with Unmeasured Confounding.
\newblock In \emph{CD@KDD}, volume~92 of \emph{Proceedings of Machine Learning Research}, 23--47. {PMLR}.

\bibitem[{Moneta et~al.(2011)Moneta, Chla{\ss}, Entner, and Hoyer}]{moneta2011causal}
Moneta, A.; Chla{\ss}, N.; Entner, D.; and Hoyer, P. 2011.
\newblock Causal search in structural vector autoregressive models.
\newblock In \emph{NIPS Mini-Symposium on Causality in Time Series}, 95--114. PMLR.

\bibitem[{Moneta et~al.(2013)Moneta, Entner, Hoyer, and Coad}]{moneta2013causal}
Moneta, A.; Entner, D.; Hoyer, P.~O.; and Coad, A. 2013.
\newblock Causal inference by independent component analysis: Theory and applications.
\newblock \emph{Oxford Bulletin of Economics and Statistics}, 75(5): 705--730.

\bibitem[{Montgomery, Jennings, and Kulahci(2015)}]{montgomery2015introduction}
Montgomery, D.~C.; Jennings, C.~L.; and Kulahci, M. 2015.
\newblock \emph{Introduction to time series analysis and forecasting}.
\newblock John Wiley \& Sons.

\bibitem[{Mooij, Janzing, and Sch{\"{o}}lkopf(2013)}]{Mooij2013deterministic_case}
Mooij, J.~M.; Janzing, D.; and Sch{\"{o}}lkopf, B. 2013.
\newblock From Ordinary Differential Equations to Structural Causal Models: the deterministic case.
\newblock In Nicholson, A.~E.; and Smyth, P., eds., \emph{Proceedings of the Twenty-Ninth Conference on Uncertainty in Artificial Intelligence, {UAI} 2013, Bellevue, WA, USA, August 11-15, 2013}. {AUAI} Press.

\bibitem[{Parmezan, Souza, and Batista(2019)}]{parmezan2019evaluation}
Parmezan, A. R.~S.; Souza, V.~M.; and Batista, G.~E. 2019.
\newblock Evaluation of statistical and machine learning models for time series prediction: Identifying the state-of-the-art and the best conditions for the use of each model.
\newblock \emph{Information sciences}, 484: 302--337.

\bibitem[{Pearl(2009)}]{pearl2009causality}
Pearl, J. 2009.
\newblock \emph{Causality}.
\newblock Cambridge university press.

\bibitem[{Runge et~al.(2019)Runge, Bathiany, Bollt, Camps-Valls, Coumou, Deyle, Glymour, Kretschmer, Mahecha, Mu{\~n}oz-Mar{\'\i} et~al.}]{runge2019inferring}
Runge, J.; Bathiany, S.; Bollt, E.; Camps-Valls, G.; Coumou, D.; Deyle, E.; Glymour, C.; Kretschmer, M.; Mahecha, M.~D.; Mu{\~n}oz-Mar{\'\i}, J.; et~al. 2019.
\newblock Inferring causation from time series in Earth system sciences.
\newblock \emph{Nature communications}, 10(1): 2553.

\bibitem[{Sigmund and Ferstl(2021)}]{sigmund2021panel}
Sigmund, M.; and Ferstl, R. 2021.
\newblock Panel vector autoregression in R with the package panelvar.
\newblock \emph{The Quarterly Review of Economics and Finance}, 80: 693--720.

\bibitem[{Sims(1980)}]{sims1980macroeconomics}
Sims, C.~A. 1980.
\newblock Macroeconomics and reality.
\newblock \emph{Econometrica: journal of the Econometric Society}, 1--48.

\bibitem[{Toner and Darlow(2024)}]{linear_analysis_toner_2024}
Toner, W.; and Darlow, L.~N. 2024.
\newblock An Analysis of Linear Time Series Forecasting Models.
\newblock In \emph{Forty-first International Conference on Machine Learning}.

\bibitem[{Wold(1938)}]{wold1938study}
Wold, H. 1938.
\newblock \emph{A study in the analysis of stationary time series}.
\newblock Ph.D. thesis, Almqvist \& Wiksell.

\bibitem[{Wunsch et~al.(2022)Wunsch, Russo, Mouchart, and Orsi}]{wunsch2022time}
Wunsch, G.; Russo, F.; Mouchart, M.; and Orsi, R. 2022.
\newblock Time and causality in the social sciences.
\newblock \emph{Time \& Society}, 31(2): 177--204.

\bibitem[{Zeng et~al.(2023)Zeng, Chen, Zhang, and Xu}]{zeng2023dlinear}
Zeng, A.; Chen, M.; Zhang, L.; and Xu, Q. 2023.
\newblock Are Transformers Effective for Time Series Forecasting?
\newblock \emph{Proceedings of the AAAI Conference on Artificial Intelligence}, 37(9): 11121--11128.

\end{thebibliography}

\clearpage
\appendix
    
\section{Difference Equations and Stochastic Processes Equilibration}
\label{app:difference_equations}

Difference Equations can serve as discrete approximations of differential equations, sharing some fundamental theoretical properties. In many applications, the former are approximation tools to simulate the latter numerically. 
As with differential equations, it can be beneficial to introduce stochastic components into the equation to model phenomena that are not entirely predictable. 
This unpredictability may stem from incomplete knowledge of the relevant variables of the process or its intrinsic randomness. 
We will denote by \DE{} the class of difference equations in general, regardless of whether they are deterministic or not. 
Including stochastic components in differential equations gives rise to various classes of equations.
This section introduces the notions of Ordinary and Random Difference Equations. Understanding these concepts is helpful for contextualizing our work, and for clarifying how it extends the ideas presented in~\citep{Mooij2013deterministic_case}. 
\subsection{Difference equations}
\paragraph{Ordinary Difference Equation}
An Ordinary Difference Equation (\ODE{}) is an expression that describes the evolution of an indexed variable $x_t$ through a functional relationship
\begin{equation*}
x_t = f(x_{<t}),
\end{equation*}
where $x_{<t}:=\{x_{t-1},x_{t-2},\dots\}$ represents the past of $x_t$. Specifically, a $p^{th}$-order \ODE{} is an expression where the preceding variables of $x_t$ include up to $x_{t-p}$, i.e, $x_{<t}=\{x_{t-i}\}_{i=1}^p$. 

\paragraph{Random Difference Equation} 
A Random Difference Equation (\RDE{}) is a \DE{} of the form
\begin{equation}\label{eq:RDE}
    \X_{t} = f(\X_{<t},\vectornoise),
\end{equation}
where $\vectornoise:\Omega\to\sR^e$ is a random variable\footnote{In ~\citep{bongers2018causal}, \(\vectornoise\) is regarded as a stochastic process that converges eventually or asymptotically to a time-independent random variable. In this work, simplifying the concept, we consider it as a random variable throughout the whole process.}  and such that for every $\omega \in \Omega$, the equation 
\begin{equation}\label{eq:RDE_sample_path_solution}
    \X_{t}(\omega) = f(\X_{<t}(\omega),\vectornoise(\omega))
\end{equation}
is an \ODE{}.

Note that $\X_t$ in \cref{eq:RDE} is a random variable and characterizes a \SP{} $\X:T\times\Omega\to\sR^n$ induced by $\vectornoise$.
In this sense, we say that $\X$ is a solution of \cref{eq:RDE}. Equivalently, we say that $\X$ is a \textit{solution} of \cref{eq:RDE} if \cref{eq:RDE_sample_path_solution} is satisfied for \textit{almost all} $\omega\in\Omega$.
\begin{example}[\RDE{}]\label{ex:rde}
The system described by 
\begin{equation}\label{eq:rde_example}
    \begin{cases}
        X_t=a(\Omega)X_{t-1} + b(\Omega)\\
        X_0=X_0(\Omega)
    \end{cases}
\end{equation} 
represents a first-order linear \RDE{} equipped with an initial condition for $X_0$, in which the stochasticity of $\omega$ generates a distribution over $X_0$ and over the parameters of the equation $a,b$. 
The general solution of \Eqref{eq:rde_example} can be found by recursive substitution, obtaining $X_t(w) = a(\omega)^t X_0(\omega) + b(\omega)\sum_{i=0}^{t-1}a(\omega)^i$.
\end{example}

\subsection{Stochastic Processes Equilibration}
Let $\X$ be a \SP{}.  
We say that $\X$ \textit{(strongly) equilibrates} if $\X_t$ converges to a random variable $\Xinf$ for $t\to\infty$ almost surely, i.e., 
\begin{equation*}
    \Prb(\limsup_{t\to\infty} \{\omega\in\Omega:|\X_t(\omega)-\Xinf(\omega)|>\varepsilon\})=0\quad \forall \varepsilon>0, 
\end{equation*}
\text{denoted by $\X_t\overset{a.s}{\to}\Xinf$}. %
We call $\Xinf$ the \textit{equilibrium state} of $\X$.
Moreover, we say that $\X$ \textit{weakly equilibrates} if $\X_t$ converges in distribution to a random variable $\Xinf$, i.e., for all $x$ for which $F_\X$ (the CDF of $\X$) is continuous, 
\begin{equation*}
    \lim_{t\to\infty}F_{\X_t}(x)=F_{\Xinf}(x),\quad \text{denoted by $\X_t\indist\Xinf$}.
\end{equation*}
In general, $\X_t\overset{a.s.}{\to}\Xinf \Rightarrow \X_t\indist\Xinf$. 
\begin{remark}
 If $\X$ is a solution of an \RDE{}, almost sure convergence is achieved if and only if, for almost all $\omega\in\Omega$, the solution of the ordinary difference equation (\ODE{}) in \Eqref{eq:RDE_sample_path_solution} converges asymptotically to a fixed value $\Xinf(\omega)$. 
 On the other hand, a \SP{} that is a solution of a \SDE{} (for convenience, we recall \Eqref{eq:sde}: $
\X_t = \bm{f}(\X_{<t}) + \bm{g}(\X_{<t}) \odot \vectornoise_t$) cannot equilibrate unless the term $g(\X_{<t})$ converges as $t\to\infty$ to $0$, thereby transforming the equation into a \RDE{}. Apart from this scenario, no solutions of a \SDE{} can strongly equilibrate.
\end{remark}
\begin{example}[\RDE{} equilibration]
    Consider example \ref{ex:rde}. Assuming $a(\omega)<1 \quad \forall \omega\in\Omega$, 
    \begin{equation*}
        X_t\overset{a.s}{\to}X_\infty=b(\Omega)\sum_{i=0}^{\infty}a(\Omega)^i = \frac{b(\Omega)}{1-a(\Omega)}.
    \end{equation*}
\end{example}

\subsection{Equilibration as a map from \RDE{}s to \SCM{}s}\label{app:equilibration_RDE_SCM}
\begin{figure}[t!]
    \centering
    \begin{tikzpicture}
    [scale=0.8, every node/.style={rectangle split, rectangle split parts=2,rectangle split horizontal=true, rounded corners=0.1cm, minimum height=1cm, align=center,scale=0.8}]

    \node[draw] (1) at (0,2) {\nodepart[text width=0.6cm]{one}$\dequation$\nodepart[text width=0.6cm]{two}$\X_t$};
    \node[draw] (2) at (4,2) {\nodepart[text width=0.6cm]{one}$\scm$\nodepart[text width=0.6cm]{two}$\X_\infty$};
    \node[draw] (3) at (0,0) {\nodepart[text width=0.6cm]{one}$\dequation^\intervention$\nodepart[text width=0.6cm]{two}$ \X_t^\intervention$};
    \node[draw] (4) at (4,0) {\nodepart[text width=0.6cm]{one}$\scm^\intervention$\nodepart[text width=0.6cm]{two}$ \X_\infty^\intervention$};

    \node [] at (2.14,1) {$=?$};

    \draw[|->,shorten >=4pt,shorten <=4pt,color=MidnightBlue] (1) -- (2) node[midway, above] {$t \to \infty$};
    \draw[|->,shorten >=4pt,shorten <=4pt,color=BrickRed] (1) -- (3) node[midway, left] {$\intervention$};
    \draw[|->,shorten >=4pt,shorten <=4pt,color=MidnightBlue] (2) -- (4) node[midway, right,xshift=0.33cm] {$\intervention$};
    \draw[|->,shorten >=4pt,shorten <=4pt,color=BrickRed] (3) -- (4) node[midway, below] {$t \to \infty$};

    \draw[thick,-stealth,in=0,out=90,shorten >=4pt,shorten <=4pt,color=MidnightBlue] (1.3,1.5) .. controls +(0:1.1cm) and +(90:1.1cm) .. (2.7,0.5);
    \draw[thick,-stealth,in=-90,out=180,shorten >=4pt,shorten <=4pt,color=BrickRed] (1.3,1.5) .. controls +(-90:1.1cm) and +(180:1.1cm) .. (2.7,0.5);
\end{tikzpicture}
\caption{Illustration of the relationship between \RDE{}s \& \SCM{}s and the effect of causal interventions on them. Applying equilibration (here denoted by $t \to \infty$) before or after intervention does not change the final resulting distribution $\X_\infty^\intervention$, i.e., the diagram commutes. For \SDE{}s, when variables are localized in times,  commutation is not preserved, that is $(\X_\infty)^{\intervention_{\scm}} \neq (\X^{\intervention_{\dequation}})_\infty$.}\label{fig:simple_diagram_commutes}
\end{figure}
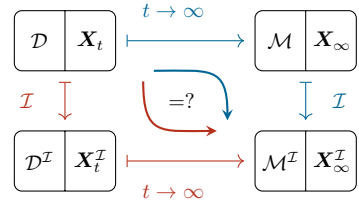

The process equilibration allows defining a map that associates each \RDE{} with an \SCM{}. 
Furthermore, this map preserves the semantic of intervention, in the same sense of \cref{subsec:mapping_sde_scm}. Visually, the diagram in \cref{fig:simple_diagram_commutes} commutes, i.e., applying equilibration before or after intervention doesn't change the resulting $\X^\intervention_\infty$.

Consider a Simple \RDE{} \(\dequation\) and its solution \(\X\). Let \(\Xinf\) be the random variable which \(\X\) equilibrates to. 
Then it is possible to associate an SCM \(\scm_{\dequation}=(\rmF,\rmE)\) where \(\rmE=\vectornoise_{\dequation}\) and \(\rmF\) is defined through the equilibrium relation
\begin{equation}
    \Xinf = f_{\dequation}(\Xinf(<t,\cdot),\rmE)=\tilde{f}(\Xinf,\rmE),
\end{equation}
where \(\Xinf(t,\cdot)=\Xinf \: \forall t\) is the constant stochastic process associated with \(\Xinf\). In other words, \(\scm_\dequation\) fully inherits the functional relationships of \(\dequation\), substituting dynamic variables with their stationary counterparts. A proof for the general case in continuous-time can be found in ~\citep{bongers2018causal}. The discrete-time variant follows directly by applying methods of numerical approximation~\citep{hildebrand1987introduction,jornet2023theory}.

\subsection{Equilibration as a map from \SDE{}s to \SCM{}s}\label{app:equilibration_SDE_SCM}
Associating an \SDE{} with an \SCM{} in non-linear settings involves two steps. The first concerns the convergence conditions of the dynamical system to a stationary (observational) distribution. The second step involves constructing the mapping between interventions in the two different models and proving their equivalence. In this work, we address the first point. As for the second, a formal treatment of the nonlinear case has yet to be completed. While intuitively the existence of an appropriate \SCM{} seems more than reasonable, its explicit characterization requires mathematical tools beyond the scope of this work. We intend to tackle this step in future research.
\paragraph{Existence of an observational limit distribution}
Several works discuss the existence and uniqueness of a limit distribution for stochastic systems ~\citep{arnold1995random, bhattacharya2003random, loskot1995limit}.
Here, we adapt the following theorem from \citep{lasota1989stochastic} to our formulation in \cref{eq:sde}:
%
\begin{theorem}\label{thm:sdetodistribution}[Adapted \cref{eq:sde} from~\citep{lasota1989stochastic}]
Consider a \SDE{}\footnote{
\cref{thm:sdetodistribution} considers Markovian systems, while our formulation allows for $f$ and $g$ to depend on multiple past states. As long as the states considered are finite in number, i.e., $s$ depends on $\X_{t-1},\X_{t-2},\dots,\X_{t-p},\: p<\infty$, it is always possible to represent the system as if it were Markovian, through the transformation $\bm{Z}_t := [\X_t,\X_{t-1},\dots,\X_{t-p+1}]$. Given this fact, for ease of exposition, we will consider strictly Markovian systems in the discussion.} 
described by 
\begin{equation*}
    \X_t = f(\X_{t-1}) + g(\X_{t-1})\odot\vectornoise_t.
\end{equation*}
If the following conditions hold: i) $f,g:\mathcal{X}\to\mathcal{X}$ are continuous in $\X_{t-1}$, and $\mathcal{X}\subset\R^{\sizex}$ is closed; ii) 
     $\vectornoise_t$ is i.i.d. white noise; and iii)
     for every $t$,
    \begin{align*}
        &(\text{iii}.1) 
        \quad \forall x,y\in\mathcal{X},\: x\neq y\\\quad&\expected{\norm{f(x)+g(x)\odot\vectornoise_t-f(y)-g(y)\odot\vectornoise_t}}< \norm{x-y} \\\\
        &(\text{iii}.2) 
        \quad \forall x\in\mathcal{X}\\
        &\expected{\norm{f(x) + g(x)\odot\vectornoise_t}^2}\leq \alpha \norm{x}^2 + \beta ,\: \alpha<1,
    \end{align*}
where $\norm{\cdot}$ is the Euclidean norm in $\sR^\sizex$. 
Then, there exists a unique invariant distribution $p(\Xinf)$ to which $p(\X_t)$ converges as $t\to\infty$, regardless of the initial state $\X_0$.
\end{theorem}
%
%

\paragraph{Rediscussing \cref{thm:sdetodistribution} in terms of Lipschitz continuity}

For a more intuitive interpretation of \cref{thm:sdetodistribution}, we show that condition (iii) holds under Lipschitz continuity\footnote{A function $f$ is Lipschitz with constant $L$ if for all $x,y$, it holds that $\|f(x)-f(y)\| \leq L \|x-y\|$.} of $f$ and $g$. 
Given $f,g:\sR^\sizex\to\sR^\sizex$ Lipschitz continuous with Lipschitz constants $L_f,L_g$, we want to find sufficient conditions on $L_f$ and $L_g$ such that, for every fixed $x,y$,
\begin{align*}
         (1) \quad &\expected{\norm{f(x)+g(x)\odot\vectornoise-f(y)-g(y)\odot\vectornoise}}< \norm{x-y}. \\
         (2) \quad &\expected{\norm{f(x) + g(x)\odot\vectornoise}^2}\leq \alpha \norm{x}^2 + \beta ,\: \alpha<1.
    \end{align*}
    $\vectornoise$ being a random vector s.t. each of its component satisfy $\expected{\noise}=0,\expected{|\noise|^2}<\infty$. For $(1)$ we calculate
    \begin{align*}
        &\expected{\norm{f(x)+g(x)\odot\vectornoise-f(y)-g(y)\odot\vectornoise}}\leq \\
        &\norm{f(x)-f(y)}+\norm{g(x)-g(y)}\: \expected{\norm{\vectornoise}}\leq \\
        &L_f \norm{x-y} + L_g \norm{x-y} \: \expected{\norm{\vectornoise}}= \\
        &(L_f + L_g \: \expected{\norm{\vectornoise}})\norm{x-y}<\norm{x-y} \\
        &\iff \quad \boxed{L_f + L_g \: \expected{\norm{\vectornoise}}< 1}
    \end{align*}

    The same procedure for $(2)$ yields
    \begin{align*}
        &\expected{\norm{f(x) + g(x)\odot\vectornoise}^2} = \\
        &\expected{\norm{f(x)}^2} + \expected{\norm{g(x)\odot\vectornoise}^2}= \quad \{\hat{f}(x):=f(x)-f(0)\} \\
        &\expected{\norm{\hat{f}(x)+f(0)}^2}+\expected{\norm{\hat{g}(x)\odot \vectornoise+g(0)\odot \vectornoise}^2}\leq \\
        &L_f^2 \norm{x}^2 + 2 |\innerproduct{\hat{f}(x)}{f(0)}|  + \norm{f(0)}^2 + \\
        &L_g^2\norm{x}^2 \sigma^2 + 2|\innerproduct{\hat{g}(x)}{g(0)}|\sigma^2 + \norm{g(0)}^2 \sigma^2 \leq \quad \{\sigma^2:=\expected{\norm{\vectornoise}^2}\} \\
        &(L_f^2+L_g^2\sigma^2)\norm{x}^2+2(L_f\norm{f(0)} +L_g\norm{g(0)}\sigma^2)\norm{x} + \\
        &\norm{f(0)}^2 + \norm{g(0)}^2 \sigma^2 =: a\norm{x}^2+m\norm{x}+b.
    \end{align*}
    Observing that every polynomial $ax^2 + mx + b$ with $a > 0$ can be bounded above by $\alpha x^2 + \beta$ assuming $\alpha > a$ and provided that $\beta$ is sufficiently large, we finally obtain
    \begin{equation*}
    \begin{split}
        &a\norm{x}^2+m\norm{x}+b \leq \alpha x^2 + \beta, \: \alpha<1 \\
        &\iff \quad \boxed{L_f^2+L_g^2\expected{\norm{\vectornoise}^2} < 1}
    \end{split}
    \end{equation*}

In the simple case in which $g\equiv 1$, we obtain the inequality $L_f<1$, i.e.,~$f$ \textit{is a contraction}. 
Condition \((2)\) disappears, as $L_f^2<L_f<1$.
In the general case, we interpret $f$ as a contraction map on $\Xinf$ and $g \odot \vectornoise_t$ as a diffusion process. The equilibrium between these processes generates the invariant distribution.

\section{Vector Autoregressive Models}
\label{app:svars}
\subsection{Properties}\label{app:svarproperties}
A VAR process $\X_t$ is stable if all roots of the determinantal polynomial of the VAR operator are outside the complex unit circle. This condition can be formally expressed as:

\begin{equation}\label{eq:determinantal}
    det(I_K - \A_1z - \ldots - \A_\order z^\order) \neq 0 \quad \forall z \in \mathbb{C}, |z| \leq 1,
\end{equation}
where $\mathbb{C}$ denotes the set of complex numbers. The stability condition ensures that the process is stationary~\citep{lutkepohl2005new}. Consequently, a stable VAR($\order$) process $\X_t$ is stationary.

In~\citep{wold1938study}, the author introduced the Wold decomposition Theorem, which states that every covariance-stationary time series $\X_t$ can be written as the sum of two uncorrelated processes as $\X_t = \bm{Z}_t + \bm{Y}_t$
where $\bm{Z}_t$ is a deterministic component that can be forecast perfectly from its past, and $\bm{Y}_t$ is a purely stochastic process with an infinite-order Moving Average (MA) representation: 
\begin{equation}\label{eq:moving_average}
\bm{Y}_t = \sum_{i=0}^\infty \Phi_i \vectornoisevar_{t-i}.
\end{equation}
Suppose the $\Phi_i$ are absolutely summable and that there exists an operator $A(L)$ with absolutely summable coefficient matrices satisfying $A(L)\Phi(L) = I_K$.
Then $\Phi(L)$ is invertible with $\Phi(L)^{-1} = A(L)$ and $\bm{Y}_t$ can be approximated arbitrarily well by a finite-order \VAR{}($\order$) with $\order$ sufficiently large, and $A(L) = I - \A_1 L - \dots - \A_p L^\order$, where $I$ is the identity matrix. 

In particular, the matrices $\Phi_i$ of a \VAR{}(p) model can be calculated by recursively applying the formula
\begin{equation}\label{eq:recursive_calcultation_phi}
    \Phi_0 = I, \quad \Phi_i = \sum_{j=1}^i \Phi_{i-j}\A_j, \quad \text{where} \; \A_j = 0 \; \text{for} \; j>p.
\end{equation}
\begin{remark*}
    Consider the model \( A(L)\X_t = \bm{\nu} + \vectornoisevar_t \), where \( \bm{\nu} \) is a (constant) intercept term. Using Wold's Theorem, we can express it by linearity of $\response$ as \( \X_t = \response(L)\bm{\nu} + \response(L)\vectornoisevar_t \). Noting that \( L\bm{\nu}=\bm{\nu} \), we obtain  
    \begin{equation*}
        \X_t=\bm{\mu} + \response(L)\vectornoisevar
    \end{equation*}
    where \( \bm{\mu}=\response(1)\bm{\nu}:=\sum_{l=0}^\infty\response_l\bm{\nu} \). Due to the invertibility of \( \response(L) \), we can express the same relation through \( \bm{\nu}=\response(1)^{-1}\bm{\mu}=A(1)\bm{\mu}=[1-\A_1-\dots-\A_p]\bm{\mu} \). From this observation, we automatically obtain the equality
    \begin{equation}\label{eq:phi1a1}
        \sum_{l=0}^\infty \response_l = \response(1) = A(1)^{-1} = [1-\A_1 - \dots - \A_p]^{-1}.
    \end{equation}
    
\end{remark*}

\subsection{Proof of \cref{thm:var_to_scm} - From \VAR{}s to \SCM{}s (\cref{subsec:mapping_var_scm})  }\label{app:explicit_solution_var_svar}

Consider the $MA(\infty)$ representation of $\X_t$ (since we can always apply a translation to it without modifying the dynamic of the stochastic part, let's assume for simplicity that \( \expected{\X_t} = 0 \)):
\begin{equation*}
    \X_t = \sum_{l=0}^{\infty} \response_l \vectornoisevar_{t-l}.
\end{equation*}
To capture the long-run relationships of the process, we apply the transformation 
\begin{equation}\label{eq:t-trasformation} \quad 
    \gT: \X_t \mapsto \bm{Z}_t = \frac{1}{\sqrt{t}} \sum_{i=1}^t \X_i,
\end{equation}

By examining the equilibration $\bm{Z}_t \to \bm{Z}_\infty$, we can link a stable $\VAR{}(p)$ process to the following \SCM{}:
\begin{equation*}
    \tilde{\X} =\tilde{\A} \tilde{\X} + \tilde{\vectornoisevar}, \quad \text{where} \quad  \tilde{\A} := \left[ \sum_{i=1}^p \A_i \right]
\end{equation*} 

Here, we outline the proof:
\begin{align*}
    \lim_{t\to\infty}\bm{Z}_t &= \lim_{t\to\infty} \frac{1}{\sqrt{t}}\sum_{i=1}^t \sum_{l=0}^{\infty} \response_l \vectornoisevar_{i-l} = \\
    &\text{(by a.s. absolute convergence)}\\
    &= \lim_{t\to\infty} \frac{1}{\sqrt{t}} \sum_{l=0}^{\infty} \sum_{i=1}^t \response_l \vectornoisevar_{i-l}= \\
    &= \lim_{t\to\infty} \sum_{l=0}^{\infty} \response_l \frac{1}{\sqrt{t}} \sum_{i=1}^t  \vectornoisevar_{i-l}= \\
    &\text{(by dominated convergence)}\\
    &=  \sum_{l=0}^{\infty} \response_l \lim_{t\to\infty} \frac{1}{\sqrt{t}} \sum_{i=1}^t  \vectornoisevar_{i-l}= \\ 
    &\text{(by central limit theorem)}\\
    &\overset{d}{=}\left[ \sum_{l=0}^{\infty} \response_l \right] \tilde{\vectornoisevar}, \quad \tilde{\vectornoisevar} \sim \normal (0,\Sigma_{\vectornoisevar})= \quad (\star)\\
    &= \response(1) \tilde{\vectornoisevar} = A(1)^{-1} \tilde{\vectornoisevar} = \left[1-\A_1-\dots-\A_p\right]^{-1} \tilde{\vectornoisevar},
\end{align*}
that is 
\begin{equation}\label{eq:explicit_scm}
\begin{split}
    \bm{Z}_\infty \overset{d}{=} A(1)^{-1} \tilde{\vectornoisevar}, \quad \text{or equivalently} \\
    \bm{Z}_\infty \overset{d}{=} [\A_1+\dots+\A_p] \bm{Z}_\infty + \tilde{\vectornoisevar} \quad \qedsymbol
\end{split}
\end{equation}
For an alternative proof of $(\star)$ see~\citep[194]{hamilton1994time} and~\citep[429]{anderson1994statistical}, from which we report the main statement:
\begin{theorem}\label{thm:teorem_hamilton}~\citep[429]{anderson1994statistical}
Let $\X$ be a \SP{} defined by
\begin{equation}
    \X_t = \bm{\mu} + \sum_{l=0}^\infty \Phi_i \vectornoisevar_{t-l}
\end{equation}
such that $\vectornoisevar_t$ is white noise and $\sum_{l=0}^\infty |\Phi_i|<\infty$. Then
\begin{equation*}
    \bm{Z}_t = \frac{1}{\sqrt{t}}\sum_{i=1}^{t} (\X_i-\bm{\mu}) \quad \indist \quad \normal(0,\sum_{i=-\infty}^{\infty}\gamma_i) \quad \text{as} \quad t\to\infty,
\end{equation*}

where $\gamma_i$ represents the autocovariance $\Var(\X_t,\X_{t-i})$. 
\end{theorem}
Exploiting the i.i.d assumption for $\vectornoisevar_t$, $\gamma_i$ admits the representation
$$\gamma_i=\sum_{l=0}^{\infty}\response_{l+i} \Sigma_{\vectornoisevar} \response_l'= \sum_{k=0}^{\infty} \sum_{l=0}^{\infty} \delta_{k,l+i} \response_{k} \Sigma_{\vectornoisevar} \response_l'.$$
Using $\sum_{i=-\infty}^{\infty} \delta_{i,j} = 1 \: \forall j$, we rewrite
\begin{align}
\begin{split}\label{autocovariance_to_response}
    \sum_{i=-\infty}^{\infty}\gamma_i &= \sum_{i=-\infty}^{\infty} \sum_{k=0}^{\infty} \sum_{l=0}^{\infty} \delta_{k,l+i} \response_{k} \Sigma_{\vectornoisevar} \response_l'= \\
    &\text{(absolute convergence)}\\
    &= \sum_{k=0}^{\infty} \sum_{l=0}^{\infty} \sum_{i=-\infty}^{\infty}\delta_{k,l+i} \response_{k} \Sigma_{\vectornoisevar} \response_l'=\\
    &= \sum_{k=0}^{\infty} \sum_{l=0}^{\infty} \response_{k} \Sigma_{\vectornoisevar} \response_l'=\\
    &= \left[ \sum_{k=0}^{\infty} \response_{k} \right] \Sigma_{\vectornoisevar} \left[ \sum_{l=0}^{\infty}\response_l' \right] = \response(1) \Sigma_{\vectornoisevar} \response(1)',
\end{split}
\end{align}
that is compatible with \cref{eq:explicit_scm}. 
Additionally, it is worth noting that given a positive definite covariance matrix  $\Sigma_{\bm{Z}}$ and a fixed order of variables in $\bm{Z}$, Cholesky decomposition~\citep{lutkepohl2005new} guarantees the existence and uniqueness of a triangular matrix $L$ such that $\Sigma_{\bm{Z}} = LL'$. 
The same holds for LDL decomposition~\citep{lutkepohl2005new}, namely $\Sigma_{\bm{Z}} = LDL'$. 
Setting $L=\response(1)$ and $D=\Sigma_{\vectornoisevar}$, we can deduce \cref{eq:explicit_scm} from~\cref{thm:teorem_hamilton}.

\paragraph{Equilibration commutes with additive intervention}

We show that applying the same additive intervention on the \VAR{} and its associated \SCM{} yields the same interventional distribution. We start by studying the long term effect of the additive term $\bm{F}$ over the \VAR{} model:

\begin{equation*}
\begin{split}
    &A(L)\X_t = \vectornoisevar_t + \bm{F} \quad \Rightarrow \quad \X_t = \Phi(L)\vectornoisevar_t + \Phi(1)\bm{F}, \\
    &\text{where} \quad \Phi(1)\bm{F} = \sum_{l=0}^\infty\Phi_l \bm{F}.
\end{split}
\end{equation*}
By applying $\gT$ (\cref{eq:t-trasformation}) and \cref{eq:explicit_scm}, we obtain
\begin{equation*}
    \bm{Z}^\intervention_\infty = \Phi(1) \tilde{\vectornoisevar} + \Phi(1)\bm{F}.
\end{equation*}

Using \cref{thm:var_to_scm}, we study the same intervention of the associated \SCM{}:

\begin{align*}
    &\tilde{\X}=\tilde{\A}\tilde{\X} + \noisescm + \bm{F} \Rightarrow \\
    & [I-\tilde{\A}]\X =\noisescm+ \bm{F} \Rightarrow \\
    & \tilde{\X} = [I-\tilde{\bm{A}}]^{-1}\tilde{\vectornoisevar} + [I-\tilde{\bm{A}}]^{-1}\bm{F}.
\end{align*}

The correspondence follows directly from $\Phi(1) = \A(1)^{-1} = [I- \tilde{\A}]^{-1}.$

\paragraph{Equilibration commutes with forcing intervention}

We report \cref{eq:do-like} for clarity of exposition:
\begin{equation*}
    \A(L)\X_t = \bm{\nu} + \vectornoisevar_t + \mathds{I}(t\geq t_0)\bm{F}\odot(\hat{\X}-\X_t)
\end{equation*}

We can rewrite the equation by shifting the term $\bm{F}\odot \X_t$ to the left, reformulating it in matrix form $\bm{F}_{diag}\X_t$, and obtaining $[\A(L)+\bm{F}_{diag}]\X_t = \bm{\nu} + \vectornoisevar_t + \mathds{I}(t\geq t_0)\bm{F}\odot\hat{\X}$. 
By defining $\A_\intervention(L):=\A(L)+\bm{F}_{diag}$ and $\tilde{\bm{F}}:= \bm{F}\odot\hat{\X}$, we obtain $\A_\intervention(L) \X_t = \bm{\nu} + \vectornoisevar_t + \mathds{I}(t\geq t_0)\tilde{\bm{F}}$, which is again an additive intervention, but on the intervened dynamic. For this reason, we can prove the case $\hat{\X}=\bm{0}$ and directly applying the result of additive intervention afterwards.
By applying $\gT$ (\cref{eq:t-trasformation}) and equilibrating, we obtain
\begin{equation*}
    \bm{Z}^\intervention_\infty = \A_\intervention(1)^{-1} \noisescm
\end{equation*}

Consider the analogue of forcing intervention over an \SCM{}:
\begin{equation*}
    \tilde{\X} = \tilde{\bm{A}} \tilde{\X} + \noisescm - \bm{F} \odot \tilde{\X}, \quad \text{where we fixed} \quad \hat{\X}=\bm{0}.
\end{equation*}

We can rewrite it as $[I-\tilde{\bm{A}}+\bm{F}_{diag}]\tilde{\X}=\noisescm$, that is $\A_\intervention(1)\tilde{\X} = \noisescm \Rightarrow \tilde{\X} = \A_\intervention(1)^{-1} \noisescm.\qed{}$

\subsection{Stability of forcing intervention is not guaranteed: an example}\label{subsec:stability_not_granted}

$\intervention_{f}$ perturbs the system dynamics by modifying the operator $\A(L)$. 
Specifically, by shifting the term $\bm{F} \odot \X_t$ to the left of the equation and rewriting it in matrix form as $\bm{F}_{diag}\X_t$, we obtain $\A_\intervention(L):=\A(L)+\bm{F}_{diag}$. 
Hence, the stability of the intervened system is not guaranteed, and it is necessary to verify \textit{that all the eigenvalues of} $\A_\intervention(L)$ \textit{are still inside the unit circle.}

When applying a forcing intervention, the intensity of $\bm{F}$ can play an important role. To show this, consider the toy model described by the $\VAR{}(1)$ equation
\begin{equation}\label{eq:pendulum_matrix}
    \begin{bmatrix}
        \Xcomp[1]\\\Xcomp[2]
    \end{bmatrix}_t
    = 
    \sqrt{2}\begin{bmatrix}
        0 & -0.5 \\
        0.5 & 1
    \end{bmatrix} 
    \begin{bmatrix}
        \Xcomp[1]\\\Xcomp[2]
    \end{bmatrix}_{t-1}
    + 
    \begin{bmatrix}
        \vectornoise^{(1)}\\\vectornoise^{(2)}
    \end{bmatrix}_t.
\end{equation}
the roots of $det(I-\A_1z)$ are $z_{1,2}=\sqrt{2}>1$, indicating that the system is stable. Applying $\intervention_f$ to component $\Xcomp[1]$, we need to examine
\begin{equation*}
    det\left(
    \begin{bmatrix}
        1+F & 0\\
        0 & 1
    \end{bmatrix}-
    \sqrt{2}\begin{bmatrix}
        0 & -0.5\\
        0.5 & 1
    \end{bmatrix}z
    \right)
\end{equation*}
which have roots $z_{1,2}=\sqrt{2}\left(1+F\pm \sqrt{F^2+F}\right)$. It is easy to verify that for $F=1$, one root is less than $1$, and thus the model is no longer stable.
On the other hand, interveening on $\Xcomp[2]$ preserves stability regardless of $F$ intensity; the same procedure yields $z_{1,2}=\sqrt{2}\left(1\pm\sqrt{-F}\right)$, so that $|z_{1,2}|\geq \sqrt{2}>1$ $\forall F>0$. 
\\

It is possible to identify sufficient conditions for $\dequation$ to be always stable under intervention. In particular, it is easy to see that if all matrices $\A_i$ of the model are strictly lower triangular, and thus $\graph_\dequation$ is acyclic, $\X_t$ has a finite Moving Average representation. This property is preserved by any intervention $\intervention_f$. Acyclicity can be weakened to allow \textit{self-loops}.

\paragraph{Stable \VAR{}s with at most self-loops are interventionally stable} Given a \VAR{} model such that all its matrices $\A_i$ are (non-strictly) lower triangular, the stability of the intervened system directly follows from the stability of the observational one. To verify this, note that the determinantal polynomial of the model $$det(I-\A_1z-\dots-\A_pz^p) \quad (\text{as in} \: \text{\cref{eq:determinantal}})$$ reduces to the product of diagonal terms since all matrices are triangular, and each i-th root satisfies the relation $p_i(z)=1-\A_1[i,i]z-\dots-\A_p[i,i]z^p=0$. Observing that $p_i(0)=1>0$ and considering the root with the smallest norm $z_{min}$, using the continuity of $p_i$, we can deduce that $p_i(z)>0 \: \forall z$ such that $|z|<|z_{min}|$. Intervening on component $i$ yields $\tilde{p_i}(z)=p_i(z)+F$, from which we can conclude $\tilde{p_i}(z)>F>0$ for all $z$ with norm lower than $z_{min}$, and in particular for all those within the unit circle.

\begin{figure}
    \centering
    \begin{subfigure}[t]{0.24\textwidth}
        \centering
    \begin{tikzpicture}[scale=1.1,
            transform shape,
            vertex/.style={circle, draw, minimum size=1.1cm, inner sep=0pt, font=\small}]
    \node[vertex] (x1) at (0, 2.5) {$X^{(1)}$};
    \node[vertex] (x2) at (2.5, 2.5) {$X^{(2)}$};
    
    \draw[edge] (x1) to [bend left=25] node[midway, above] {$\frac{\sqrt{2}}{2}$} (x2);
    \draw[edge] (x2) to [bend left=25] 
    node[midway, below] {$-\frac{\sqrt{2}}{2}$}
    (x1);
    \draw[edge] (x2) to [loop right] 
    node[midway, right] {$\sqrt{2}$}
    (x2);
    \end{tikzpicture}
    \end{subfigure}%
    \begin{subfigure}[t]{0.24\textwidth}
        \centering
        \includegraphics[height=1.2in]{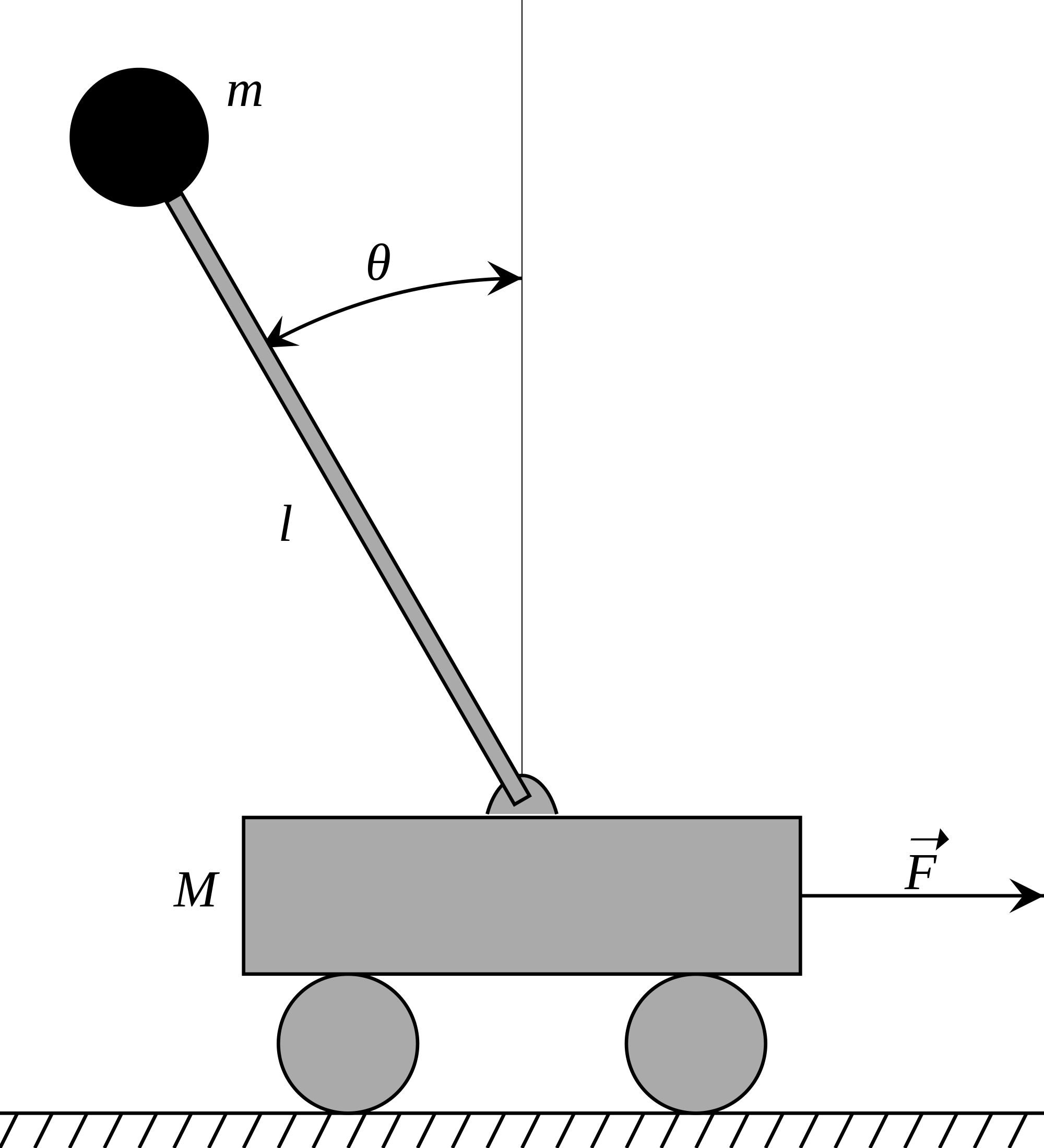}
    \end{subfigure}
    \caption{(left) A simple model that lacks interventional stability: \( \Xcomp[1] \) functions as a stabilizer for \( \Xcomp[2] \), which exhibits a divergent dynamic. Intervening on \( \Xcomp[1] \) disrupts the negative feedback loop, making the system unstable. (right) A concrete example of the model from control theory is the inverted pendulum: \( \Xcomp[1] \) is the position of the cart, while \( \Xcomp[2] \) is the angle \( \theta \); fixing the cart makes the system unstable, i.e., the pendulum falls. As a practical example, consider optimal control applications to nuclear reactors management~\citep{edwards1992robust}.}
    \label{fig:negativefeedback}
\end{figure}
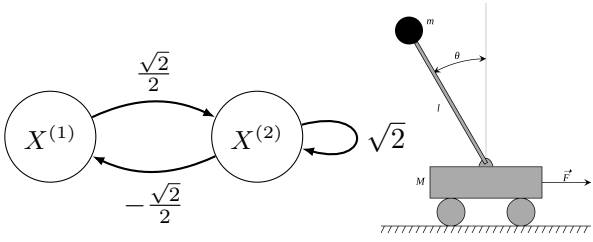


\section{Answering causal queries in practice}\label{app:queries_practice}
In this section, we report the actual execution of the procedures used in our work to calculate forecasting (\ref{app:forecasting}), interventional forecasting (\ref{app:interventional_forecast}), and retrospective counterfactuals (\ref{app:counterfactuals_retro}).

\subsection{Forecasting}\label{app:forecasting}
Forecasting is one of the main objectives of time series analysis in general~\citep{montgomery2015introduction}. We discuss predictors based on \VAR{} processes that minimize the Mean Square Error (MSE). 

\paragraph{Objective} Suppose we have a time series $\X_t$ up to time $t_0$ and we want to predict the state of the system after $h$ steps $\X_{t_0+h}$. We call a predictor for this quantity the $h$-steps predictor. In particular, the optimal predictor with respect to MSE is 
\begin{equation*}
    \expected{\X_{t_0+h}|\X_{\leq t_0}}.
\end{equation*}
The optimality of the expected value relies on the assumption that $\vectornoisevar_t$ in \cref{eq:VAR} is \textit{i.i.d. white noise}.
For a \VAR{}(p) as in \cref{eq:VAR}, the optimal $h$-step predictor can be evaluated by recursively applying
\begin{equation}\label{eq:recursive_expected_var}
\begin{split}
    &\expected{\X_{t+k}|\X_{\leq t}} =\bm{\nu} +\A_1\expected{\X_{t+k-1}|\X_{\leq t}} + \\
    &\cdots + \A_p \expected{\X_{t+k-p}|\X_{\leq t}},
\end{split}
\end{equation}
where $\expected{\X_{t'}|\X_{\leq t}} = \X_{t'}$ if $t' \leq t$. In other words, it is possible to forecast all subsequent values of $\X_{t_0}$ by applying the original equation of the \VAR{} without considering the noise $\vectornoisevar_t$. For example, considering the $1$-step predictor, we get
\begin{equation*}
    \expected{\X_{t_0+1}|\X_{\leq t_0}} = \bm{\nu} + \A_1 \X_{t_0} + \cdots + \A_p \X_{t_0+1-p}.
\end{equation*}
Similarly, by exploiting the possibility of recursively applying \cref{eq:recursive_expected_var}, we obtain the estimate of the forecast error covariance (or MSE) matrix as
\begin{equation*}
\Sigma_{\X}(k) = \sum_{i=0}^{h-1} \Phi_i \Sigma_{\vectornoisevar} \Phi'_i.
\end{equation*}
Note that each term in the sum is a positive definite matrix, and consequently, the MSE matrix is monotonically non-decreasing. $\Sigma_{\X}(k)$ approaches the covariance matrix of $\X_t$, $\Sigma_{\X}$, for $k\to\infty$. In fact, the optimal long-range forecast $h \to \infty$ is just the process mean. In other words, the past of the process contains no information on the development of the process in the distant future.

\subsection{Interventional Forecasting}\label{app:interventional_forecast}

Consider again a \VAR{}(p) model described by \cref{eq:VAR} and suppose an intervention is applied to the system at time $t_0$. We report the procedure for performing interventional forecasts for additive and forcing interventions separately.

\paragraph{Additive intervention}
The model equations are modified only by adding the term $\bm{F}$, consequently, \cref{eq:recursive_expected_var} simply becomes
\begin{equation}\label{eq:forecast_additive_intervention}
\begin{split}
    &\expected{\X_{t+k}|\X_{\leq t}} = \hat{\bm{\nu}}+ \A_1\expected{\X_{t+k-1}|\X_{\leq t}} +\\
    &\cdots + \A_p \expected{\X_{t+k-p}|\X_{\leq t}}, \quad \text{where} \quad  \hat{\bm{\nu}} = \bm{\nu}+\bm{F}
    \end{split}
\end{equation}
The MSE matrix remains unchanged, as both $\Sigma_{\vectornoisevar}$ and all the matrices (and consequently the matrices $\Phi_i$, see \cref{eq:recursive_calcultation_phi}) are not modified.

Alternatively, exploiting the linearity of the predictor, it is possible to directly estimate the expected causal effect at time $h$ through
\begin{equation*}
    \ce_{t_0+h} 
    = \sum_{l=0}^h \response_l \bm{F}, 
\end{equation*}
and add this quantity to the observational forecast. The resulting interventional forecast is equivalent.

\paragraph{Forcing intervention}
In this case, the intervention modifies the matrices $\A_i$ of the process, and the procedure is more complex. Let us proceed step by step. The intervened model is described by (see \cref{eq:do-like})

\begin{equation}\label{eq:forecast_forcing_intervention}
\begin{split}
    & \X_t = \bm{\nu} + \bm{F}\odot(\hat{\X}-\X_t) + \sum_{i=1}^p \A_i \X_{t-i} + \vectornoisevar_t \\
    \Rightarrow \quad & [I+\bm{F}_{diag}] \X_t = \bm{\nu} + \bm{F}\odot\hat{\X} + \sum_{i=1}^p \A_i \X_{t-i} + \vectornoisevar_t \\
    & \text{where} \quad \bm{F}_{diag} := I \cdot \bm{F} \\
    \Rightarrow \quad & [I+\bm{F}_{diag}] \X_t = \hat{\bm{\nu}} + \sum_{i=1}^p \A_i \X_{t-i} + \vectornoisevar_t \\
    & \text{where} \quad \hat{\bm{\nu}} := \bm{\nu} + \bm{F}\odot\hat{\X} \\
    \Rightarrow \quad & \X_t = [I+\bm{F}_{diag}]^{-1} \left(\hat{\bm{\nu}} + \sum_{i=1}^p \A_i \X_{t-i} + \vectornoisevar_t \right) \\
    \Rightarrow \quad & \X_t = \tilde{\bm{\nu}} + \sum_{i=1}^p \tilde{\A}_i \X_{t-i} + \tilde{\vectornoisevar}_t,
\end{split}
\end{equation}

where $\tilde{\bm{\nu}}, \tilde{\A}_i, \tilde{\vectornoisevar}_t$ are the results of left-multiplying their respective elements by $[I+\bm{F}_{diag}]^{-1}$.

The resulting equation has the form of a standard \VAR{}(p) model, so we can apply the forecasting exactly as in \cref{app:forecasting}. The same applies to the MSE matrix, calculated using the modified matrices $\Sigma_{\tilde{\vectornoisevar}}$ and $\tilde{\Phi}_i$.

Unlike the additive intervention, there is no way to directly calculate the causal effect. The estimate is therefore obtained through the difference.
\begin{equation*}
    \ce{}_{t_0+h} = \expected{\tilde{\X}_{t_0+h}|\X_{\leq t}} - \expected{\X_{t_0+h}|\X_{\leq t}},
\end{equation*}
that is \textit{the difference between the interventional forecast and the observational one}.

\subsection{Retrospective Counterfactuals}\label{app:counterfactuals_retro}
In the \textit{ladder of causality}~\citep{pearl2009causality,BareinboimCII22}, counterfactual statements belong to the last level, distinct from the interventional level, and a mature causal model should be able to address both. Embracing this perspective, we show how counterfactual queries can be formulated for causal inference over time. The procedure follows the same steps used in the \SCM{} framework.
\begin{itemize}
\item \textbf{Abduction} Based on the observed trajectory $\X_t$, we estimate the values of $\vectornoisevar_t$ by calculating the residuals $\hat{\vectornoisevar}_t = \X_t - \expected{\X_t|\X_{<t}}$. For a \VAR{}(p) model, in particular, $\expected{\X_t|\X_{<t}} = \A_1 \X_{t-1} + \cdots + \A_p \X_{t-p}$.
\item \textbf{Action} We apply a specific intervention $\intervention$ by modifying the model's equations.
\item \textbf{Prediction} We simulate the evolution of the new process $\X^{\intervention}_t$ using $\hat{\vectornoisevar}_t$ in the equation.
\end{itemize}

Let $\X_t$ be a time series generated by a \VAR{}(p) model. We want to evaluate what the trajectory of $\X_t$ would have been up to the present time $t_1$ assuming a specific intervention was applied at a particular past time $t_0$. We proceed as follows:
\begin{enumerate}
    \item Perform the abduction step and retrieve the residuals $\{\hat{\vectornoisevar}_t\}_{t=t_0}^{t_1}$.
    \item Apply the intervention to the model as in \cref{app:interventional_forecast}.
    \item Simulate the counterfactual trajectory from $\X_{t_0}$ to $\X_{t_1}$ through the recursive application of
    \begin{equation*}
        \X_t = \hat{\bm{\nu}} + \A_1 \X_{t-1} + \dots + \A_p \X_{t-p} + \hat{\vectornoisevar}_t.
    \end{equation*}
    where $\hat{\bm{\nu}}$ is defined as in \cref{eq:forecast_additive_intervention} or \cref{eq:forecast_forcing_intervention}, depending on the type of intervention. Note that \textit{if using forcing intervention}, the whole term on the right must be left-multiplied by $[I+\bm{F}_{diag}]^{-1}$ before performing the simulation.
\end{enumerate} 
\paragraph{Observation} Performing retrospective counterfactual without applying any intervention results in the perfect recovery of the factual trajectory, so that the difference with the interventional one can be interpreted in the same way as in \ref{app:interventional_forecast}, i.e. as the causal effect over time along the trajectory.

\section{Experimental details and extra results}
\label{app:extra_results}

\definecolor{myorange}{HTML}{E69F00}
\definecolor{mypurple}{HTML}{8B008B}

In this section, we expand the description of the experimental section from \cref{sec:experiments}, and provide the reader with additional results in the following subsections.
First, we outline the details that are consistent across all experiments and then explore the specifics of each experiment in their respective subsections.

\paragraph{Hardware}
To run all experiments, we used a machine with 18 cores, 36 threads, equipped with Intel(R) Core (TM) i9-10980XE CPU @ 3.00GHz, OS Ubuntu 20.04.3, programming language Python 3.12.0. 

\paragraph{Datasets}\label{app:dataset}
This section provides all the information on datasets used in the empirical evaluation of~\cref{sec:experiments} of the main manuscript and the following subsections. 

\vspace{0.2cm}

\textbf{German} We aimed to capture relationships between the variables that appeared intuitive to us and, to a certain extent, reflected a real-world loan approval scenario. Therefore, we selected the following variables from the German Credit dataset\footnote{\url{https://www.kaggle.com/datasets/uciml/german-credit}}: 
\textit{Expertise} ($E$), 
\textit{Responsibility} ($R$), 
\textit{Loan Amount} ($L$), 
\textit{Loan Duration} ($D$), 
\textit{Income} ($I$), 
\textit{Savings} ($S$) 
and \textit{Credit Score} ($C$). 
We have indexed the variables from $0$ to $6$, following the sequence in which they are mentioned.
The corresponding causal graph is designed as in~\cref{fig:german_graph}. 
%
%
%
%
To generate a $7$-dimensional vector-valued stationary time series, we simulate a stable \svar{}$(4)$ process, considering four lags to be a reasonable timeframe for conducting our analysis.
In modeling the autoregressive behavior of socio-economic variables, we use different matrices tailored to capture temporal dependencies in the data. Below, we provide interpretations for such matrices and specify all the non-zero coefficients:
\begin{itemize}
    \item $A_0$ denotes the instantaneous impact between variables and affects only those nominal features (theoretical value of a good or financial instrument defined by a consensus or standard), e.g., the \textit{Credit Score}. 
    This is highlighted because the tangible attributes of an applicant require a certain period to change. In our model, we set $A_0[3, 6] = 0.5$; $A_0 [4, 6] = -0.3$; $A_0 [5, 6] = -0.5$. 
    Furthermore, all diagonal entries of $A_0$ are set to $1$;
    \item $A_1$ has diagonal values that are all $0.95$ except for the last one (ie., \textit{Credit Score)}, which is $0$. This matrix gives the dynamic system a memory function.
    \item $A_2$ includes interactions among variables we classify as rapid. These effects are considered short-term and are typically the most direct and intense, as the system's memory extends only one step back in time. 
    In our model, we set $A_2 [1, 4] = 0.3$; $A_2 [2, 6] = 0.5$; $A_2 [4, 5] = 0.2$;
    \item $A_3$ contains medium-term effects, illustrating how conditions from three steps back affect the current state of the system. In our model, we set $A_3 [2, 3] = 0.5$;
    \item $A_4$ reflects long-term relationships within the system. In our model, we set $A_4 [0, 1] = 0.3$; $A_4 [0, 4] = 0.8$.
\end{itemize}

\textbf{Pendulum} For the description of the model see~\cref{subsec:stability_not_granted}. To generate a $2$-dimensional vector-valued stationary time series, we simulate a stable \SVAR{}(1) process. In this case, we set $A_0$ as an identity matrix, while $A_1$ is set as in~\cref{eq:pendulum_matrix}. 

\vspace{0.15cm}

\textbf{Census} We consider open data extracted from the United States Census Bureau\footnote{\url{https://www.census.gov}}.
This source provides a panel dataset spanning yearly demographic characteristics for $227$ countries, starting from $1950$ up to $2023$.
We conducted a preprocessing procedure to align the data with our experimental requisites. Specifically, we selected the $50$ countries with high income\footnote{\url{https://blogs.worldbank.org/en/opendata/world-bank-country-classifications-by-income-level-for-2024-2025}}. 
Since most countries had missing values in the initial decades of data collection, we decided to start the dataset from $1991$, considering the $32$-year period sufficient for our work. The total number of instances is therefore $1600$.
Then, we chose only the essential variables for our analysis, which led to the construction of the ground truth causal graph. The chosen variables include: \textit{Total Population} for three age groups namely $0-14$, $15-64$, and $65-99$, \textit{Net Migration}, \textit{Total Births}, \textit{Total Deaths}. The assumed causal graph is shown in~\cref{fig:census_graph}. 
To adapt the data to the autoregressive model, we modified the assignment of values for \textit{Net Migration}, \textit{Total Births}, and \textit{Total Deaths} by shifting them to the previous year's values. 

\paragraph{Metrics} 
We measure the discrepancy between the \textit{h}-step forecast $\hat{\X}_{t+h}|\X_{<t}$ and the true value $\X_{t+h}$ on the test set using as metrics Root Mean Square Error (RMSE) and
Symmetric Mean Absolute Percentage Error (SMAPE)~\citep{montgomery2015introduction}.
We report scores focusing on the target variables (i.e., \textit{Credit Score} for German, $X^{(1)}$ for Pendulum, and age groups for Census).

\paragraph{Training and evaluation methodology}
We generated synthetic time-series datasets with varying training samples. We used $300$ validation and $2400$ test samples for German while $100$ validation and $2200$ test samples for Pendulum.

We implemented DLinear, TiDe, and TSMixer models using the Darts library~\cite{herzen2022darts}. Each model was trained for $200$ epochs, with final evaluations performed on the test set. 
We employed an early stopping mechanism to prevent overfitting, monitoring the validation loss with patience set to $10$ and a minimum delta of $1e-5$. 
We used the Adam optimizer~\cite{KingmaB14} with an initial learning rate of $1e-3$ and an exponential learning rate scheduler\footnote{\url{https://pytorch.org/docs/stable/generated/torch.optim.lr_scheduler.ExponentialLR.html}} with a decay rate of $0.99$.
For hyperparameter tuning, we always select the best hyperparameter combination according to validation loss, reporting results from the test dataset in the manuscript. 
We applied Mean Squared Error as the loss function for DLinear and quantile regression as the likelihood function for TiDE and TSMixer.

For the Census dataset, we constrained the fitting of the \VAR{} framework by setting the matrix coefficients without dependencies in the causal graph to zero.

All experiments were repeated $10$ times, with results reported as averages and standard deviations.

\begin{table}[t]
    \centering
    \caption{\textbf{Observational forecasting.} MAE scores in~\cref{tab:comparative_forecasting}. Here, we also report the standard deviation in subscript.}\label{tab:observational_forecasting_MAE}
    \setlength{\tabcolsep}{0.8pt}
    \footnotesize
    \renewcommand{\arraystretch}{1.18}
    \begin{tabular}{lcccccccc}
    \toprule
    & & & \multicolumn{5}{c}{MAE} \\
    \cmidrule(lr){4-8}
    Dataset & Size & Horizon & Oracle & VAR & DLinear & TiDE & TSMixer \\
    \midrule
    German & \multirow{4}{*}{\centering100} & \multirow{2}{*}{\centering1} & $\mathtt{.004}$ & $\mathbf{.008}_{.004}$ & $.009_{.004}$ & $.011_{.000}$ & $.014_{.001}$ \\
Pendulum & & & $\mathtt{.042}$ & $\mathbf{.043}_{.001}$ & $\mathbf{.043}_{.001}$ & $.218_{.000}$ & $.217_{.000}$ \\
\cmidrule(lr){3-8}
German & & \multirow{2}{*}{\centering10} & $\mathtt{.014}$ & $\mathbf{.055}_{.030}$ & $\mathbf{.055}_{.027}$ & $.094_{.003}$ & $.139_{.005}$ \\
Pendulum & & & $\mathtt{.399}$ & $\mathbf{.420}_{.018}$ & $.440_{.009}$ & $1.43_{.002}$ & $1.43_{.001}$ \\
\midrule
German & \multirow{4}{*}{\centering500} & \multirow{2}{*}{\centering1} & $\mathtt{.004}$ & $\mathbf{.004}_{.000}$ & $\mathbf{.004}_{.000}$ & $.011_{.000}$ & $.014_{.000}$ \\
Pendulum & & & $\mathtt{.042}$ & $\mathbf{.042}_{.000}$ & $\mathbf{
.042}_{.000}$ & $.218_{.001}$ & $.217_{.000}$ \\
\cmidrule(lr){3-8}
German & & \multirow{2}{*}{\centering10} & $\mathtt{.014}$ & $\mathbf{.015}_{.001}$ & $\mathbf{.015}_{.000}$ & $.093_{.002}$ & $.135_{.004}$ \\
Pendulum & & & $\mathtt{.399}$ & $\mathbf{.401}_{.002}$ & $.405_{.010}$ & $1.43_{.003}$ & $1.43_{.001}$ \\
\bottomrule
    \end{tabular}
\end{table}

\subsection{Observational Forecasting}\label{app:obs_forecasting}
\paragraph{How does the VAR performance compare with SOTA
models for forecasting multivariate time series?}
\cref{tab:observational_forecasting_MAE} presents results that
mirror those in~\cref{tab:comparative_forecasting} with standard deviation in subscript. 
In addition to the results from~\cref{subsec:observational_forecasting}, \cref{tab:observational_forecasting_combined} provides the forecasting accuracy, measured using RMSE and SMAPE scores, across different data sizes and forecast horizons for German and Pendulum datasets. 
The patterns observed here are consistent with those in the main text.
\VAR{} continues to lead in performance, often aligning closely with the Oracle. DLinear remains competitive for shorter forecast horizons thanks to its linear structure, occasionally surpassing \VAR{} by a small margin.

\begin{table*}[ht!]
    \centering
    \caption{\textbf{Observational forecasting.} RMSE and SMAPE scores (\emph{lower is better}) for \VAR{}, DLinear~\citep{zeng2023dlinear}, TiDE~\cite{das2023tide} and TSMixer~\cite{chen2023tsmixer}, benchmarked against the oracle forecaster. Results averaged over ten runs, with standard deviation in subscript.}
    \label{tab:observational_forecasting_combined}
    \setlength{\tabcolsep}{5pt}
    \footnotesize
    \renewcommand{\arraystretch}{1.18}
    \begin{tabular}{lcccccccccccc}
    \toprule
    & & & \multicolumn{5}{c}{RMSE} & \multicolumn{5}{c}{SMAPE} \\
    \cmidrule(lr){4-8} \cmidrule(lr){9-13}
    Dataset & Size & Horizon & Oracle & VAR & DLinear & TiDE & TSMixer & Oracle & VAR & DLinear & TiDE & TSMixer \\
    \midrule
    German & \multirow{4}{*}{\centering100} & \multirow{2}{*}{\centering1} & $\mathtt{.005}$ & $\mathbf{.010}_{.004}$ & $.011_{.005}$ & $.063_{.009}$ & $.071_{.002}$ & $\mathtt{2.01}$ & $\mathbf{3.81}_{1.44}$ & $4.12_{1.56}$ & $14.6_{2.16}$ & $20.4_{0.33}$ \\
    Pendulum & & & $\mathtt{.053}$ & $\mathbf{.053}_{.001}$ & $.054_{.001}$ & $.333_{.033}$ & $.264_{.000}$ & $\mathtt{15.7}$ & $\mathbf{15.8}_{0.22}$ & $15.8_{0.23}$ & $62.0_{3.73}$ & $54.2_{0.00}$ \\
    \cmidrule(lr){3-13}
    German & & \multirow{2}{*}{\centering10} & $\mathtt{.018}$ & $.069_{.038}$ & $\mathbf{.068}_{.033}$ & $.306_{.053}$ & $.284_{.008}$ & $\mathtt{7.33}$ & $17.2_{6.14}$ & $\mathbf{17.0}_{5.36}$ & $53.7_{3.75}$ & $58.8_{0.46}$ \\
    Pendulum & & & $\mathtt{.494}$ & $\mathbf{.520}_{.022}$ & $.545_{.013}$ & $1.89_{.069}$ & $1.74_{.000}$ & $\mathtt{82.7}$ & $\mathbf{87.2}_{4.61}$ & $88.9_{3.21}$ & $175_{2.32}$ & $180_{0.03}$ \\
    \midrule
    German & \multirow{4}{*}{\centering500} & \multirow{2}{*}{\centering1} & $\mathtt{.005}$ & $\mathbf{.005}_{.000}$ & $\mathbf{.005}_{.000}$ & $.037_{.003}$ & $.074_{.002}$ & $\mathtt{2.01}$ & $2.15_{0.08}$ & $\mathbf{2.13}_{0.08}$ & $8.45_{0.46}$ & $17.7_{0.44}$ \\
    Pendulum & & & $\mathtt{.053}$ & $\mathbf{.053}_{.000}$ & $\mathbf{.053}_{.000}$ & $.352_{.014}$ & $.264_{.000}$ & $\mathtt{15.7}$ & $\mathbf{15.7}_{0.02}$ & $\mathbf{15.7}_{0.03}$ & $63.8_{1.72}$ & $54.2_{0.00}$ \\
    \cmidrule(lr){3-13}
    German & & \multirow{2}{*}{\centering10} & $\mathtt{.018}$ & $\mathbf{.019}_{.001}$ & $\mathbf{.019}_{.001}$ & $.200_{.014}$ & $.320_{.008}$ & $\mathtt{7.33}$ & $\mathbf{7.55}_{0.19}$ & $7.68_{0.20}$ & $40.9_{1.63}$ & $56.5_{0.35}$ \\
    Pendulum & & & $\mathtt{.494}$ & $\mathbf{.496}_{.002}$ & $.502_{.011}$ & $1.93_{.030}$ & $1.74_{.000}$ & $\mathtt{82.7}$ & $\mathbf{83.3}_{0.75}$ & $83.8_{1.08}$ & $174_{0.70}$ & $180_{0.04}$ \\
    \midrule
    German & \multirow{4}{*}{\centering1000} & \multirow{2}{*}{\centering1} & $\mathtt{.005}$ & $\mathbf{.005}_{.000}$ & $\mathbf{.005}_{.000}$ & $.025_{.001}$ & $.051_{.002}$ & $\mathtt{2.01}$ & $2.09_{0.03}$ & $\mathbf{2.07}_{0.04}$ & $6.57_{0.21}$ & $12.4_{0.41}$ \\
    Pendulum & & & $\mathtt{.053}$ & $\mathbf{.053}_{.000}$ & $\mathbf{.053}_{.000}$ & $.373_{.018}$ & $.264_{.000}$ & $\mathtt{15.7}$ & $\mathbf{15.7}_{0.02}$ & $\mathbf{15.7}_{0.03}$ & $65.9_{1.89}$ & $54.1_{0.01}$ \\
    \cmidrule(lr){3-13}
    German & & \multirow{2}{*}{\centering10} & $\mathtt{.018}$ & $\mathbf{.018}_{.000}$ & $\mathbf{.018}_{.000}$ & $.161_{.006}$ & $.264_{.007}$ & $\mathtt{7.33}$ & $\mathbf{7.38}_{0.09}$ & $7.43_{0.13}$ & $34.2_{0.66}$ & $51.8_{0.84}$ \\
    Pendulum & & & $\mathtt{.494}$ & $\mathbf{.495}_{.001}$ & $.498_{.006}$ & $1.97_{.037}$ & $1.74_{.000}$ & $\mathtt{82.7}$ & $\mathbf{82.8}_{0.54}$ & $83.2_{0.66}$ & $173_{0.90}$ & $180_{0.01}$ \\
    \bottomrule
    \end{tabular}
\end{table*}
\begin{table*}[ht!]
    \centering
    \caption{\textbf{Observational forecasting.} RMSE and SMAPE scores (\emph{lower is better}) for \VAR{}, DLinear, TiDE and TSMixer on the Census dataset.}
    \label{tab:census_forecasting}
    \setlength{\tabcolsep}{3pt}
    \footnotesize
    \renewcommand{\arraystretch}{1.18}
    \begin{tabular}{lcccccccccc}
    \toprule
    & & & \multicolumn{4}{c}{RMSE} & \multicolumn{4}{c}{SMAPE} \\
    \cmidrule(lr){4-7} \cmidrule(lr){8-11}
    Dataset & Size & Horizon & VAR & DLinear & TiDE & TSMixer & VAR & DLinear & TiDE & TSMixer \\
    \midrule
    \multirow{2}{*}{\centering Census} & \multirow{2}{*}{\centering50$\times$32} & \multirow{1}{*}{\centering1} & $\mathbf{.002}$ & $.007$ & $\mathbf{.002}$ & $.009$ & $.409$ & $1.76$ & $\mathbf{.401}$ & $1.79$ \\
    \cmidrule(lr){3-11}
     & & \multirow{1}{*}{\centering5} & $.018$ & $.026$ & $\mathbf{.013}$ & $.024$ & $3.25$ & $6.36$ & $\mathbf{2.92}$ & $6.28$ \\
    \bottomrule
    \end{tabular}
\end{table*}

In contrast, TSMixer and TiDE generally underperform in comparison, with their scores consistently following behind \VAR{} and DLinear. The challenging nature of the Pendulum dataset is apparent, with all models (including Oracle) showing greater difficulty in maintaining the accuracy as for German.
\cref{tab:census_forecasting} presents the results for the Census dataset.  
The performance gap between the models narrows. For $1$-step forecasts, both \VAR{} and TiDE achieve similar, strong results, while for $5$-step forecasts, TiDE slightly outperforms \VAR{} in both metrics.

\subsection{Interventional Forecasting}\label{app:int_forecasting}

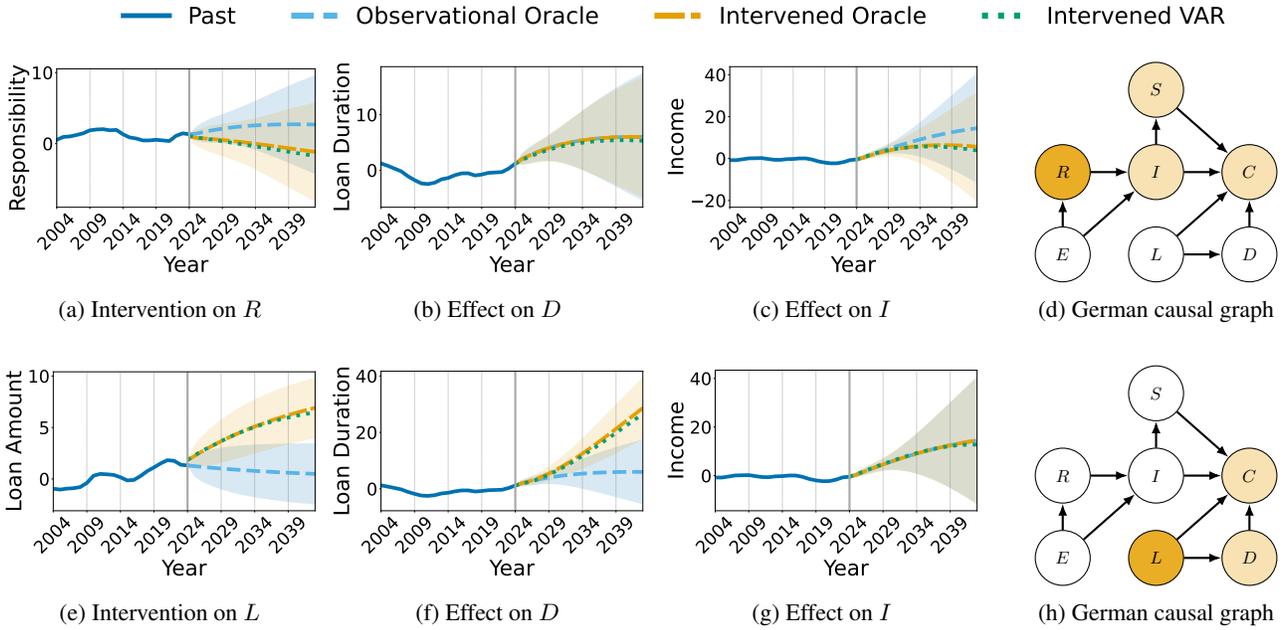
\begin{figure*}[t!]
  \centering
  \includegraphics[width=0.98\textwidth]{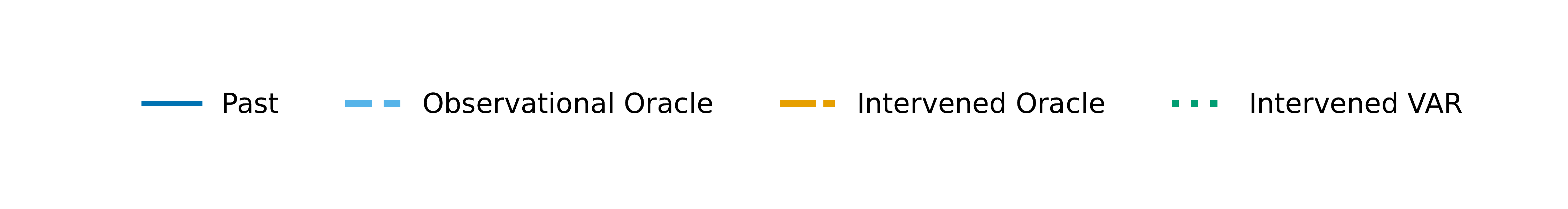}
  \begin{minipage}[t]{0.49\textwidth}
    \centering
    \begin{minipage}{\textwidth}
    \vspace{-0.6cm} 
    \begin{subfigure}[b]{0.49\textwidth}
      \centering
      \includegraphics[width=\textwidth]{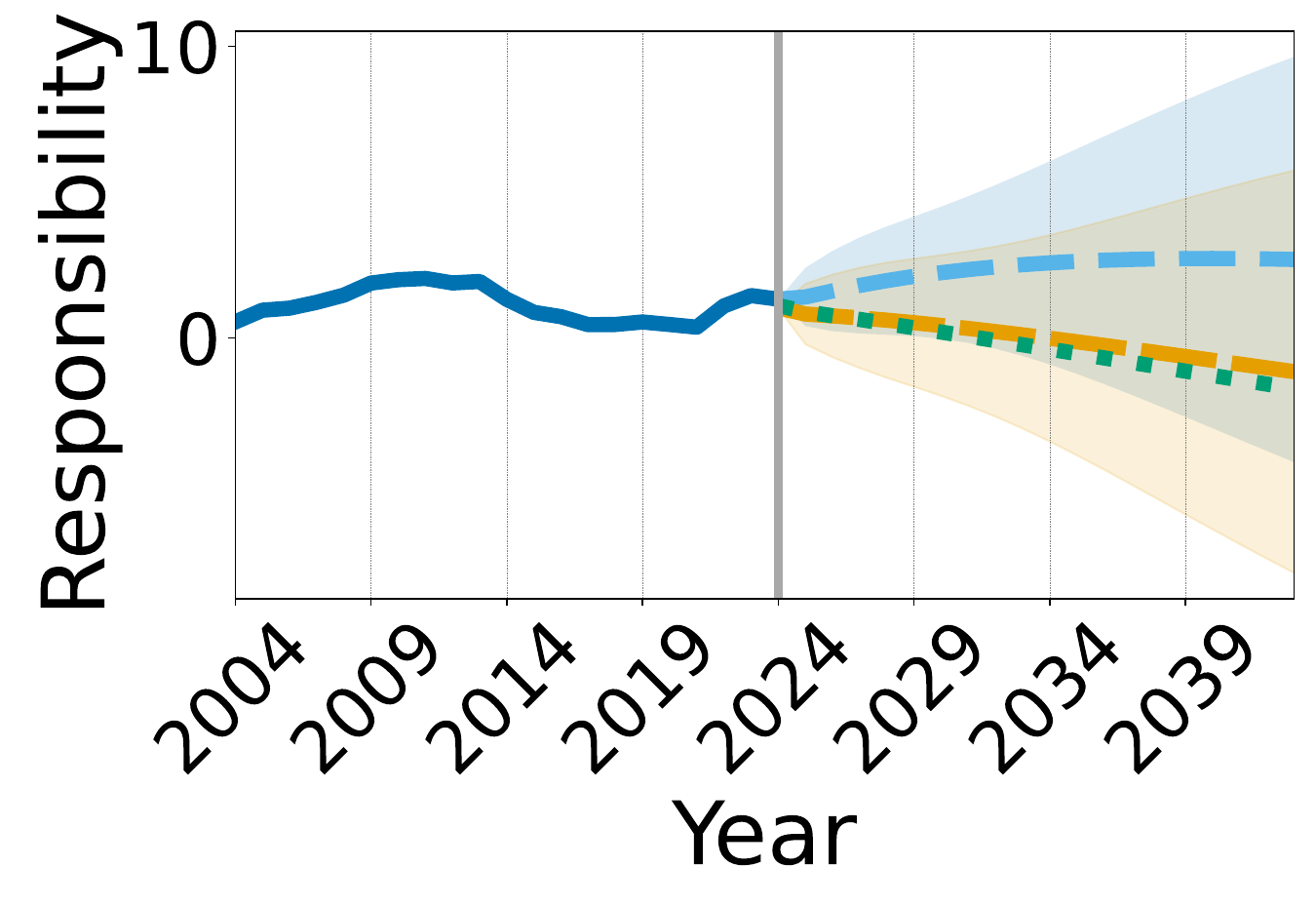}
      \caption{Intervention on $R$}
    \end{subfigure}
    \begin{subfigure}[b]{0.49\textwidth}
      \centering
      \includegraphics[width=\textwidth]{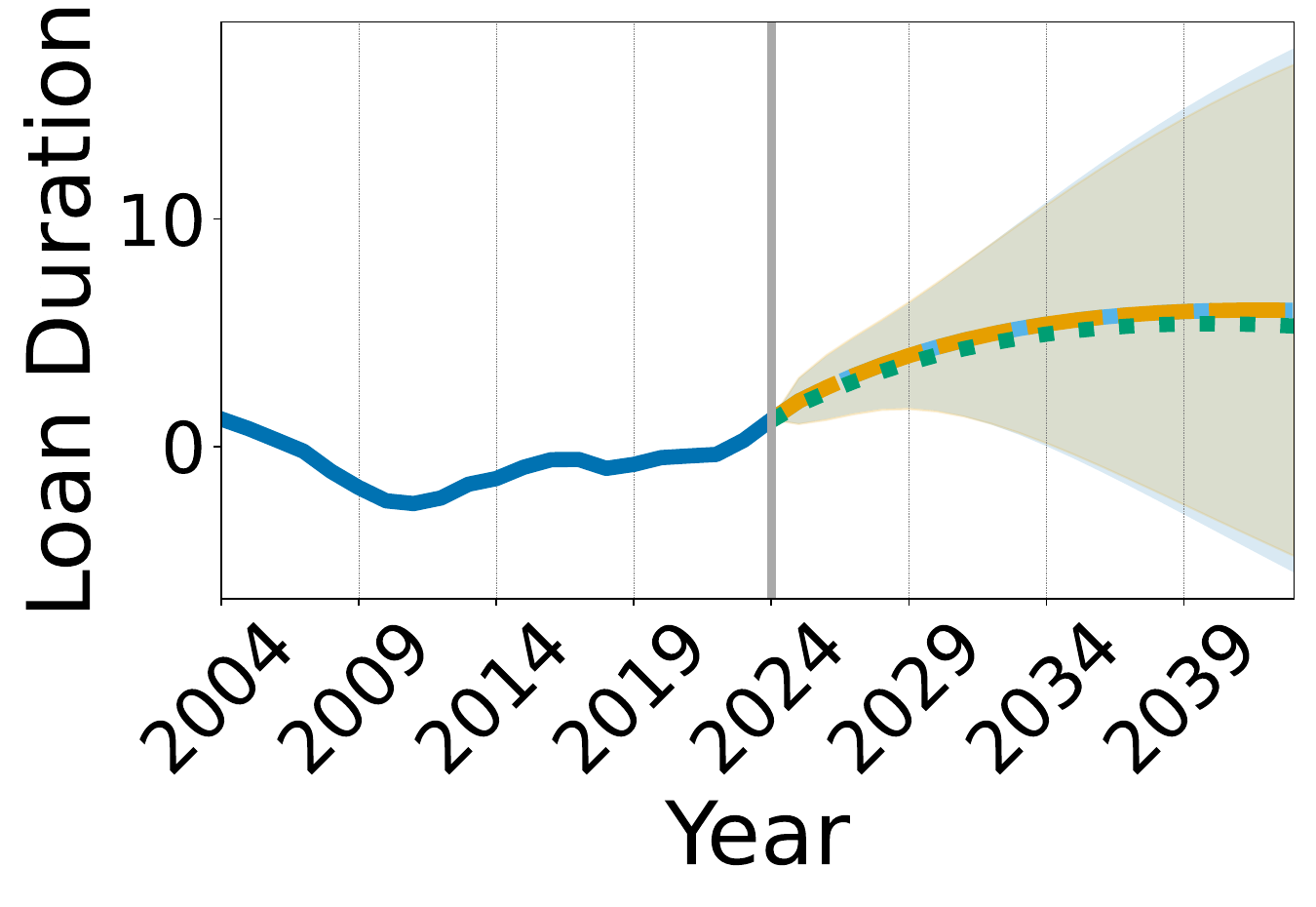}
      \caption{\centering Effect on $D$}
    \end{subfigure}
    \end{minipage}
  \end{minipage}
  \begin{minipage}[t]{0.49\textwidth}
    \centering
    \begin{minipage}{\textwidth}
    \vspace{-0.6cm} 
    \begin{subfigure}[b]{0.49\textwidth}
      \centering
      \includegraphics[width=\textwidth]{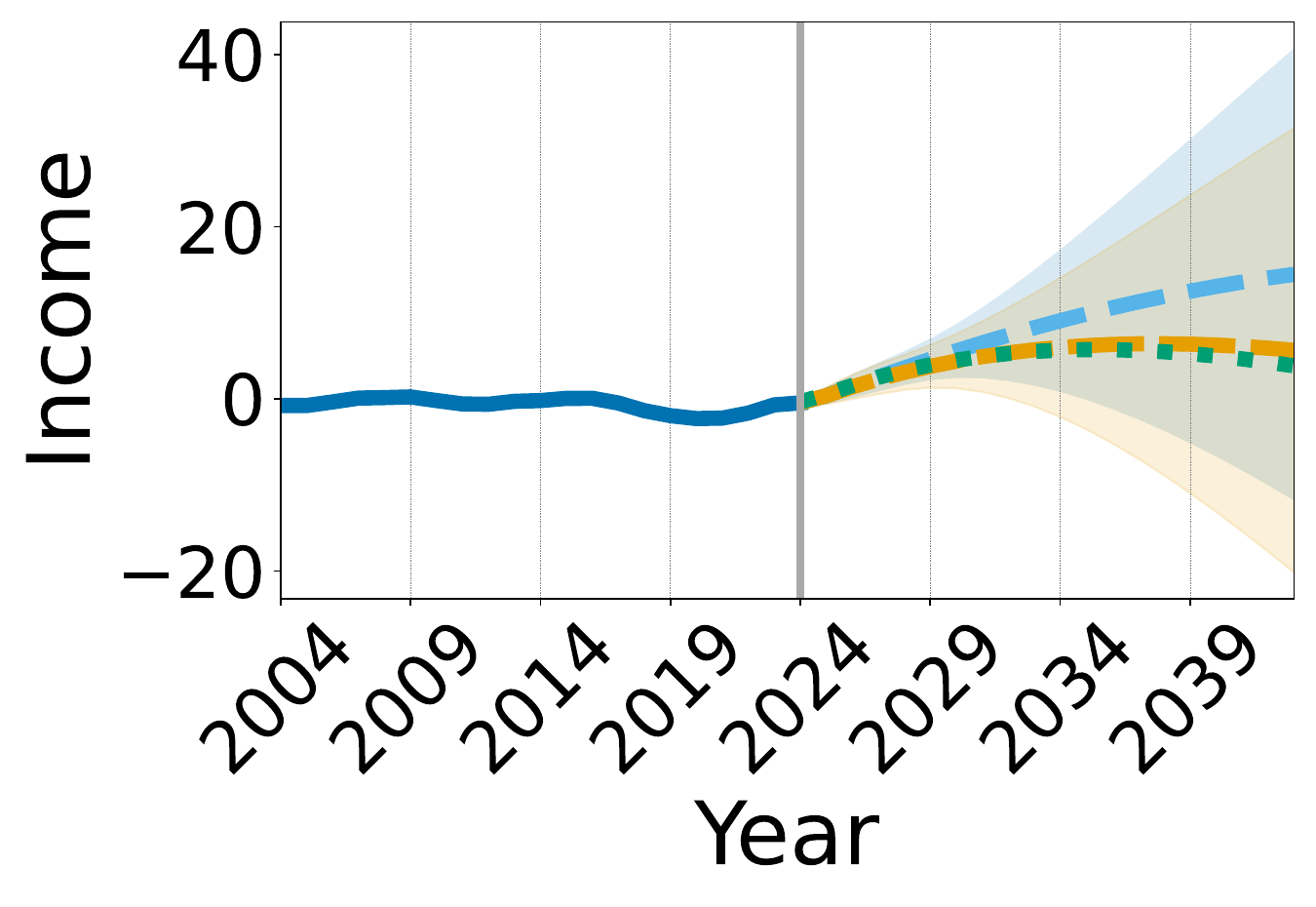}
      \caption{\centering Effect on $I$}
    \end{subfigure}
    \hfill
    \begin{subfigure}[b]{0.49\textwidth}
      \centering
      \begin{tikzpicture}[
            scale=0.73,
            transform shape,
            vertex/.style={circle, draw, minimum size=1cm, inner sep=0pt, font=\footnotesize}
        ]
        \node[vertex, fill=myorange!85!white] (r) at (0,1.5) {$R$};
        \node[vertex] (e) at (0,0) {$E$};
        \node[vertex] (l) at (1.7,0) {$L$};
        \node[vertex] (d) at (3.4,0) {$D$};
        \node[vertex, fill=myorange!30!white] (i) at (1.7,1.5) {$I$};
        \node[vertex, fill=myorange!30!white] (s) at (1.7,3) {$S$};
        \node[vertex, fill=myorange!30!white] (y) at (3.4,1.5) {$C$};
        
        \draw[edge] (e) -- (r);
        \draw[edge] (l) -- (d);
        \draw[edge] (r) -- (i);
        \draw[edge] (e) -- (i);
        \draw[edge] (i) -- (s);
        \draw[edge] (l) -- (y);
        \draw[edge] (d) -- (y);
        \draw[edge] (i) -- (y);
        \draw[edge] (s) -- (y);
        \end{tikzpicture}
        \caption{German causal graph}
    \end{subfigure}
     \end{minipage}
  \end{minipage} 
  \centering
  \begin{minipage}[t]{0.49\textwidth}
    \centering
    \begin{minipage}{\textwidth}
    \vspace{0.5cm} 
    \begin{subfigure}[b]{0.49\textwidth}
      \centering
      \includegraphics[width=\textwidth]{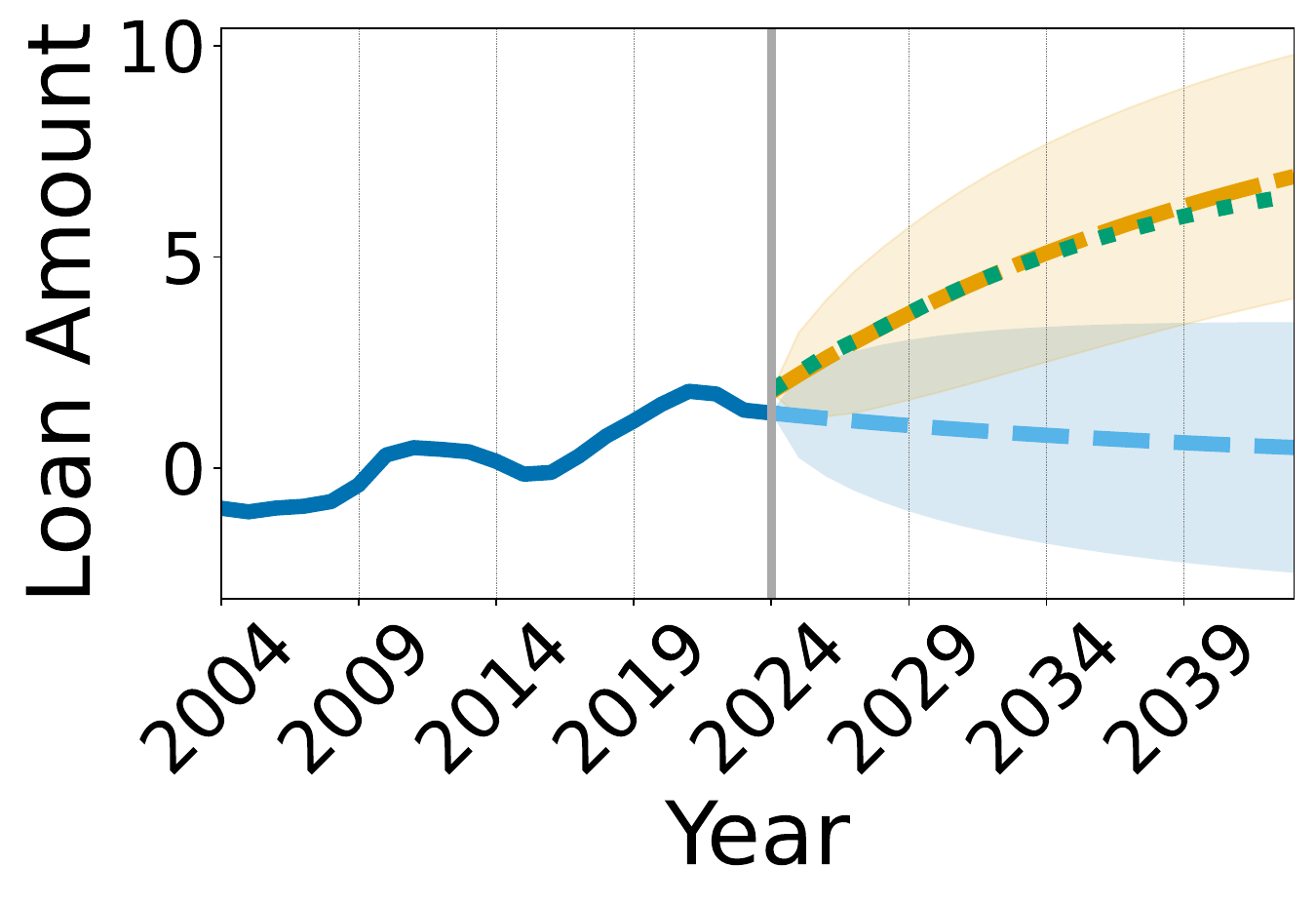}
      \caption{Intervention on $L$}
    \end{subfigure}
    \begin{subfigure}[b]{0.49\textwidth}
      \centering
      \includegraphics[width=\textwidth]{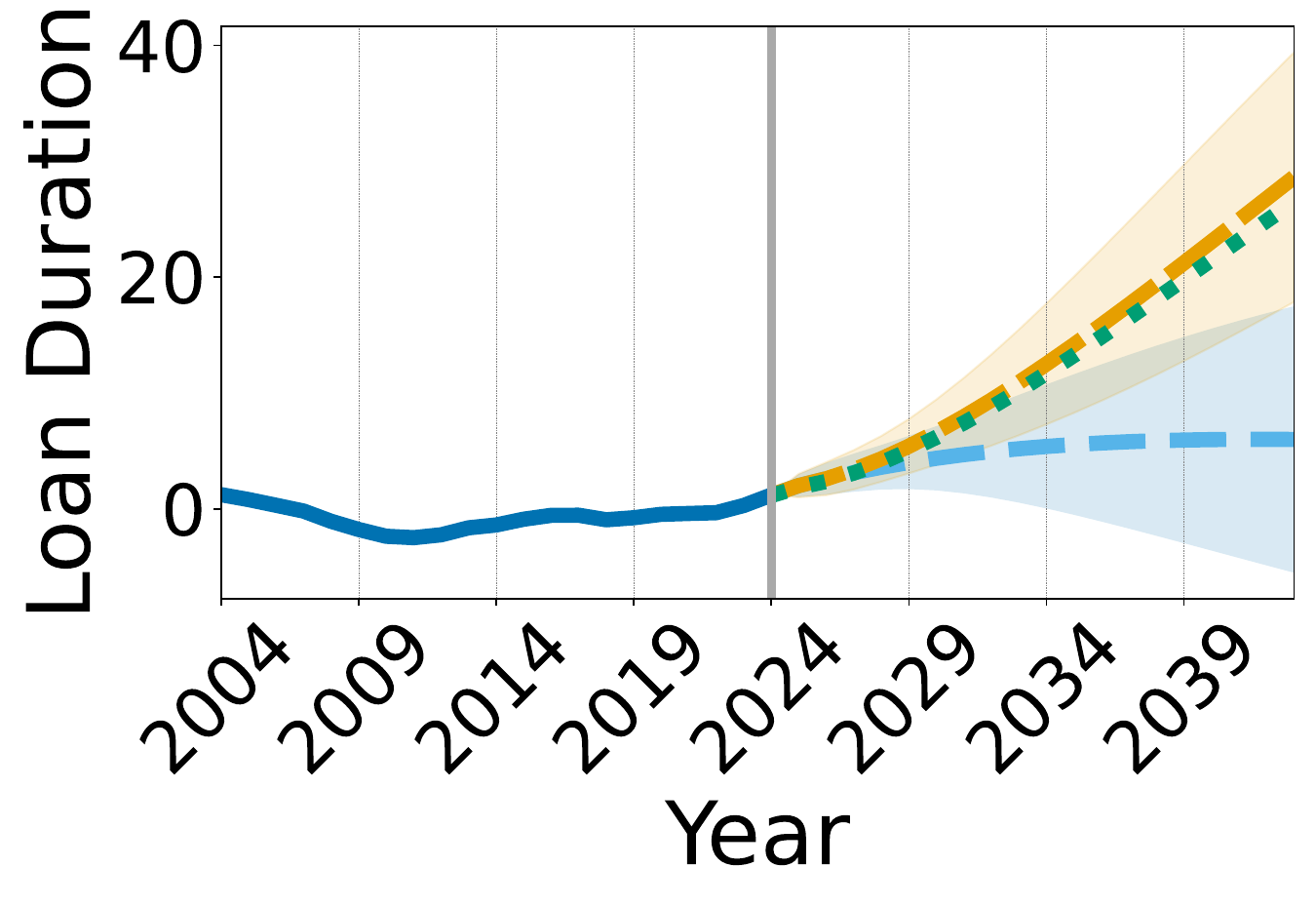}
      \caption{\centering Effect on $D$}
    \end{subfigure}
    \end{minipage}
  \end{minipage}
  \begin{minipage}[t]{0.49\textwidth}
    \centering
    \begin{minipage}{\textwidth}
    \vspace{0.5cm} 
    \begin{subfigure}[b]{0.49\textwidth}
      \centering
      \includegraphics[width=\textwidth]{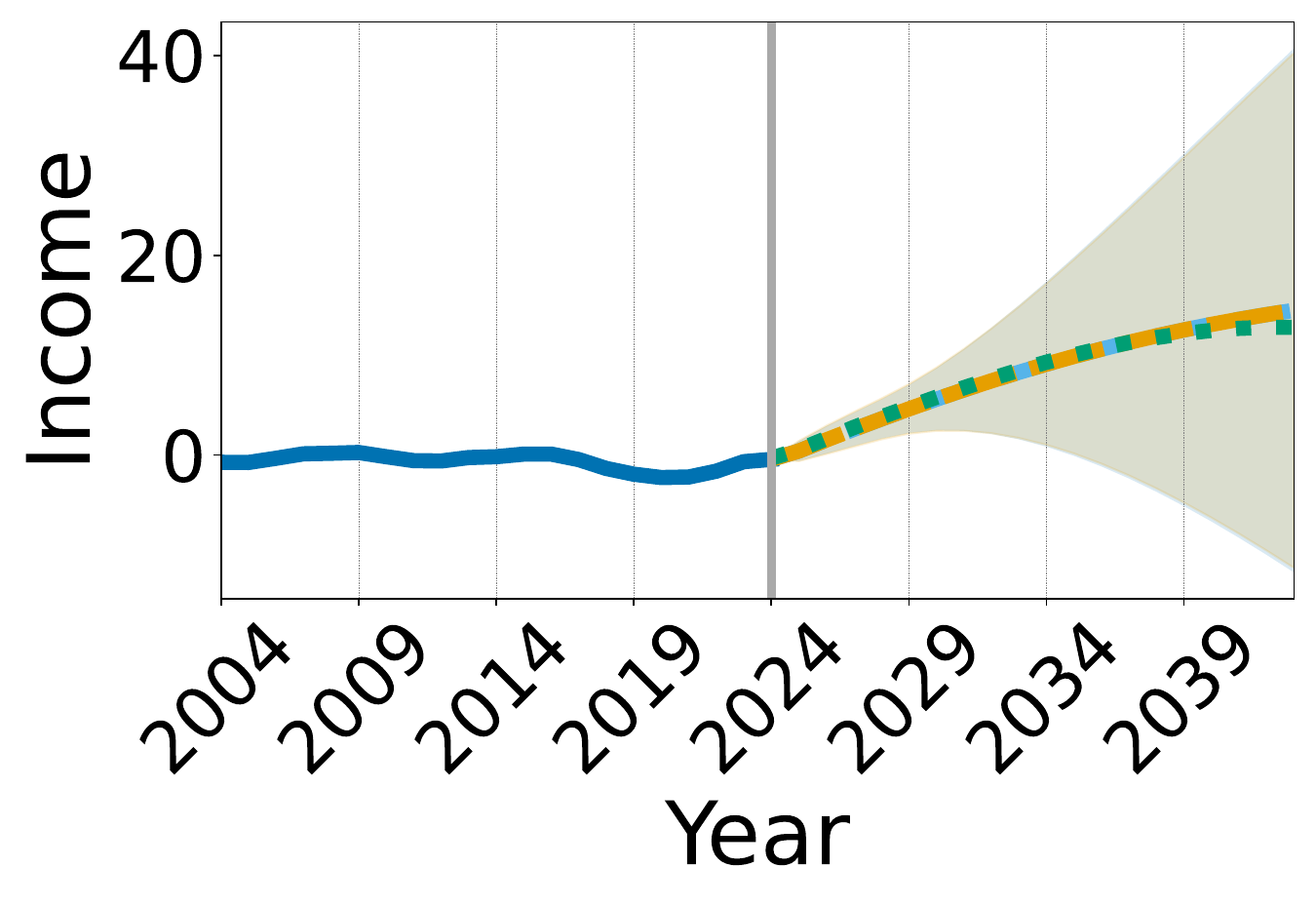}
      \caption{\centering Effect on $I$}
    \end{subfigure}
    \hfill
    \begin{subfigure}[b]{0.49\textwidth}
      \centering
      \begin{tikzpicture}[
            scale=0.73,
            transform shape,
            vertex/.style={circle, draw, minimum size=1cm, inner sep=0pt, font=\footnotesize}
        ]
        \node[vertex] (r) at (0,1.5) {$R$};
        \node[vertex] (e) at (0,0) {$E$};
        \node[vertex, fill=myorange!85!white] (l) at (1.7,0) {$L$};
        \node[vertex, fill=myorange!30!white] (d) at (3.4,0) {$D$};
        \node[vertex] (i) at (1.7,1.5) {$I$};
        \node[vertex] (s) at (1.7,3) {$S$};
        \node[vertex, fill=myorange!30!white] (y) at (3.4,1.5) {$C$};
        
        \draw[edge] (e) -- (r);
        \draw[edge] (l) -- (d);
        \draw[edge] (r) -- (i);
        \draw[edge] (e) -- (i);
        \draw[edge] (i) -- (s);
        \draw[edge] (l) -- (y);
        \draw[edge] (d) -- (y);
        \draw[edge] (i) -- (y);
        \draw[edge] (s) -- (y);
        \end{tikzpicture}
        \caption{German causal graph}
    \end{subfigure}
     \end{minipage}
  \end{minipage} 
\caption{\textbf{Additive Interventions.} 
(a) Intervention on \textit{Responsibility} with $\bm{F} = -0.3$ and the resulting effect on: (b) \textit{Loan Duration} and (c) \textit{Income}. 
(e) Intervention on \textit{Loan Amount} with $\bm{F} = 0.5$ and the resulting effect on: (f) \textit{Loan Duration} and (g) \textit{Income}.
In (d) and (h), the deep orange node represents the intervened variable, while the lighter orange nodes are those influenced by the intervention. In all plots, shaded regions denote $95\%$ confidence bounds. 
} \label{fig:additive_int_appendix}
\end{figure*}

\begin{figure*}[th!]
  \centering
  \includegraphics[width=0.98\textwidth]{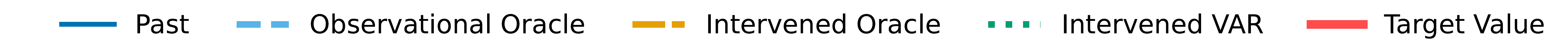}
  \begin{minipage}[t]{0.49\textwidth}
    \centering
    \begin{minipage}{\textwidth}
    \vspace{0.25cm} 
    \begin{subfigure}[b]{0.49\textwidth}
      \centering
      \includegraphics[width=\textwidth]{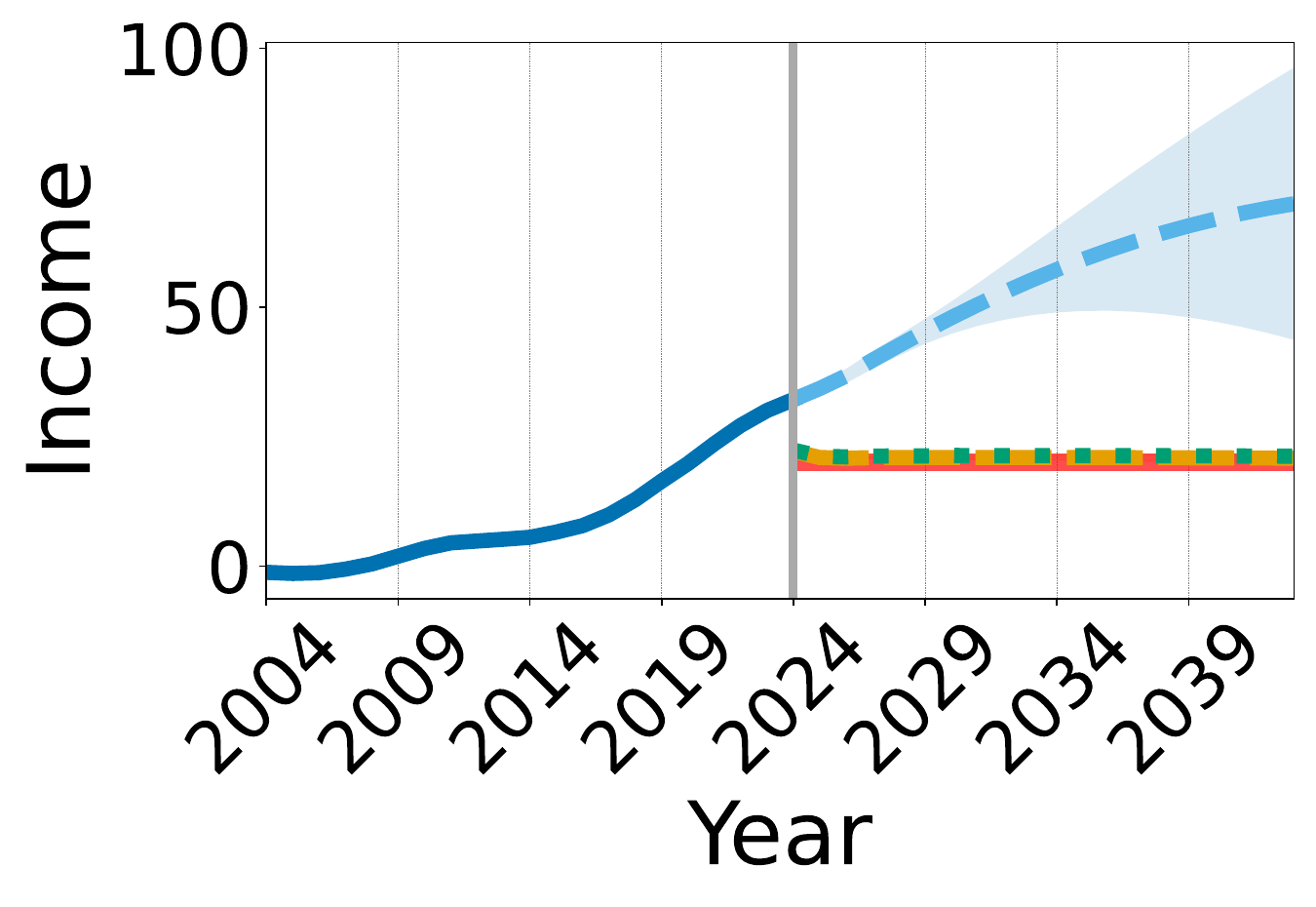}
      \caption{Intervention on $I$}
    \end{subfigure}
    \begin{subfigure}[b]{0.49\textwidth}
      \centering
      \includegraphics[width=\textwidth]{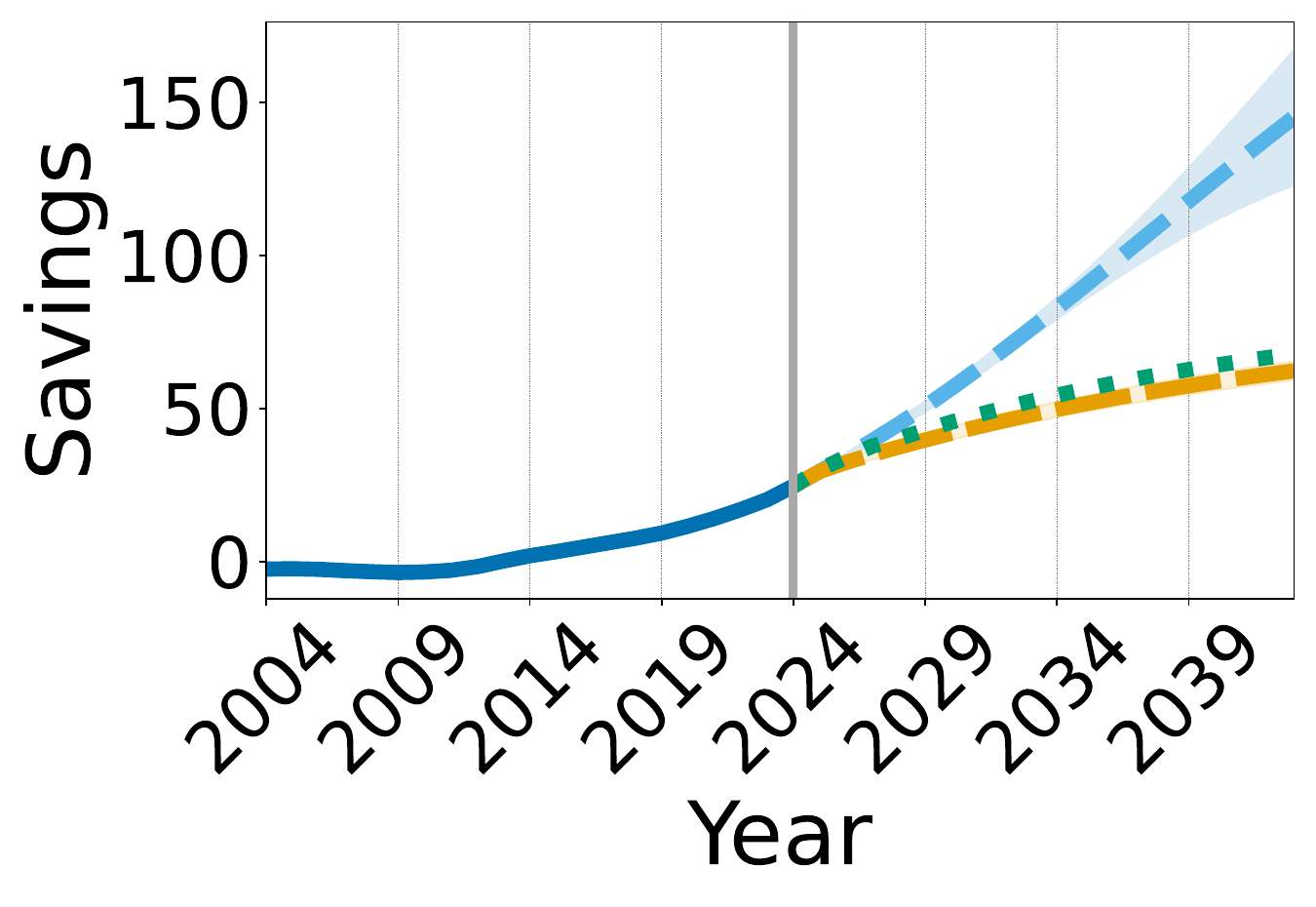}
      \caption{\centering Effect on $S$}
    \end{subfigure}
    \end{minipage}
  \end{minipage}
  \begin{minipage}[t]{0.49\textwidth}
    \centering
    \begin{minipage}{\textwidth}
    \vspace{0.25cm} 
    \begin{subfigure}[b]{0.49\textwidth}
      \centering
      \includegraphics[width=\textwidth]{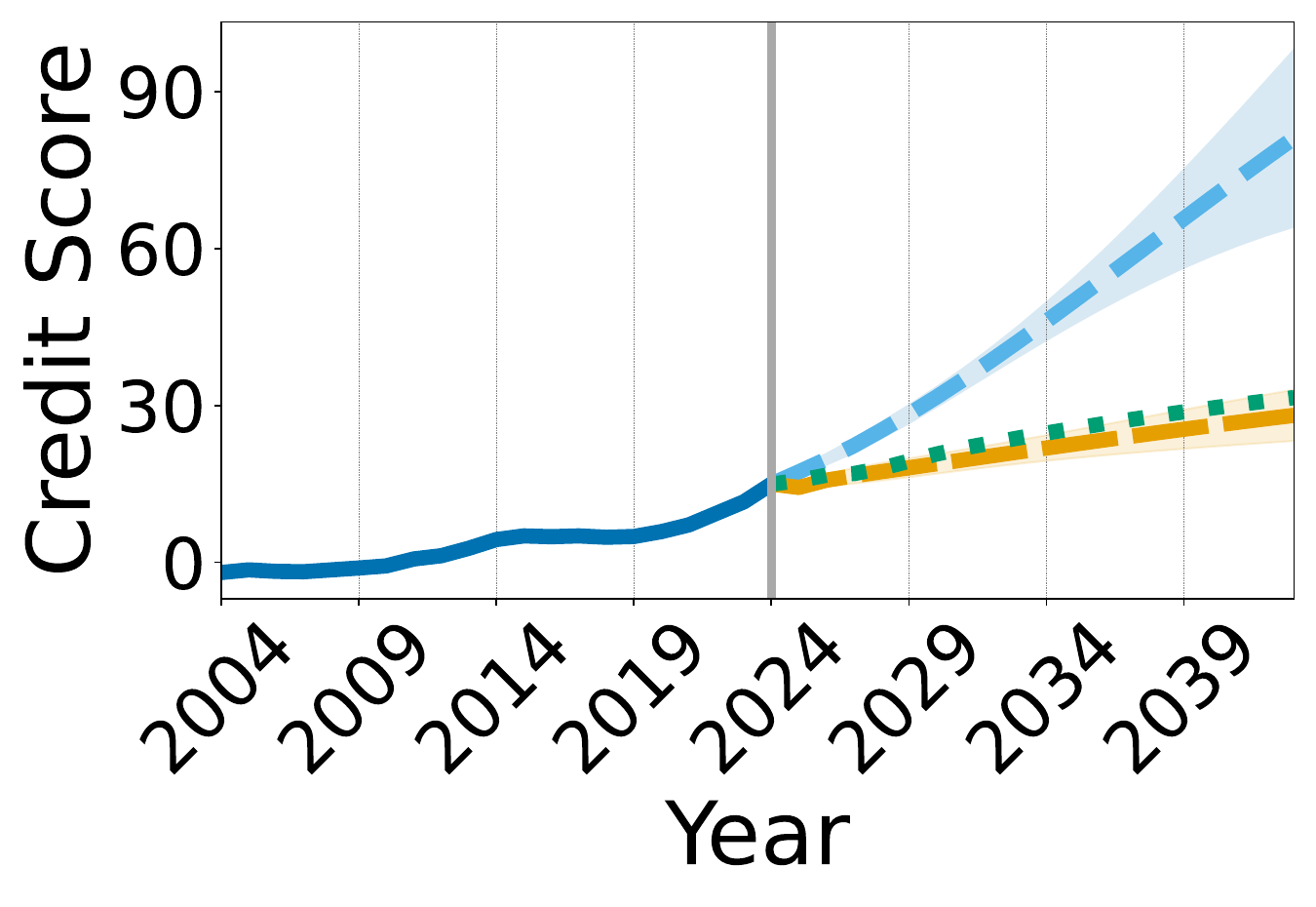}
      \caption{\centering Effect on $C$}
    \end{subfigure}
    \hfill
    \begin{subfigure}[b]{0.49\textwidth}
      \centering
      \begin{tikzpicture}[
            scale=0.73,
            transform shape,
            vertex/.style={circle, draw, minimum size=1cm, inner sep=0pt, font=\footnotesize}
        ]
        \node[vertex] (r) at (0,1.5) {$R$};
        \node[vertex] (e) at (0,0) {$E$};
        \node[vertex] (l) at (1.7,0) {$L$};
        \node[vertex] (d) at (3.4,0) {$D$};
        \node[vertex, fill=myorange!85!white] (i) at (1.7,1.5) {$I$};
        \node[vertex, fill=myorange!30!white] (s) at (1.7,3) {$S$};
        \node[vertex, fill=myorange!30!white] (y) at (3.4,1.5) {$C$};
        
        \draw[edge] (e) -- (r);
        \draw[edge] (l) -- (d);
        \draw[edge] (r) -- (i);
        \draw[edge] (e) -- (i);
        \draw[edge] (i) -- (s);
        \draw[edge] (l) -- (y);
        \draw[edge] (d) -- (y);
        \draw[edge] (i) -- (y);
        \draw[edge] (s) -- (y);
        \end{tikzpicture}
        \caption{German causal graph}
    \end{subfigure}
     \end{minipage}
  \end{minipage} 
  \centering
  \begin{minipage}[t]{0.49\textwidth}
    \centering
    \vspace{0.5cm}
    \begin{minipage}{\textwidth}
    \begin{subfigure}[b]{0.49\textwidth}
      \centering
      \includegraphics[width=\textwidth]{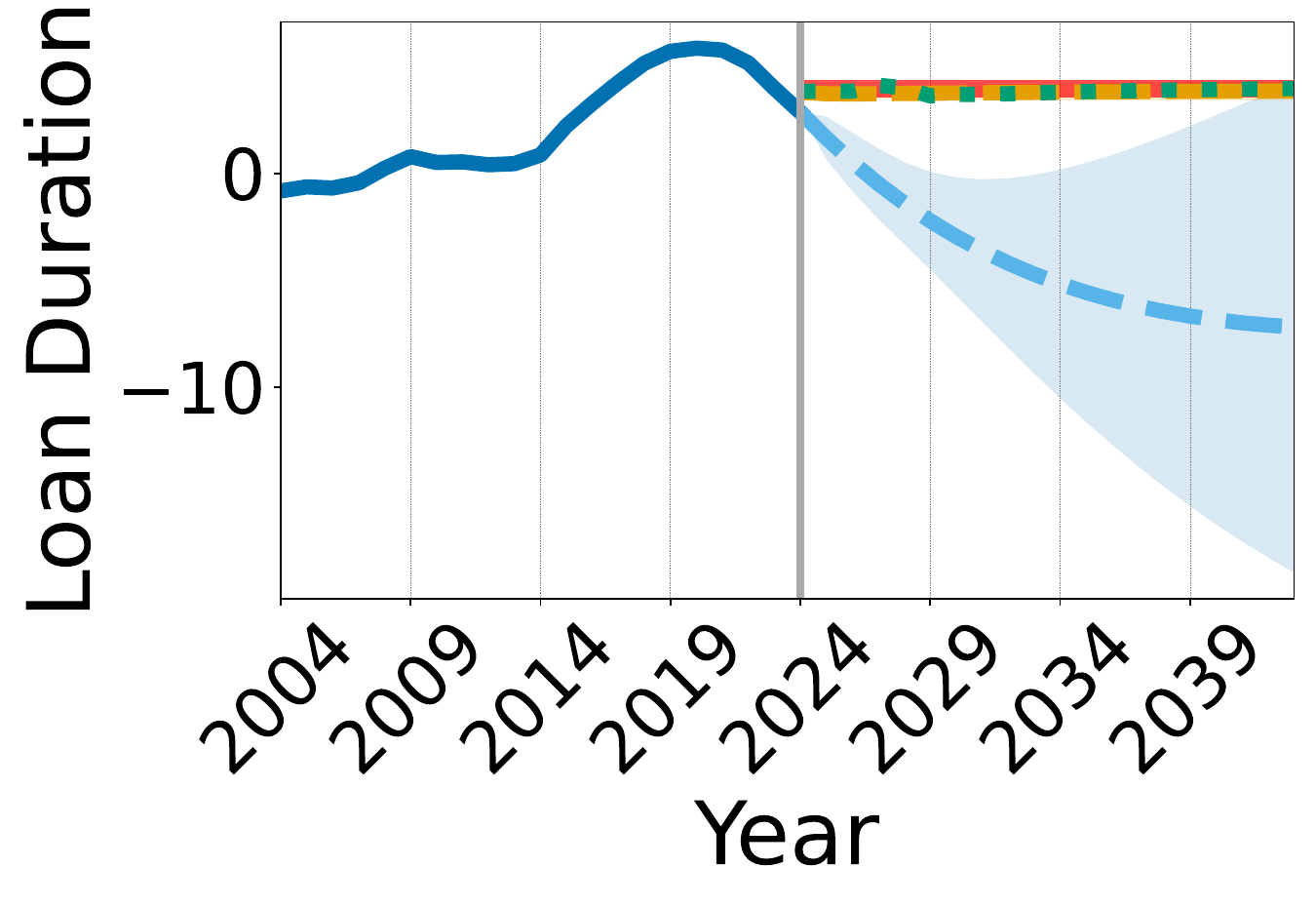}
      \caption{Intervention on $D$}
    \end{subfigure}
    \begin{subfigure}[b]{0.49\textwidth}
      \centering
      \includegraphics[width=\textwidth]{figures/Forcing_Intervention_on_Income_with_Force=20.pdf}
      \caption{\centering Intervention on $I$}
    \end{subfigure}
    \end{minipage}
  \end{minipage}
  \begin{minipage}[t]{0.49\textwidth}
    \centering
    \vspace{0.5cm}
    \begin{minipage}{\textwidth}
    \begin{subfigure}[b]{0.49\textwidth}
      \centering
      \includegraphics[width=\textwidth]{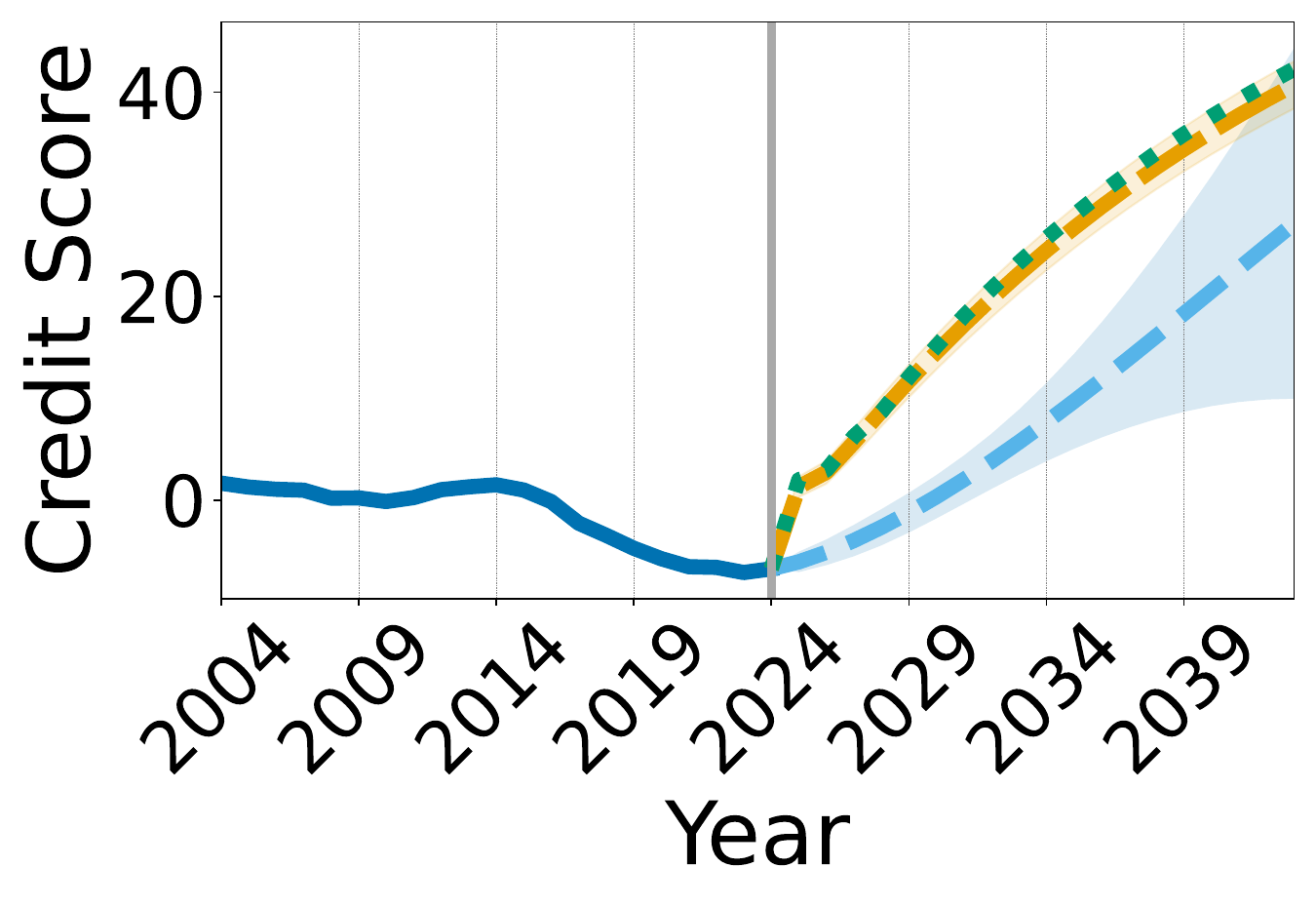}
      \caption{\centering Effect on $C$}
    \end{subfigure}
    \hfill
    \begin{subfigure}[b]{0.49\textwidth}
      \centering
      \begin{tikzpicture}[
            scale=0.73,
            transform shape,
            vertex/.style={circle, draw, minimum size=1cm, inner sep=0pt, font=\footnotesize}
        ]
        \node[vertex] (r) at (0,1.5) {$R$};
        \node[vertex] (e) at (0,0) {$E$};
        \node[vertex] (l) at (1.7,0) {$L$};
        \node[vertex, fill=myorange!85!white] (d) at (3.4,0) {$D$};
        \node[vertex, fill=myorange!85!white] (i) at (1.7,1.5) {$I$};
        \node[vertex, fill=myorange!30!white] (s) at (1.7,3) {$S$};
        \node[vertex, fill=myorange!30!white] (y) at (3.4,1.5) {$C$};
        
        \draw[edge] (e) -- (r);
        \draw[edge] (l) -- (d);
        \draw[edge] (r) -- (i);
        \draw[edge] (e) -- (i);
        \draw[edge] (i) -- (s);
        \draw[edge] (l) -- (y);
        \draw[edge] (d) -- (y);
        \draw[edge] (i) -- (y);
        \draw[edge] (s) -- (y);
        \end{tikzpicture}
        \caption{German causal graph}
    \end{subfigure}
     \end{minipage}
  \end{minipage} 
  \caption{\textbf{Forcing Interventions.}  
  (a) Intervention on \textit{Income} with $\bm{F} = 5$ and target value $\hat{I} = 20$.  The resulting effect on (b) \textit{Savings} and (c) \textit{Credit Score}. (e) Intervention on \textit{Loan Duration} with $\bm{F} = 3$ and target value $\hat{D} = 4$. (b) Intervention on \textit{Income} with $\bm{F} = 3$ and target value $\hat{I} = 20$. (g) Effect of interventions in (e-f) on \textit{Credit Score}.
  In (d) and (h), the deep orange node represents the intervened variable, while the lighter orange nodes are those influenced by the intervention. In all plots, shaded regions denote $95\%$ confidence bounds.
  }
\label{fig:forcing_int_appendix}
\end{figure*}

\paragraph{How do additive and forcing interventions affect the system dynamics?}
~\cref{fig:additive_int_appendix} illustrates two additive interventions applied to German. Panel (a) depicts an intervention on \textit{Responsibility} with $\bm{F} = -0.3$, while panel (e) shows an intervention on \textit{Loan Amount} with $\bm{F} = 0.5$.
The causal graphs in panels (d) and (h) show how these interventions propagate through the network. Panels (b) and (c) display the effects of the \textit{Responsibility} intervention on \textit{Loan Duration} and \textit{Income}, respectively. Panels (f) and (g) illustrate the corresponding effects for the \textit{Loan Amount} intervention.
Specifically, in panels (a-c), we observe that the intervention on \textit{Responsibility} affects \textit{Income} while having no impact on \textit{Loan Duration}. 
This behavior aligns with the causal relationships depicted in panel (d) as \textit{Responsibility} is not an ancestor of \textit{Loan Duration}. Similarly, panels (e-g) illustrate that the intervention on \textit{Loan Amount} affects only \textit{Loan Duration}, as \textit{Loan Amount} is not an ancestor of \textit{Income}.

~\cref{fig:forcing_int_appendix} shows two forcing interventions applied to German.
Panel (a) displays an intervention on \textit{Income} with $\bm{F} = 5$ and target value $\hat{I} = 20$. Panels (b) and (c) show how this intervention affects \textit{Savings} and \textit{Credit Score}, respectively.
Panel (e) presents an intervention on \textit{Loan Duration} with $\bm{F} = 3$ and target value $\hat{D} = 4$, while panel (f) shows a different intervention on \textit{Income} with $\bm{F} = 3$ and target value $\hat{I} = 20$. 
Panels (a-c) present results that are consistent with those demonstrated in the manuscript, namely that interventional forecasting precisely aligns with the specified target value.
Panels (e-f) demonstrate the capability of the causal \VAR{} framework to implement multiple interventions within the system simultaneously. Panel (g) displays the combined impact of these interventions on \textit{Credit Score}. 
By visualizing this cumulative effect, we could gain insights into the complex interplay between multiple variables and their joint influence on the target variable.

\begin{table*}[th!]
    \centering
    \caption{\textbf{Interventional Forecasting.} RMSE and SMAPE scores (\emph{lower is better}) for the proposed causal \VAR{} framework on German and Inverted Pendulum datasets. RMSE scores are scaled by a factor of $10^2$ to ease readability.
    Results averaged over ten runs, with standard deviation in subscript.}
    \label{tab:interventional_forecasting_combined}
    \setlength{\tabcolsep}{13pt}
    \footnotesize
    \renewcommand{\arraystretch}{1.18}
    \begin{tabular}{lccccccc}
    \toprule
    & & & \multicolumn{2}{c}{RMSE (scaled)} & \multicolumn{2}{c}{SMAPE} \\
    \cmidrule(lr){4-5} \cmidrule(lr){6-7}
    Dataset & Train Steps & Horizon & Additive & Forcing & Additive & Forcing \\
    \midrule
    German & \multirow{4}{*}{\centering100} & \multirow{2}{*}{\centering1} & $.000_{.000}$ & $.000_{.000}$ & $.000_{.000}$ & $.000_{.000}$ \\
    Pendulum & & & $.042_{.031}$ & $.062_{.046}$ & $.146_{.107}$ & $.319_{.236}$ \\
    \cmidrule(lr){3-7}
    German & & \multirow{2}{*}{\centering10} & $.446_{.295}$ & $1.53_{1.27}$ & $.829_{.547}$ & $5.80_{4.57}$ \\
    Pendulum & & & $6.66_{4.47}$ & $.102_{.051}$ & $2.19_{1.57}$ & $.262_{.140}$ \\
    \midrule
    German & \multirow{4}{*}{\centering500} & \multirow{2}{*}{\centering1} & $.000_{.000}$ & $.000_{.000}$ & $.000_{.000}$ & $.000_{.000}$ \\
    Pendulum & & & $.024_{.023}$ & $.035_{.034}$ & $.083_{.080}$ & $.180_{.173}$ \\
    \cmidrule(lr){3-7}
    German & & \multirow{2}{*}{\centering10} & $.173_{.139}$ & $.464_{.338}$ & $.353_{.276}$ & $1.89_{1.33}$ \\
    Pendulum & & & $2.02_{1.52}$ & $.049_{.035}$ & $.690_{.451}$ & $.111_{.077}$ \\
    \midrule
    German & \multirow{4}{*}{\centering1000} & \multirow{2}{*}{\centering1} & $.000_{.000}$ & $.000_{.000}$ & $.000_{.000}$ & $.000_{.000}$ \\
    Pendulum & & & $.010_{.007}$ & $.014_{.011}$ & $.034_{.026}$ & $.074_{.056}$ \\
    \cmidrule(lr){3-7}
    German & & \multirow{2}{*}{\centering10} & $.101_{.079}$ & $.299_{.220}$ & $.210_{.163}$ & $1.24_{.892}$ \\
    Pendulum & & & $1.11_{.853}$ & $.023_{.012}$ & $.376_{.257}$ & $.054_{.031}$ \\
    \bottomrule
    \end{tabular}
\end{table*}

\begin{table}[th!]
    \centering
    \caption{\textbf{Interventional Forecasting.} MAE scores (\emph{lower is better}) for the proposed causal \VAR{} on the Pendulum dataset. Results averaged over $10$ runs, with standard deviation in subscript. Scores are scaled by a factor of $10^3$ to ease readability.}\label{tab:interventional_forecasting_pendulum}
    \setlength{\tabcolsep}{5pt}
    \footnotesize
    \renewcommand{\arraystretch}{1.2}
    \begin{tabular}{ccccc}
    \toprule
    & & & \multicolumn{2}{c}{MAE} \\
    \cmidrule(lr){4-5}
    Dataset & Train Steps & Horizon & Additive & Forcing \\
    \midrule
    \multirow{4}{*}{Pendulum} &  \multirow{2}{*}{100} & 1 & $.094_{.069}$ & $.310_{.230}$ \\
& & 10 & $6.32_{4.44}$ & $.264_{.142}$ \\
\cmidrule(lr){2-5}
 & \multirow{2}{*}{500} & 1 & $.053_{.052}$ & $.174_{.169}$ \\
& & 10 & $1.95_{1.39}$ & $.112_{.078}$ \\
\bottomrule
    \end{tabular}
    \end{table}

\paragraph{Interventions} 
We provide the intervention type and the force values used to evaluate the interventional forecasting described in the following paragraph. 
For German, we perform causal interventions on root node \textit{Expertise}, 
and observe the resulting effect on the target variable \textit{Credit Score}.
For the additive case, we apply $\bm{F}=0.2$, while for the forcing case, we use $\bm{F}=1$ with a target value of $\hat{E} = 5$. 
For Pendulum, we perform causal interventions on $X^{(2)}$, and observe the resulting effect on $X^{(1)}$.
For the additive case, we apply $\bm{F}=0.4$, while for the forcing case, we use $\bm{F}=1$ with a target value of $\hat{X^{(2)}}= 1$.

\paragraph{How accurate is the causal VAR framework in estimating the causal effect of interventions over time?}
\cref{tab:interventional_forecasting_combined} provides RMSE and SMAPE scores for interventional forecasting across different data sizes and forecast horizons on the German and Pendulum datasets. These scores further validate the effectiveness of the proposed causal \VAR{} framework.
For $1$-step forecasts, the Pendulum dataset shows the intervention effect immediately, in contrast to the German dataset, where variables change slowly and require several time steps to exhibit measurable effects. 
For $10$-step forecasts, the scores demonstrate that the model achieves high accuracy. Notably, the RMSE values remain low across both datasets and intervention types, indicating the robustness of the model in estimating causal effects over longer horizons. Additionally, while the SMAPE scores show some variation, the results suggest that the model is well-suited for handling more complex forecasting scenarios. 

\cref{tab:interventional_forecasting_pendulum} outlines the results of interventional forecasting, detailing errors across different data sizes and forecast horizons for the Pendulum dataset. 
Our causal effect estimates remain highly accurate for both forecast horizons. In contrast to the German dataset in~\cref{tab:interventional_forecasting_german}, the intervention effect is observable even at the $1$-step forecast.
These results reinforce the findings presented in the main paper, highlighting the effectiveness of the proposed causal \VAR{} framework in interventional forecasting tasks.

\begin{figure*}[th!]
  \centering
  \includegraphics[width=0.98\textwidth]{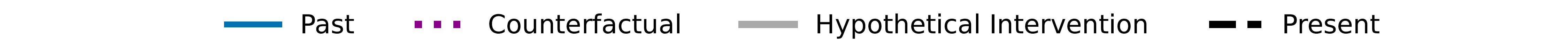}
  \begin{minipage}[t]{0.49\textwidth}
    \centering
    \begin{minipage}{\textwidth}
    \vspace{0.5cm} 
    \begin{subfigure}[b]{0.49\textwidth}
      \centering
      \includegraphics[width=\textwidth]{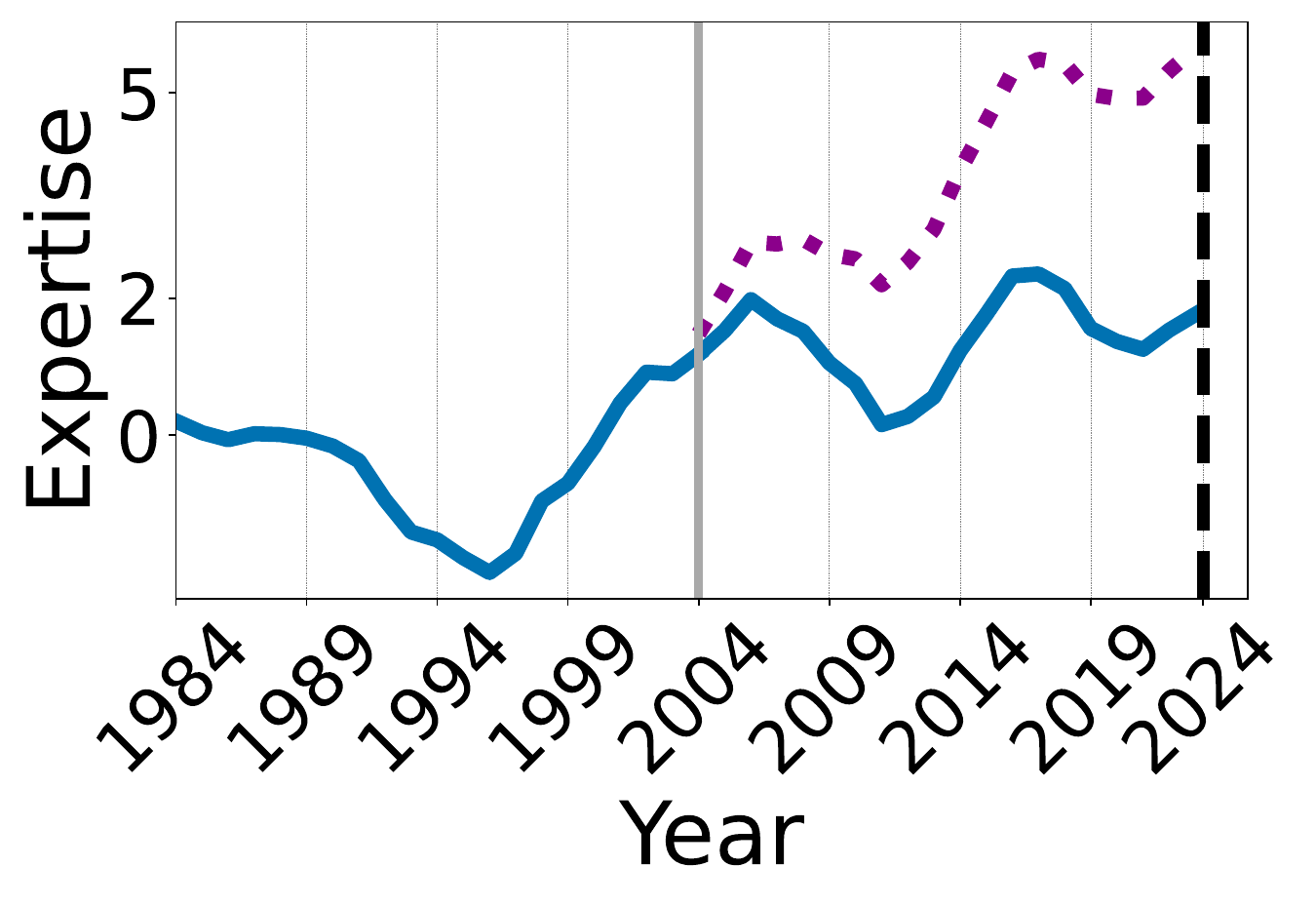}
      \caption{Intervention on $E$}
    \end{subfigure}
    \begin{subfigure}[b]{0.49\textwidth}
      \centering
      \includegraphics[width=\textwidth]{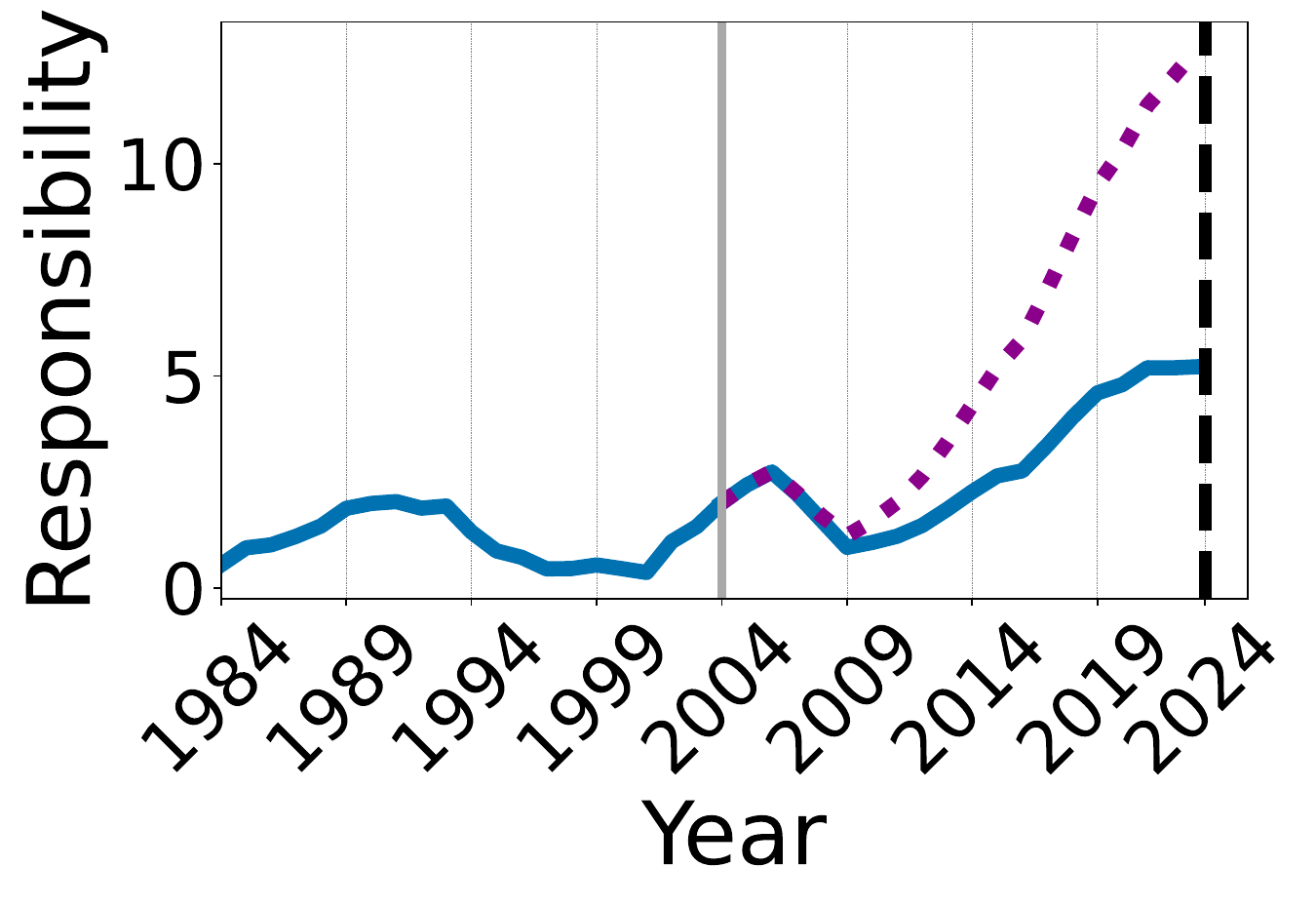}
      \caption{\centering Effect on $R$}
    \end{subfigure}
    \end{minipage}
  \end{minipage}
  \begin{minipage}[t]{0.49\textwidth}
    \centering
    \begin{minipage}{\textwidth}
    \vspace{0.5cm} 
    \begin{subfigure}[b]{0.49\textwidth}
      \centering
      \includegraphics[width=\textwidth]{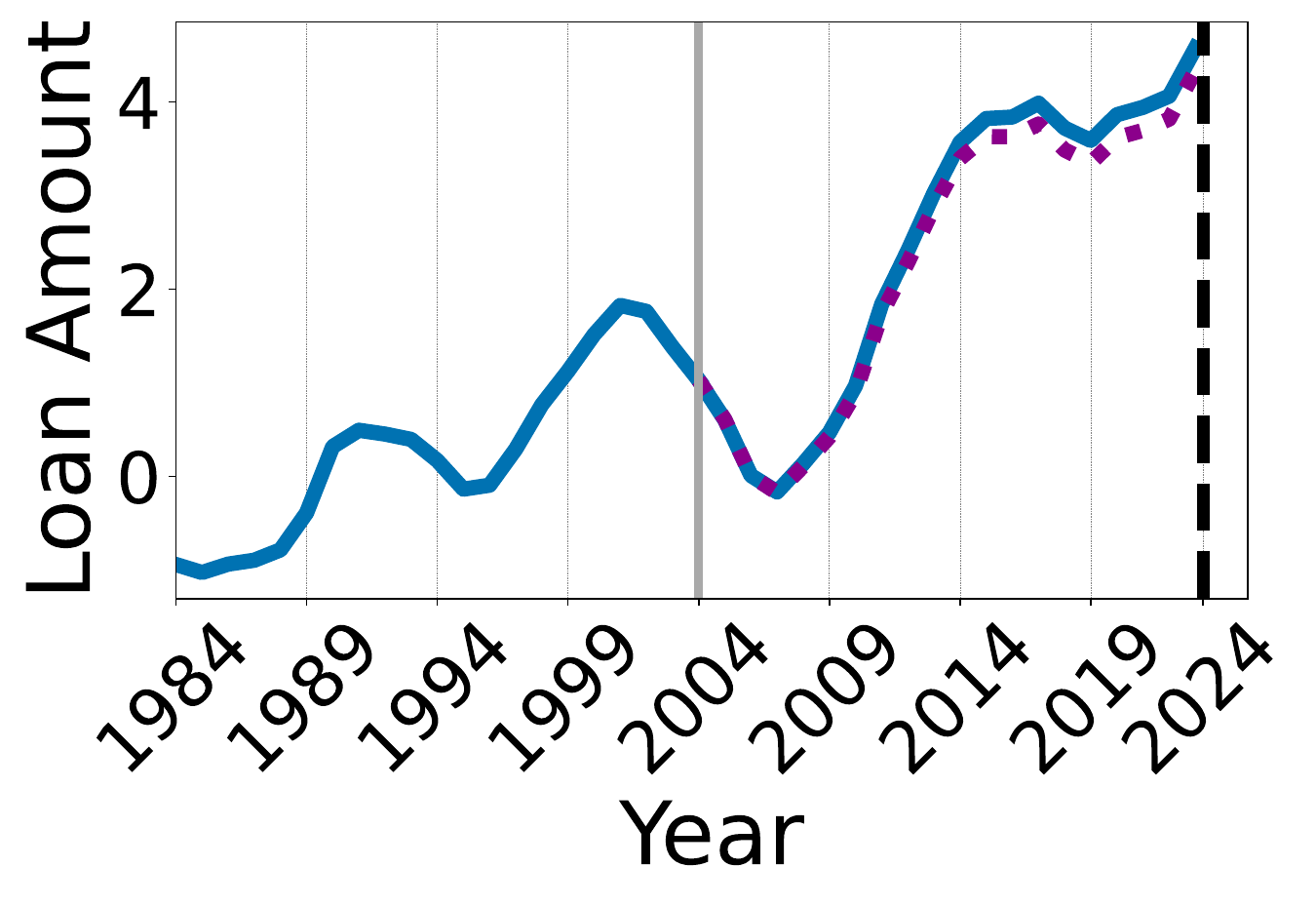}
      \caption{\centering Effect on $L$}
    \end{subfigure}
    \hfill
    \begin{subfigure}[b]{0.49\textwidth}
      \centering
      \begin{tikzpicture}[
            scale=0.73,
            transform shape,
            vertex/.style={circle, draw, minimum size=1cm, inner sep=0pt, font=\footnotesize}
        ]
        \node[vertex, fill=mypurple!15!white] (r) at (0,1.5) {$R$};
        \node[vertex, fill=mypurple!45!white] (e) at (0,0) {$E$};
        \node[vertex] (l) at (1.7,0) {$L$};
        \node[vertex] (d) at (3.4,0) {$D$};
        \node[vertex, fill=mypurple!15!white] (i) at (1.7,1.5) {$I$};
        \node[vertex, fill=mypurple!15!white] (s) at (1.7,3) {$S$};
        \node[vertex, fill=mypurple!15!white] (y) at (3.4,1.5) {$C$};
        
        \draw[edge] (e) -- (r);
        \draw[edge] (l) -- (d);
        \draw[edge] (r) -- (i);
        \draw[edge] (e) -- (i);
        \draw[edge] (i) -- (s);
        \draw[edge] (l) -- (y);
        \draw[edge] (d) -- (y);
        \draw[edge] (i) -- (y);
        \draw[edge] (s) -- (y);
        \end{tikzpicture}
        \caption{German causal graph}
    \end{subfigure}
     \end{minipage}
  \end{minipage} 
  \centering

  \begin{minipage}[t]{0.49\textwidth}
    \centering
    \begin{minipage}{\textwidth}
    \vspace{0.5cm} 
    \begin{subfigure}[b]{0.49\textwidth}
      \centering
      \includegraphics[width=\textwidth]{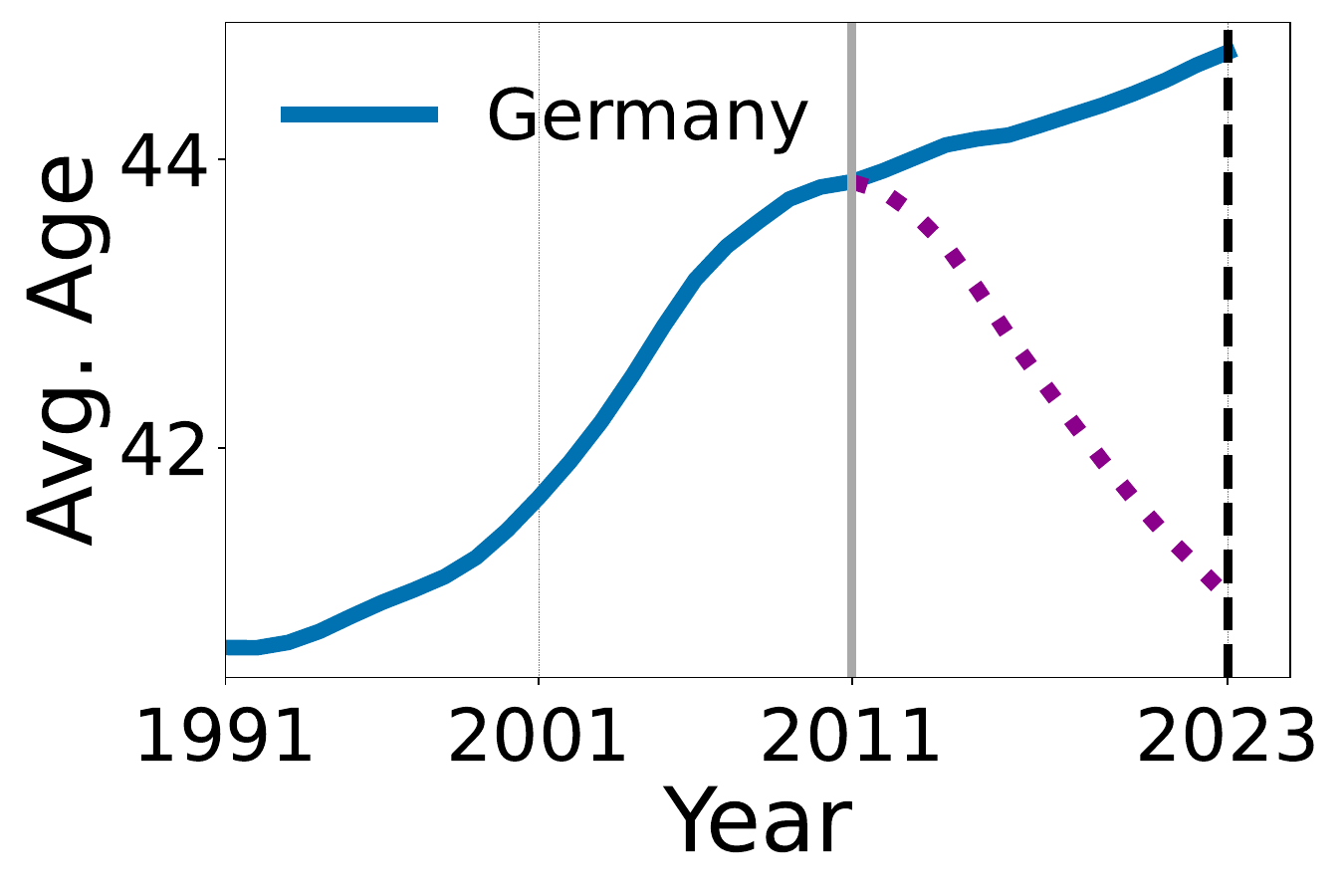}
      \caption{\centering Germany}
    \end{subfigure}
    \begin{subfigure}[b]{0.49\textwidth}
      \centering
      \includegraphics[width=\textwidth]{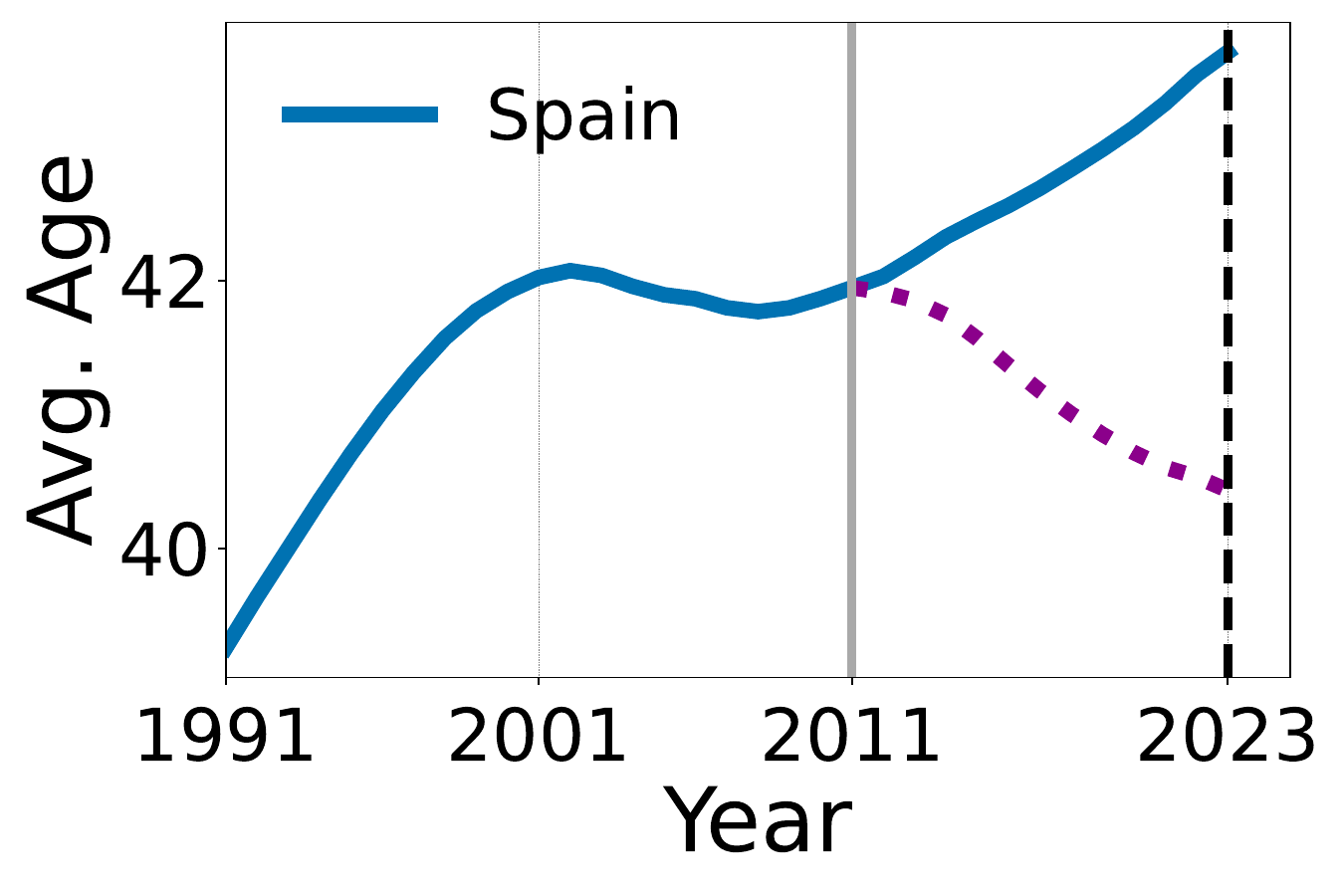}
      \caption{\centering Spain}
    \end{subfigure}
    \end{minipage}
  \end{minipage}
  \centering
  \begin{minipage}[t]{0.49\textwidth}
    \centering
    \begin{minipage}{\textwidth}
    \vspace{0.5cm} 
    \begin{subfigure}[b]{0.49\textwidth}
      \centering
      \includegraphics[width=\textwidth]{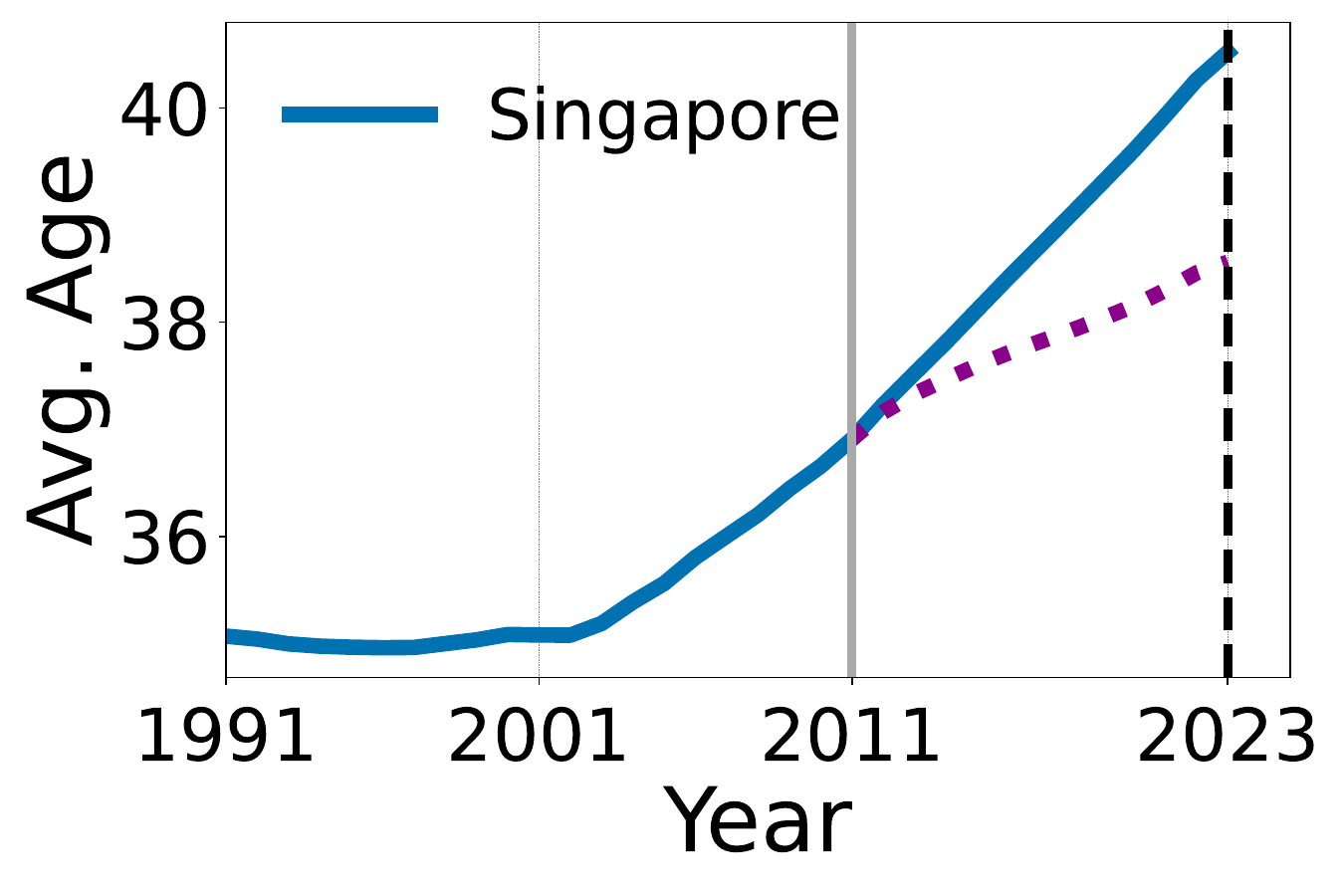}
      \caption{\centering Singapore}
    \end{subfigure}
    \hfill
    \begin{subfigure}[b]{0.49\textwidth}
      \centering
      \begin{tikzpicture}[
            scale=0.68,
            transform shape,
            vertex/.style={circle, draw, minimum size=1.1cm, inner sep=0pt, font=\small}
        ]
        \node[vertex, fill=mypurple!45!white] (b) at (0, 1.5) {Birth};
        \node[vertex, fill=mypurple!15!white] (d) at (3.4, 3) {Death};
        \node[vertex] (m) at (3.4,0) {Migra.};
        \node[vertex, fill=mypurple!15!white] (a-0) at (1.7,1.5) {0-14};
        \node[vertex, fill=mypurple!15!white] (a-1) at (3.4,1.5) {15-64};
        \node[vertex, fill=mypurple!15!white] (a-2) at (5.1,1.5) {65-99};
        
        \draw[edge] (b) -- (a-0);
        \draw[edge] (a-1) to [out=120,in=60](b); 
        \draw[edge] (d) -- (a-2);
        \draw[edge] (d) -- (a-1);
        \draw[edge] (d) -- (a-0);
        \draw[edge] (m) -- (a-0);
        \draw[edge] (m) -- (a-1);
        \draw[edge] (m) -- (a-2);
        \draw[edge] (a-0) -- (a-1);
        \draw[edge] (a-1) -- (a-2);
        \draw[edge] (m) to [out=190,in=-60](b);
        \draw[edge] (a-2) to [out=90, in=0](d);
        \end{tikzpicture}
        \caption{Census causal graph}
    \end{subfigure}
     \end{minipage}
  \end{minipage} 
  \caption{\textbf{Retrospective Counterfactuals.}
  Panels (a-d) focus on the German dataset.  
  (a) Intervention on \textit{Expertise} with $\bm{F} = 0.3$. 
  Effect of the intervention on (b) \textit{Responsibility} and (c) \textit{Loan Amount}. 
  Panels (e-h) examine the Census dataset with an intervention on \textit{Births} with $\bm{F} = 0.004$. Effect of the intervention on the population's average age of (e) Germany, (f) Spain, and (g) Singapore. 
  In (d) and (h), the deep purple node represents the intervened variable, while the lighter purple nodes are those influenced by the intervention.}
  \label{fig:counterfactuals}
\end{figure*}
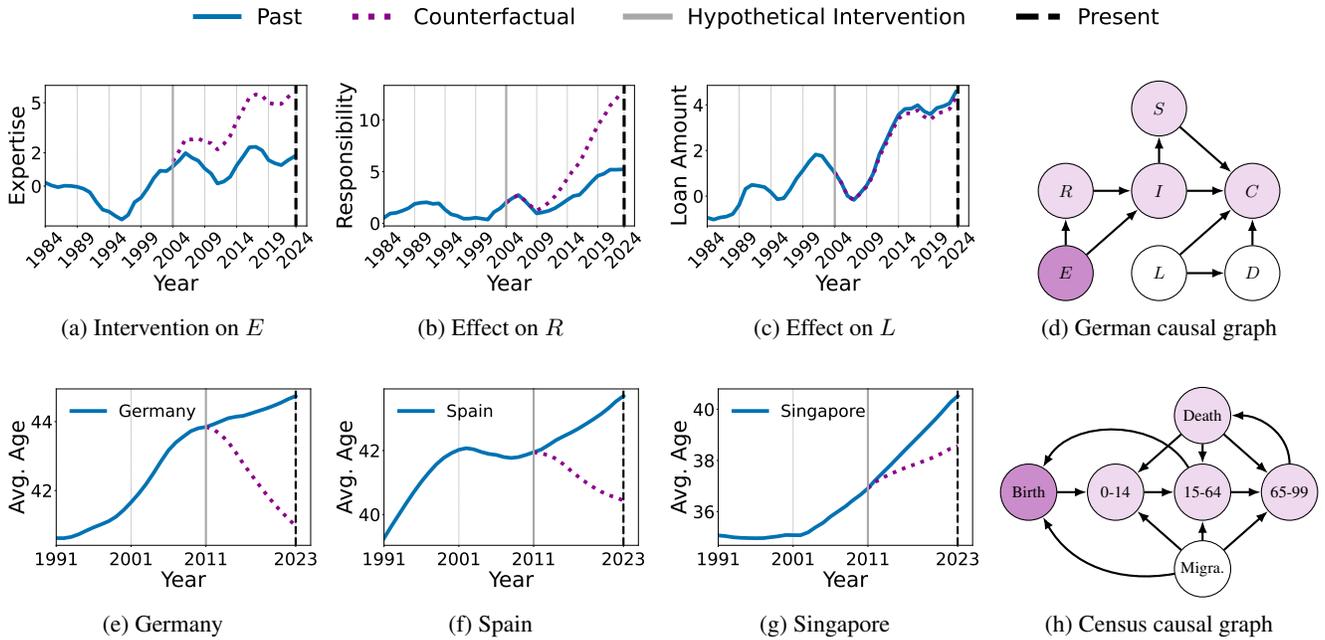

\subsection{Retrospective Counterfactuals}\label{app:counterfactuals}
In this section, we demonstrate that our causal VAR framework can also address the causal queries discussed~\cref{subsec:practical_implications} in terms of retrospective counterfactuals.

\cref{fig:counterfactuals} illustrates two counterfactual interventions, with panels (a-d) related to German and panels (e-h) focused on Census.
For German, panel (a) shows an intervention in 2004 on \textit{Expertise} with $\bm{F} = 0.3$ and the effect of this intervention on (b) \textit{Responsibility} and (c) \textit{Loan Amount}. 
This scenario could potentially answer the question: ``What would be the outcome in 2024 if an applicant had improved their expertise in 2004?''. 
Panel (b) shows that an increase in \textit{Expertise} would have led to a corresponding rise in \textit{Responsibility}. Notably, panel (c) reveals the \textit{Loan Amount} remains unchanged, reflecting the structure of our causal graph where \textit{Expertise} is not an ancestor of this node.
For the Census dataset, we examine an intervention in 2011 on \textit{Births} with $\bm{F} = 0.004$ and its effect on the population's average age (computed as a
weighted mean of age groups) of three countries: (e) Germany, (f) Spain, and (g) Singapore. 
The force value indicates that \textit{Births} increase by $0.4\%$ of each country's total population annually.
This analysis could potentially explore the question: ``How would a policy promoting increased births in 2011 have affected a country’s current average population age?''. 
Panels (e-f) display that such an intervention would have significantly lowered the average age in Germany and Spain, given their relatively older population structures. 
In contrast, panel (g) shows that the effect on Singapore would have been more moderate, consistent with its younger demographic profile at the intervention time.

\end{document}